\documentclass[runningheads]{llncs}

\usepackage{eccv}

\usepackage{eccvabbrv}
\usepackage{graphicx}
\usepackage{booktabs}
\usepackage{subcaption}
\usepackage{algorithm}
\usepackage{algpseudocode}
\usepackage{amsmath}
\usepackage{amssymb}
\usepackage{multirow}
\usepackage{pifont}
\usepackage{tabularx}
\usepackage{xcolor}
\usepackage{xspace}
\usepackage{tikz}
\usepackage{float}
\usepackage{caption}
\usepackage[accsupp]{axessibility}
\usepackage{hyperref}

\usepackage{soul}
\usepackage{mwe}
\usepackage{multirow}
\usepackage{listings}
\usepackage{xcolor}
\usepackage{makecell}

\lstdefinestyle{promptstyle}{
    basicstyle=\small\ttfamily,
    breaklines=true,
    frame=l,
    framesep=10pt,
    xleftmargin=10pt,
    moredelim=**[is][\color{blue}\bfseries]{@}{@},
    aboveskip=1em,
    belowskip=1em,
    breakindent=0pt,
    breakautoindent=false,
    columns=fullflexible,
    keepspaces=true
}

\makeatletter
\DeclareRobustCommand\onedot{\futurelet\@let@token\@onedot}
\def\@onedot{\ifx\@let@token.\else.\null\fi\xspace}

\def\eg{\emph{e.g}\onedot} 
\def\ie{\emph{i.e}\onedot}

\def\etal{\emph{et al}\onedot}
\makeatother

\def \customparskip {.3em}
\renewcommand{\paragraph}[1]{\vspace{\customparskip}\noindent\textbf{#1}}

\newcommand{\bx}{\mathbf{x}}
\newcommand{\by}{\mathbf{y}}
\newcommand{\bA}{\mathbf{A}}
\newcommand{\bX}{\mathbf{X}}
\newcommand{\fourier}{\mathcal{F}}
\newcommand{\denoise}{\mathcal{D}}
\newcommand{\stopgrad}[1]{\left[ #1 \right]_{\text{sg}}}

\newcommand{\loss}[1]{\mathcal{L}_{\text{#1}}}

\newcommand{\supp}[0]{\texttt{Supplementary Material}\xspace}

\newcommand{\flair}[0]{FLAIR~\cite{flair}\xspace}
\newcommand{\bld}[0]{BLD-SD3.5~\cite{Avrahami_2023,esser2024}\xspace}
\newcommand{\flowchef}[0]{FlowChef~\cite{flowchef}\xspace}
\newcommand{\flowdps}[0]{FlowDPS~\cite{flowdps}\xspace}
\newcommand{\brushnet}[0]{BrushNet~\cite{ju2024brushnet}\xspace}
\newcommand{\sdthreefive}[0]{StableDiffusion~3.5~\cite{esser2024}\xspace}

\newcommand{\ssim}[0]{\cite{wang2004ssim}}
\newcommand{\lpips}[0]{\cite{zhang2018unreasonableeffectivenessdeepfeatures}}
\newcommand{\fid}[0]{\cite{NIPS2017_8a1d6947}}

\newcommand{\ffhq}[0]{FFHQ~\cite{karras2019stylegan}\xspace}
\newcommand{\divtwok}[0]{DIV2K~\cite{div2k}\xspace}
\newcommand{\brushbench}[0]{BrushBench~\cite{ju2024brushnet}\xspace}

\definecolor{color1}{rgb}{0.9, 0.65, 0.65}
\definecolor{color2}{rgb}{0.95, 0.8, 0.8}
\definecolor{color3}{rgb}{1.0, 0.9, 0.9}
\newcommand{\best}[1]{\hspace{-\fboxsep}\colorbox{color1}{\textbf{#1}}\hspace{-\fboxsep}}
\newcommand{\second}[1]{\hspace{-\fboxsep}\colorbox{color2}{\textit{#1}}\hspace{-\fboxsep}}
\newcommand{\third}[1]{\hspace{-\fboxsep}\colorbox{color3}{\textit{#1}}\hspace{-\fboxsep}}

\begin{document}

\title{SONIC: Spectral Optimization of Noise for Inpainting with Consistency}
\titlerunning{SONIC}

\author{
Seungyeon Baek\inst{1} \and
Erqun Dong\inst{1} \and
Shadan Namazifard\inst{1} \and
Mark J. Matthews\inst{2}\thanks{Participated in an advisory capacity only.} \and
Kwang Moo Yi\inst{1}}
\authorrunning{S. Baek et al.}

\institute{
University of British Columbia, Canada\\
\email{\{syeonb, eqdong, kmyi\}@cs.ubc.ca}\\
\email{shadan82@student.ubc.ca}
\and
Google DeepMind, USA\\
\email{mjmatthews@google.com}
}

\maketitle

\begin{figure*}[t]
\centering
\setlength{\tabcolsep}{2pt}
\renewcommand{\arraystretch}{0.85}
\resizebox{\linewidth}{!}{
\begin{tabular}{ccccc}
    \includegraphics[width=0.195\linewidth]{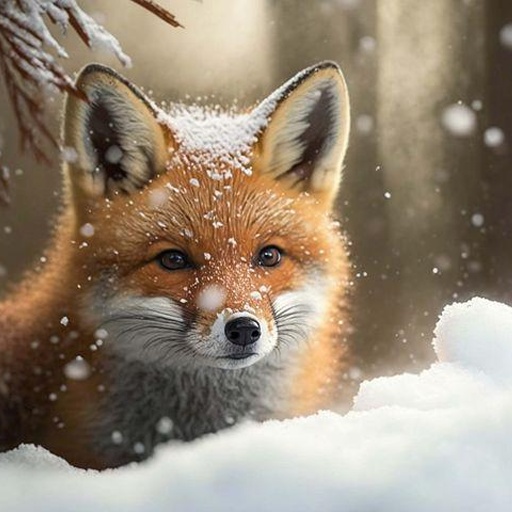} &
    \includegraphics[width=0.195\linewidth]{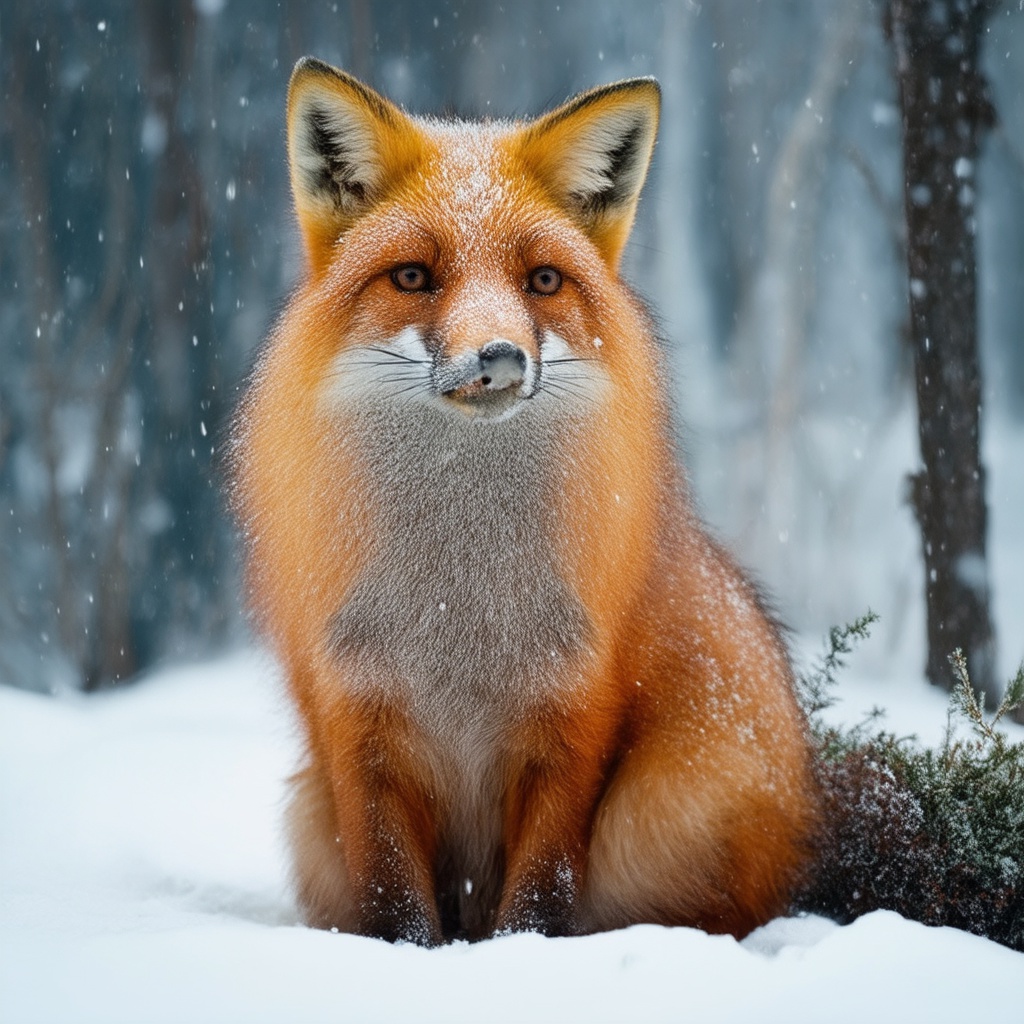} &
    \includegraphics[width=0.195\linewidth]{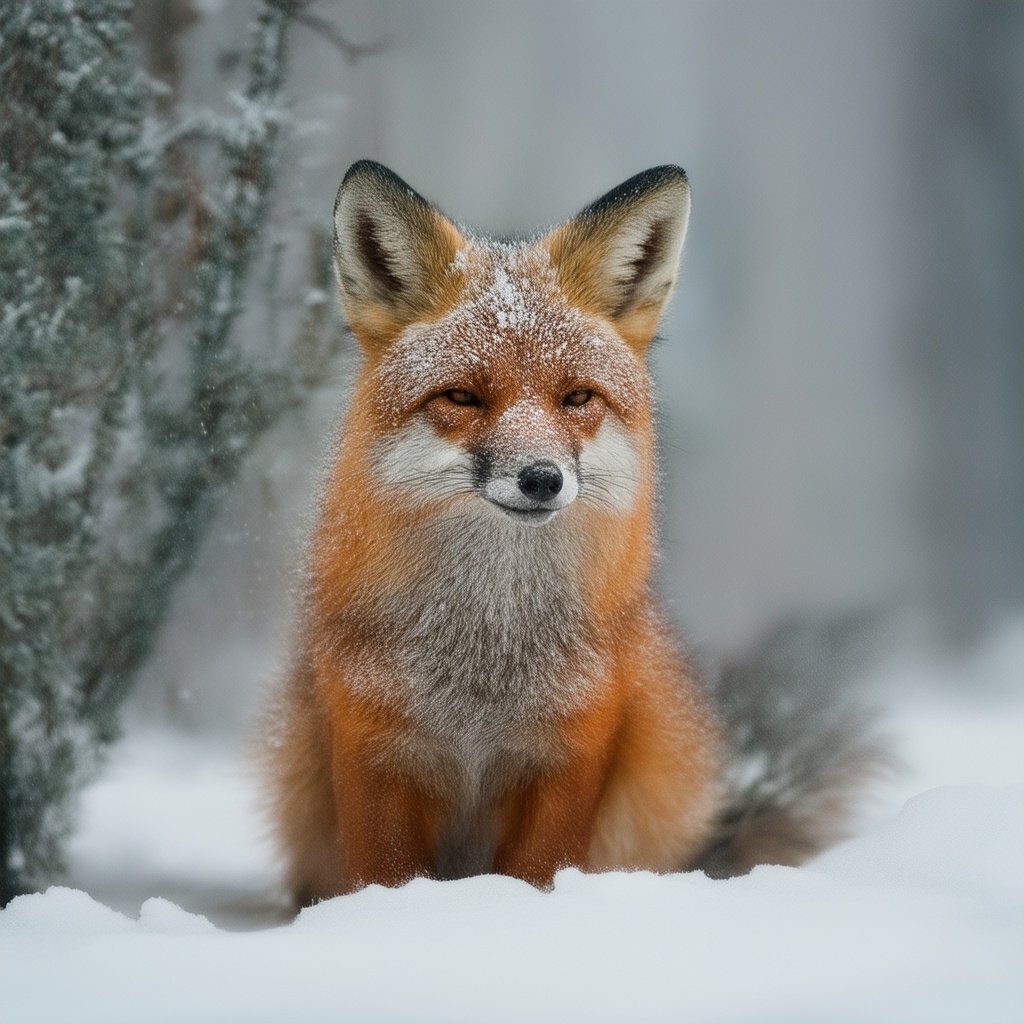} &
    \includegraphics[width=0.195\linewidth]{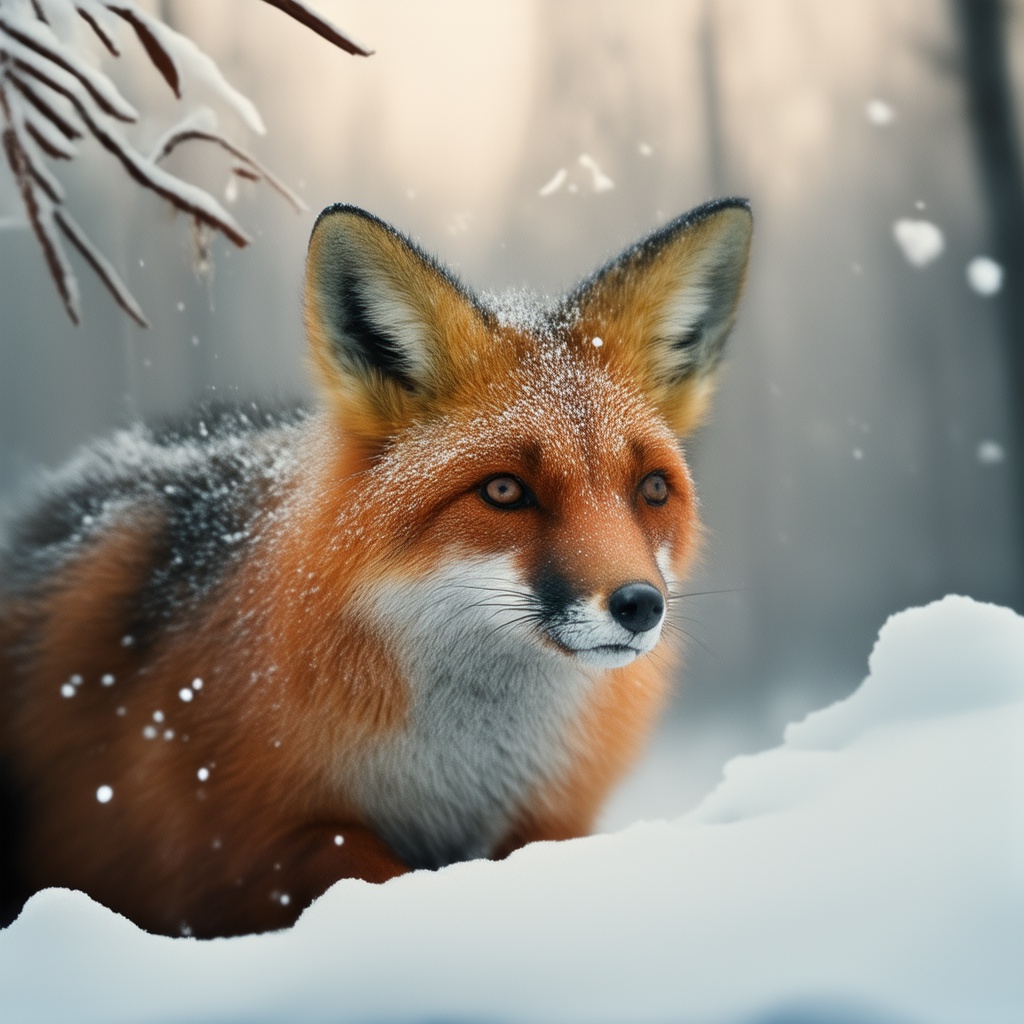} &
    \includegraphics[width=0.195\linewidth]{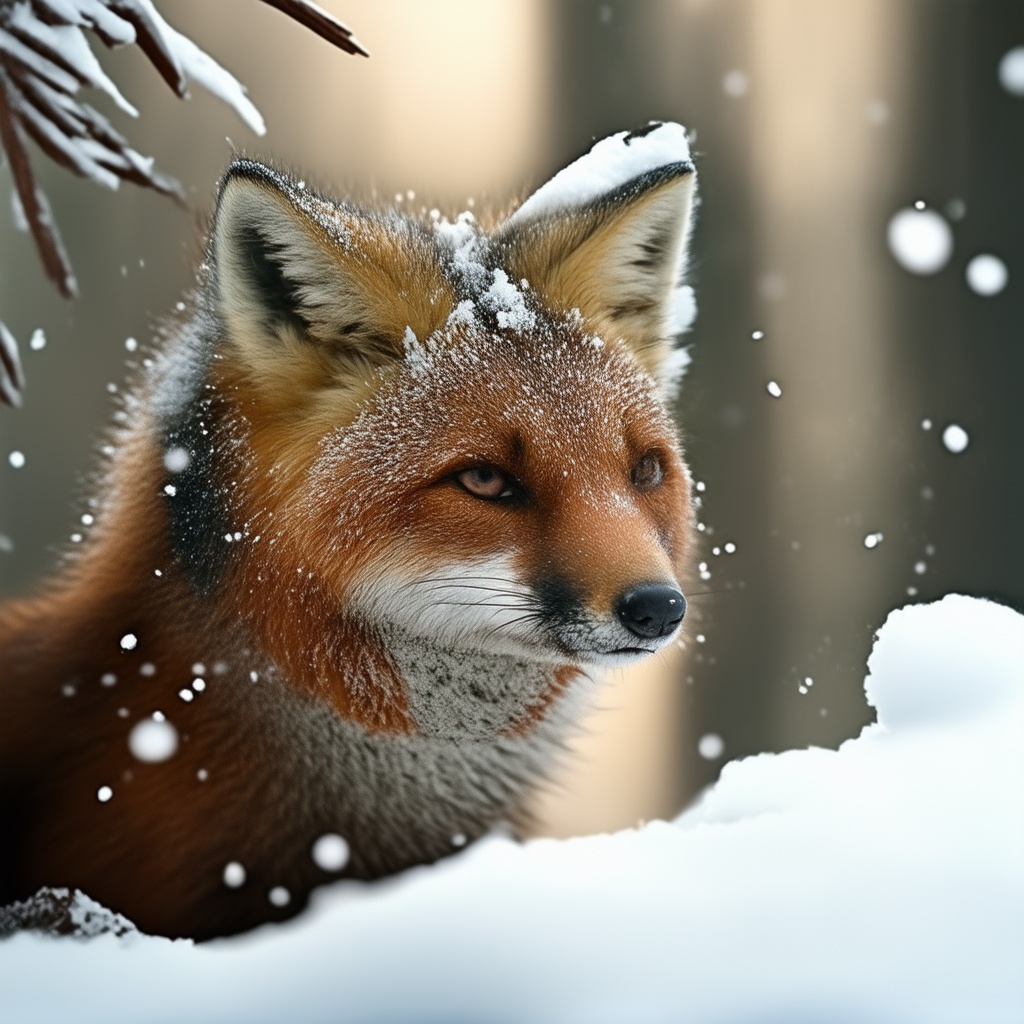} \\
    {\tiny Ground truth} &
    {\tiny Ours (0 iterations)} &
    {\tiny Ours (5 iterations)} &
    {\tiny Ours (10 iterations)} &
    {\tiny Ours (20 iterations)} \\[0.5em]
    \includegraphics[width=0.195\linewidth]{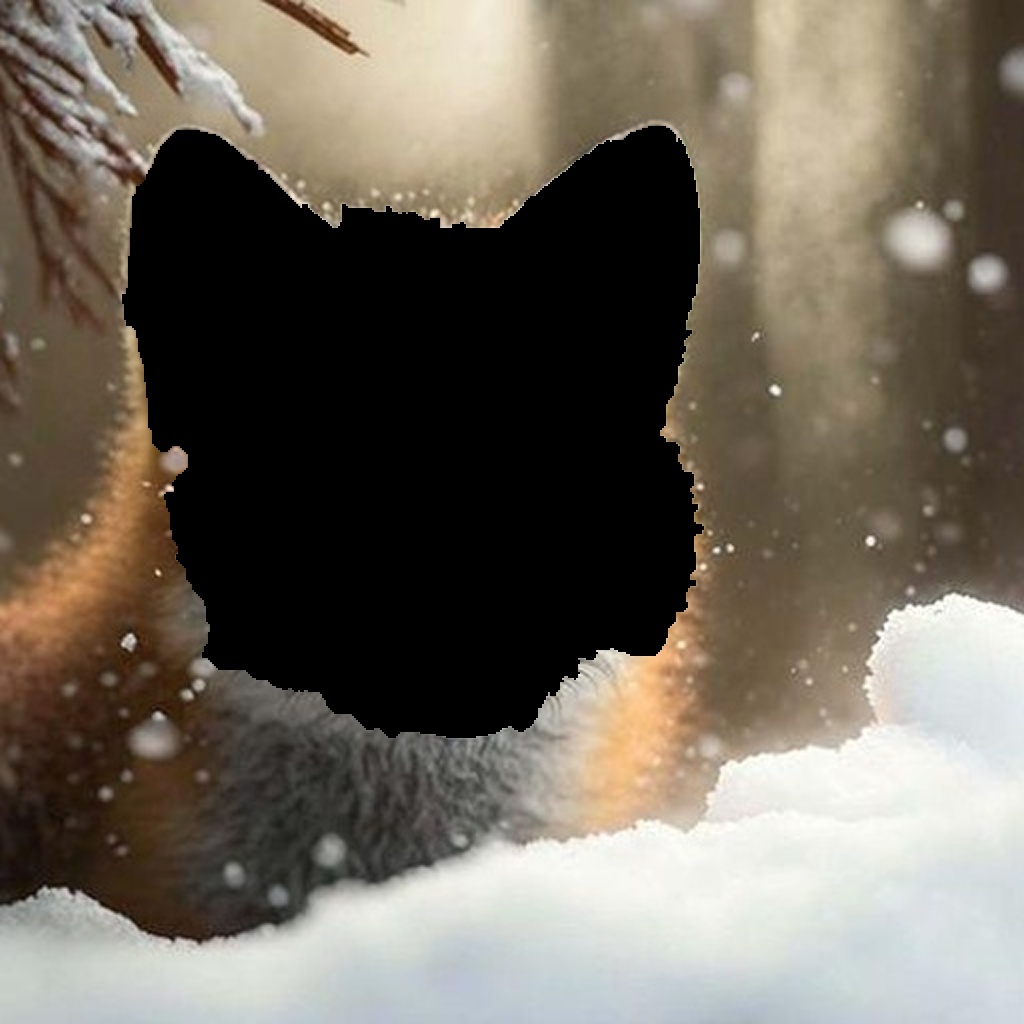} &
    \includegraphics[width=0.195\linewidth]{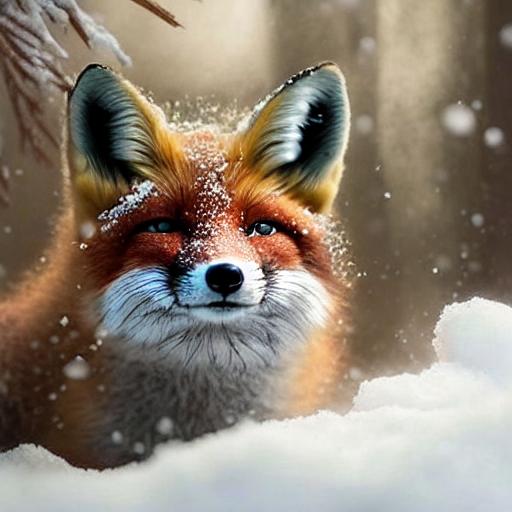} &
    \includegraphics[width=0.195\linewidth]{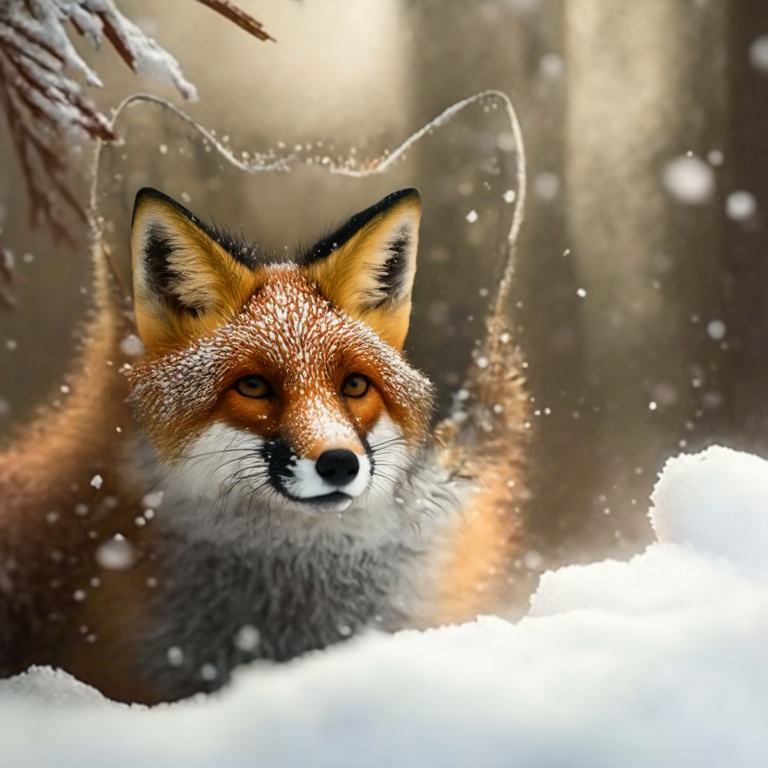} &
    \includegraphics[width=0.195\linewidth]{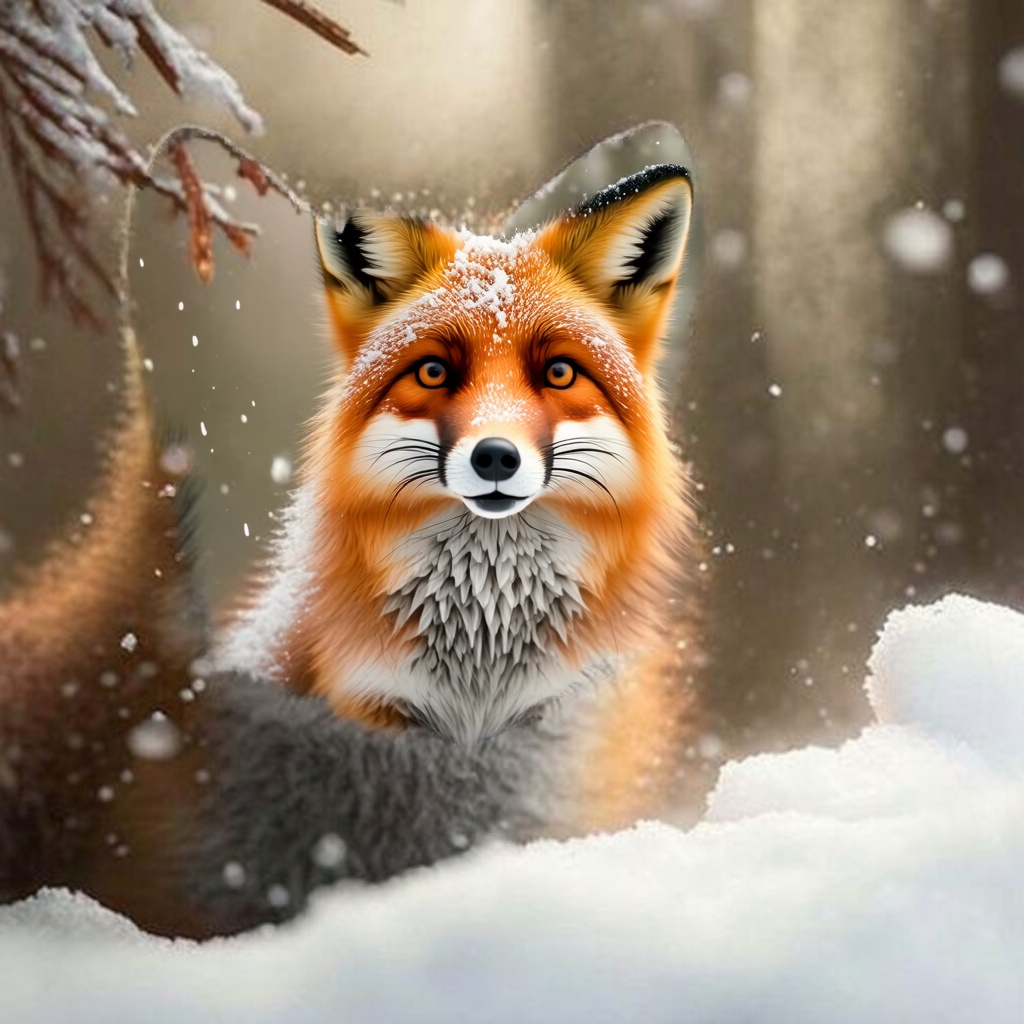} &
    \includegraphics[width=0.195\linewidth]{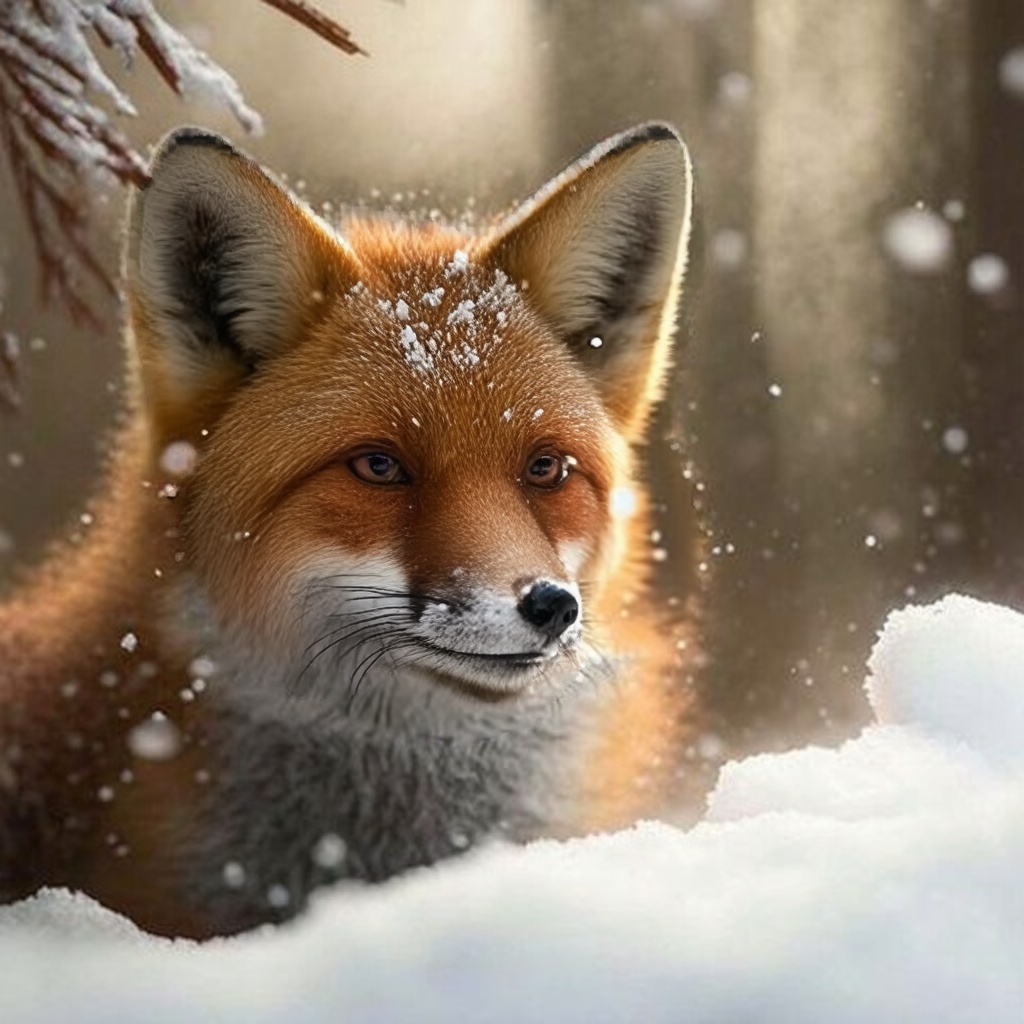} \\
    {\tiny Input image} &
    {\tiny \brushnet} &
    {\tiny \flair} &
    {\tiny \bld} &
    {\tiny Ours (19 iter.) + \cite{Avrahami_2023}}
\end{tabular}
}
\renewcommand{\arraystretch}{1.0}
\setlength{\tabcolsep}{6pt}
\vspace{-0.5em}
\caption{
    {\bf Teaser --} We propose a novel training-free method of inpainting that focuses exclusively on the initial noise sample.
    {\bf (Top row)} We show the denoising result of an initial noise sample, as we optimize the seed noise using our method.
    We optimize the seed noise to faithfully regenerate the non-masked regions of the input image, so as to obtain more consistent inpainting results.
    {\bf (Bottom row)} Inpainting results of competing methods, with our final result on the right.
}
\label{fig:teaser}
\vspace{-2em}
\end{figure*}

\setcounter{footnote}{0}

\begin{abstract}
We propose a novel training-free method for inpainting with off-the-shelf text-to-image models.
While guidance-based methods in theory allow \emph{generic} models to be used for inverse problems such as inpainting --- in practice their effectiveness is limited, leading to the necessity of \emph{specialized} inpainting-specific models.
In this work, we argue the missing ingredient for training-free generic model usage is proper optimization of the \emph{initial noise sample}.
We optimize the initial noise to approximately reproduce the unmasked image, in as few as tens of optimization steps, then use it with a conventional training-free inpainting method.
Critically, we propose two core ideas that make this possible: (i) we perform linear approximation that avoids the costly and often impractical unrolling required to relate the initial noise sample to model output---which potentially is why this relationship was previously overlooked; and (ii) perform spectral preconditioning by optimizing the initial noise sample in the spectral domain with Adam, which stabilizes the optimization.
We demonstrate our method on various inpainting tasks, outperforming the state of the art. Project website: \href{https://ubc-vision.github.io/sonic/}{https://ubc-vision.github.io/sonic/}
\end{abstract}
\section{Introduction}
Denoising diffusion models~\cite{ho2020denoising,song2020denoising} and more recently flow models~\cite{lipman2023flow,liu2023rectifiedflow} have become the go-to solution for various inverse problems in Computer Vision~\cite{daras2024surveydiffusionmodelsinverse}.
For example, most modern image inpainters~\cite{Huang_2025,Avrahami_2023} utilize them in one way or another, including those that use pre-trained models with guidance~\cite{liu2025corrfillenhancingfaithfulnessreferencebased,Avrahami_2023,flowchef}, and those that train a conditional inpainting specific model~\cite{zhuang2023task,ju2024brushnet}.
Diffusion and flow methods have also been used for inverse problems in other domains, such as image super-resolution and deblurring~\cite{flowdps, flowchef, flair}.

Among the many, we are similarly drawn to off-the-shelf models~\cite{Avrahami_2023,flowdps,flowchef,flair} for inpainting, due to the versatility afforded by bypassing expensive training~\cite{controlnet,ju2024brushnet} and task-specific augmentations that may not always generalize.\footnote{We show later in experiments that, \eg, \brushnet produces inpainting mask specific results in some cases.}
Unfortunately, the performance of existing training-free methods~\cite{Avrahami_2023,flowdps,flowchef, flair} is not as good as their more specialized counterparts~\cite{ju2024brushnet}; see \cref{fig:teaser}.

Here, we argue that the shortcomings of existing training-free methods, which rely on posterior sampling~\cite{flowdps,flair} or conditioning~\cite{Avrahami_2023,flowchef}, are due to an oversight---optimizing the initial noise sample.
The concept of optimizing the initial noise sample is not new in itself.
Ever since deep generative models have been used to solve inverse problems~\cite{bora2017compressed}, one of the very first approaches has been to treat generative models as transport functions and find the input condition (the initial noise sample for denoising models) that matches the observed data~\cite{bora2017compressed, yeh2017semantic}.

Manipulating the initial noise sample for denoising models, however, requires back-propagating through the denoising chain~\cite{ben2024d}.
This can be costly, both in terms of compute and memory requirements.
Thus, with larger modern models, this back-propagation is deemed impractical, making way for guidance methods~\cite{luo2024readoutguidance, he2023manifoldpreservingguideddiffusion, bansal2023universalguidancediffusionmodels} that change the denoising trajectory during runtime.
Still, recent works~\cite{Li2025ReliableSeeds,Wang2025Seeds} have demonstrated that, already at the initial denoising step, much of the structure of the final denoised outcome is decided.
Indeed, in \cref{fig:init_noise} we show the profound impact that the initial noise sample has on the final inpainted structure.
Thus, the initial noise sample \emph{must} be considered.

\begin{table}[t]
\centering

\begin{minipage}[t]{0.49\linewidth}
    \centering

    {%
    \setlength{\tabcolsep}{1pt}
    \newcommand{\vtextheight}{1.75cm}
    \renewcommand{\arraystretch}{0.85}

    \resizebox{\linewidth}{!}{%
    \begin{tabular}{c@{\hspace{0.2cm}}ccc}
        \includegraphics[width=0.31\linewidth]{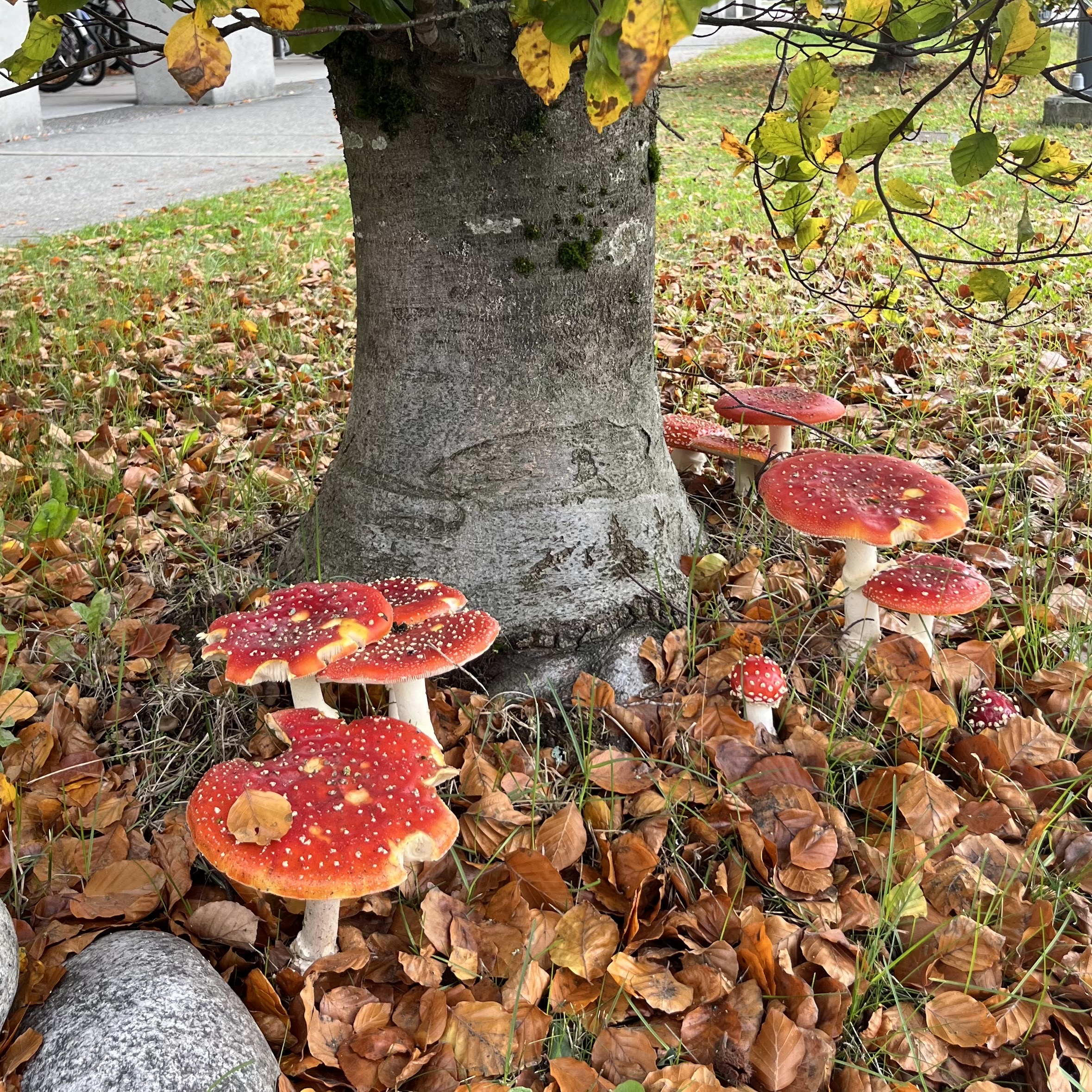} &
        {\rotatebox[origin=l]{90}{\parbox[b]{\vtextheight}{\centering\tiny Seed A}}} &
        \includegraphics[width=0.31\linewidth]{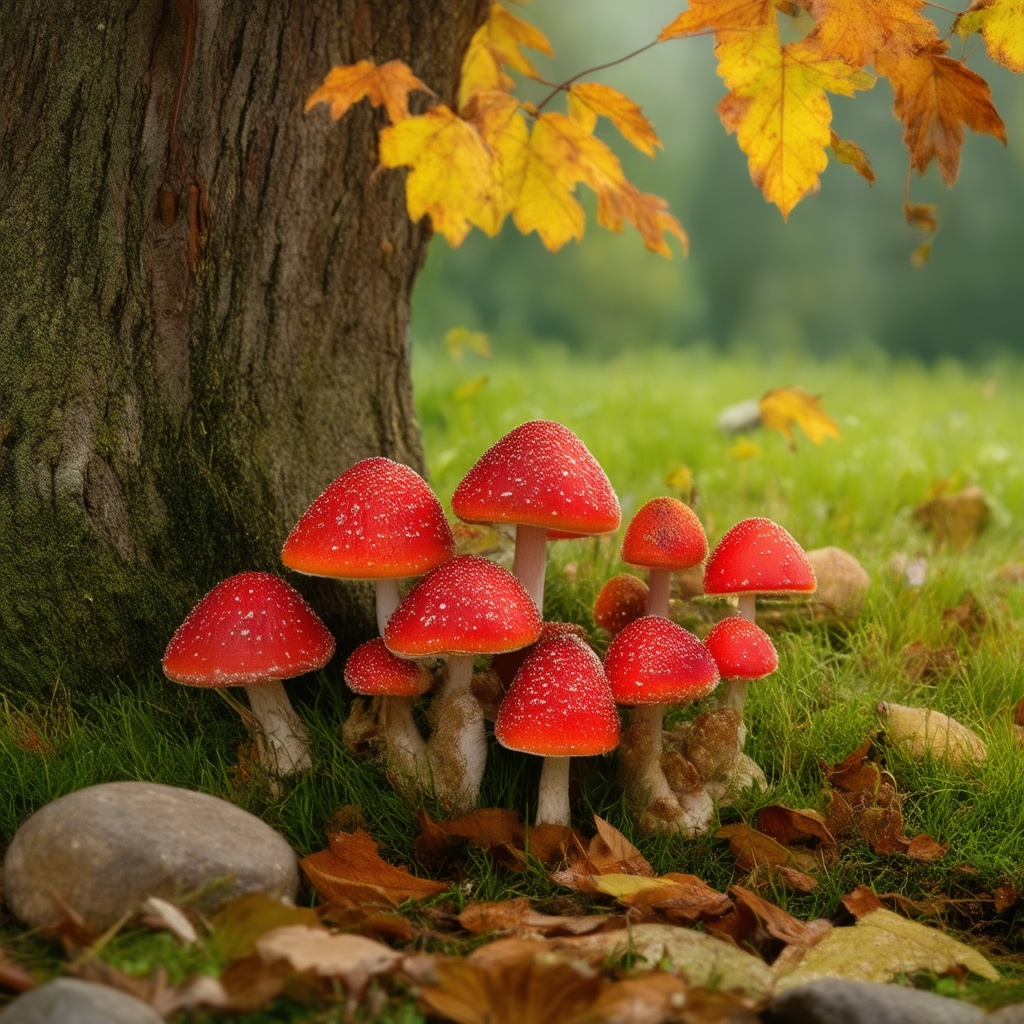} &
        \includegraphics[width=0.31\linewidth]{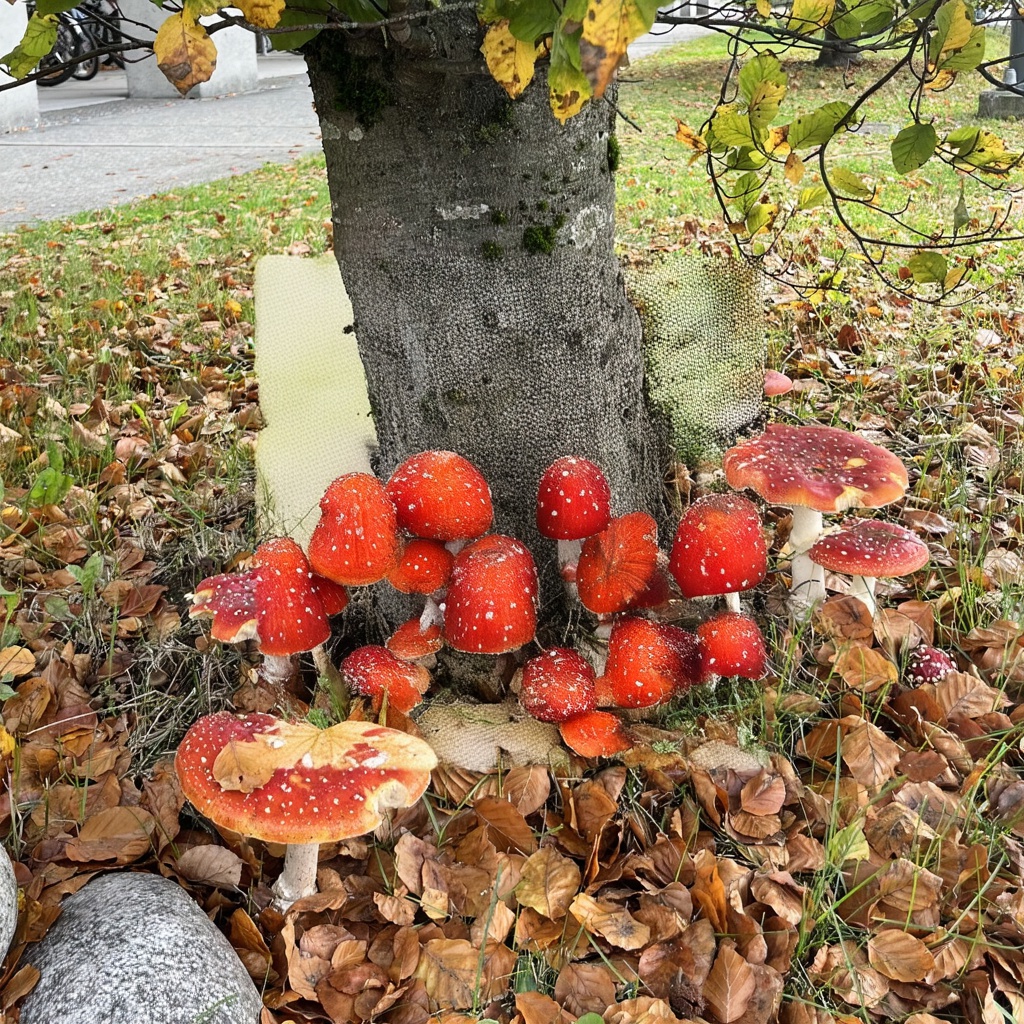} \\
        \includegraphics[width=0.31\linewidth]{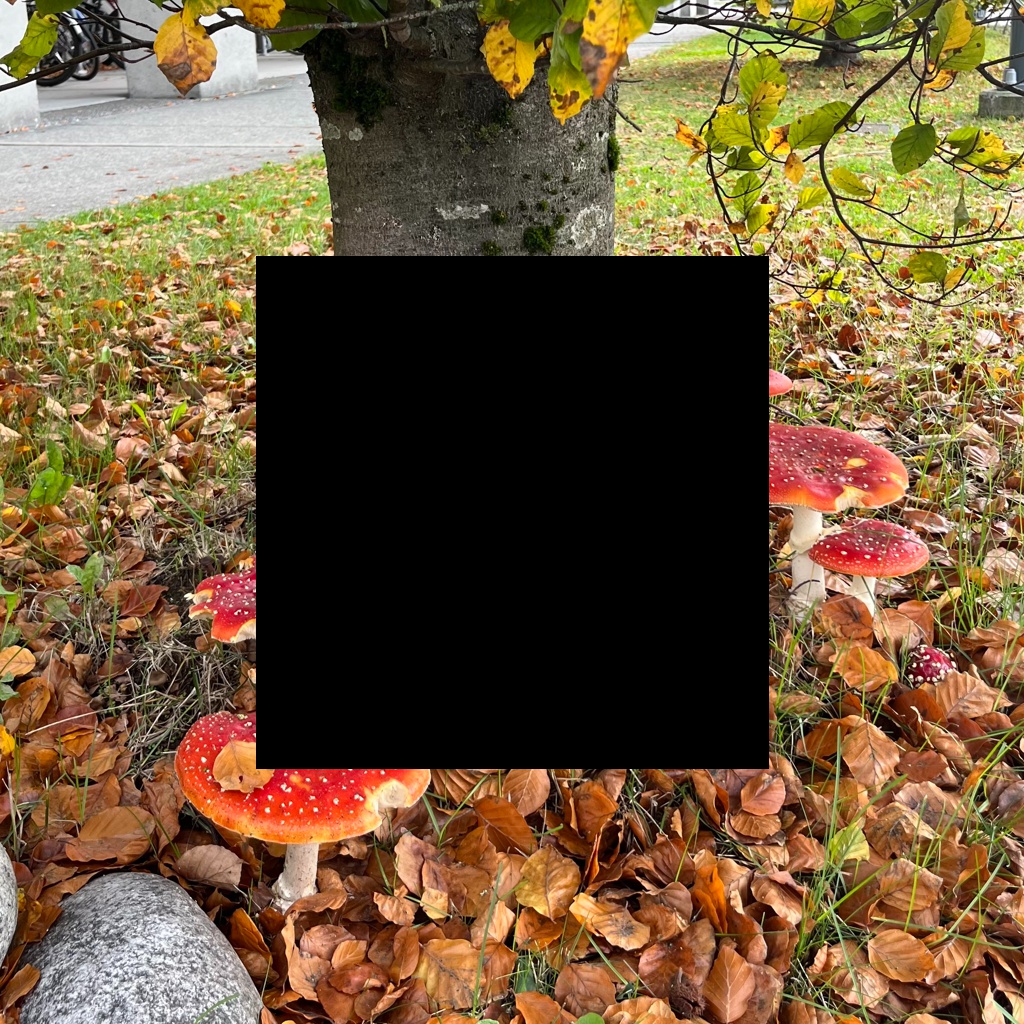} &
        {\rotatebox[origin=l]{90}{\parbox[b]{\vtextheight}{\centering\tiny Seed B}}} &
        \includegraphics[width=0.31\linewidth]{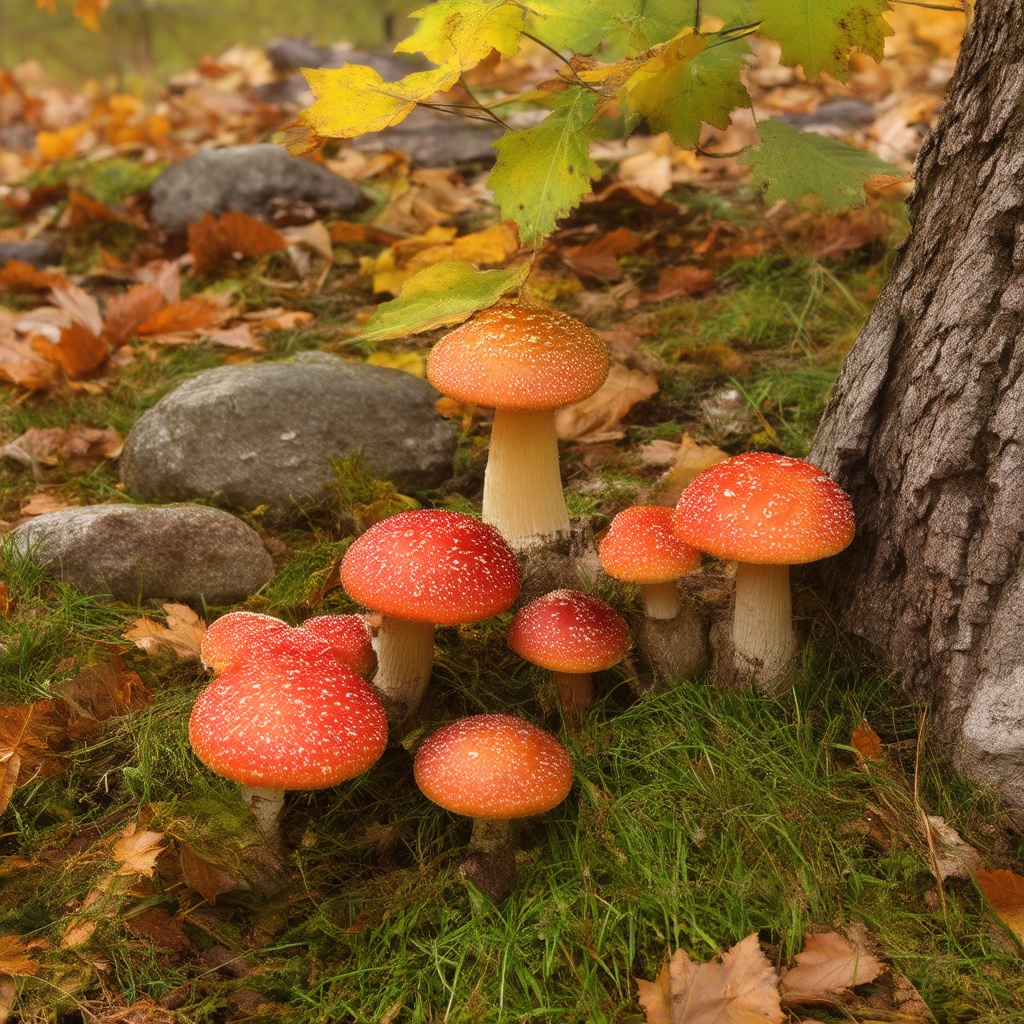} &
        \includegraphics[width=0.31\linewidth]{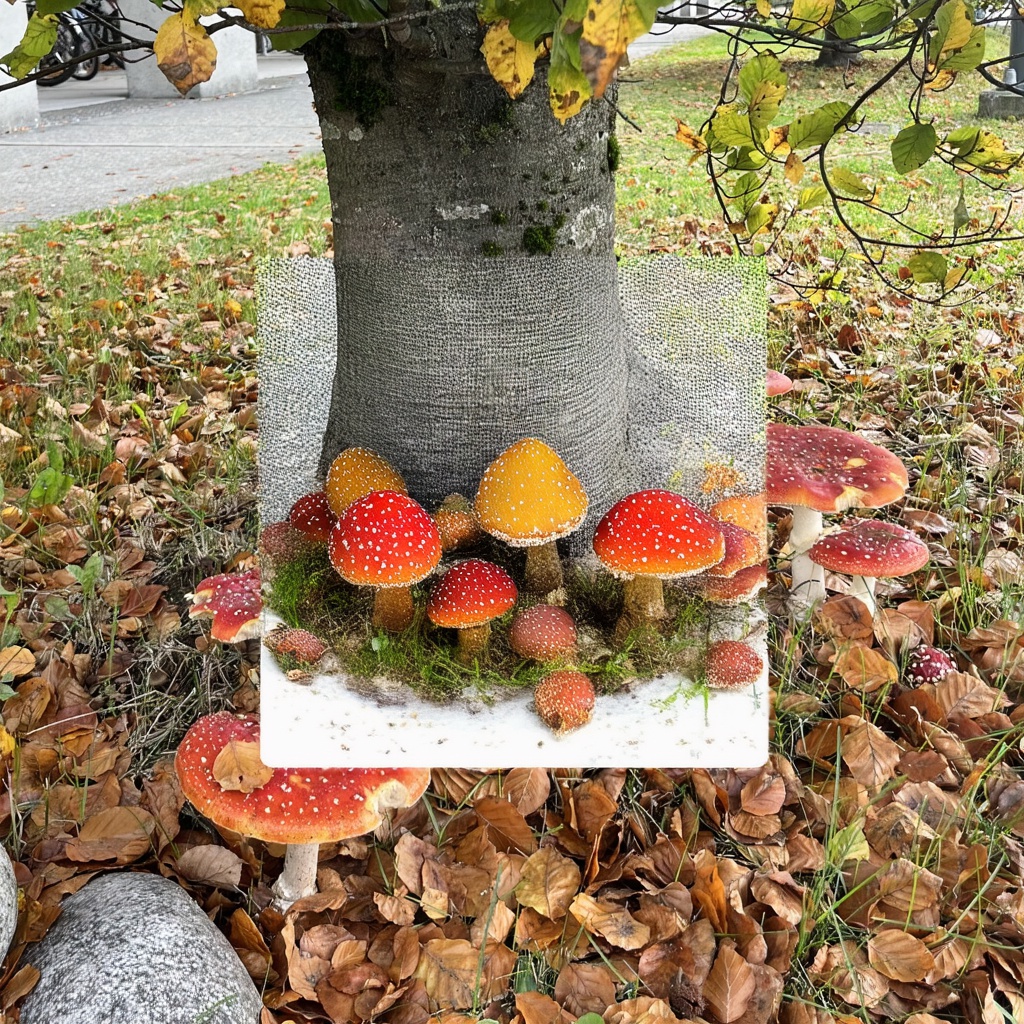} \\
        {\tiny \makecell{Ground truth\\ and mask}} & & {\tiny \makecell{Denoised initial\\ noise sample}} & {\tiny \makecell{Inpainting\\ outcomes}}
    \end{tabular}%
    }%

    }%

    \captionof{figure}{%
        {\bf initial noise sample already determines structure --}
        We show example outcomes of BLD~\cite{Avrahami_2023} with \sdthreefive using different initial noise samples.
        Different initial noise denoise to different scene structure even with the \emph{same} prompt, and \emph{also} the final inpainting outcomes.
        Note how the content within the inpainted region is structurally similar to the denoised initial noise sample.    }
    \label{fig:init_noise}
\end{minipage}
\hfill
\begin{minipage}[t]{0.49\linewidth}
    \centering

    {%
    \newcommand{\vtextheight}{1.75cm}
    \setlength{\tabcolsep}{1.5pt}
    \renewcommand{\arraystretch}{0.85}

    \resizebox{\linewidth}{!}{%
    \begin{tabular}{c ccc}
        {\rotatebox[origin=l]{90}{\parbox[b]{\vtextheight}{\centering\tiny w/o precond.}}} &
        \includegraphics[width=0.31\linewidth]{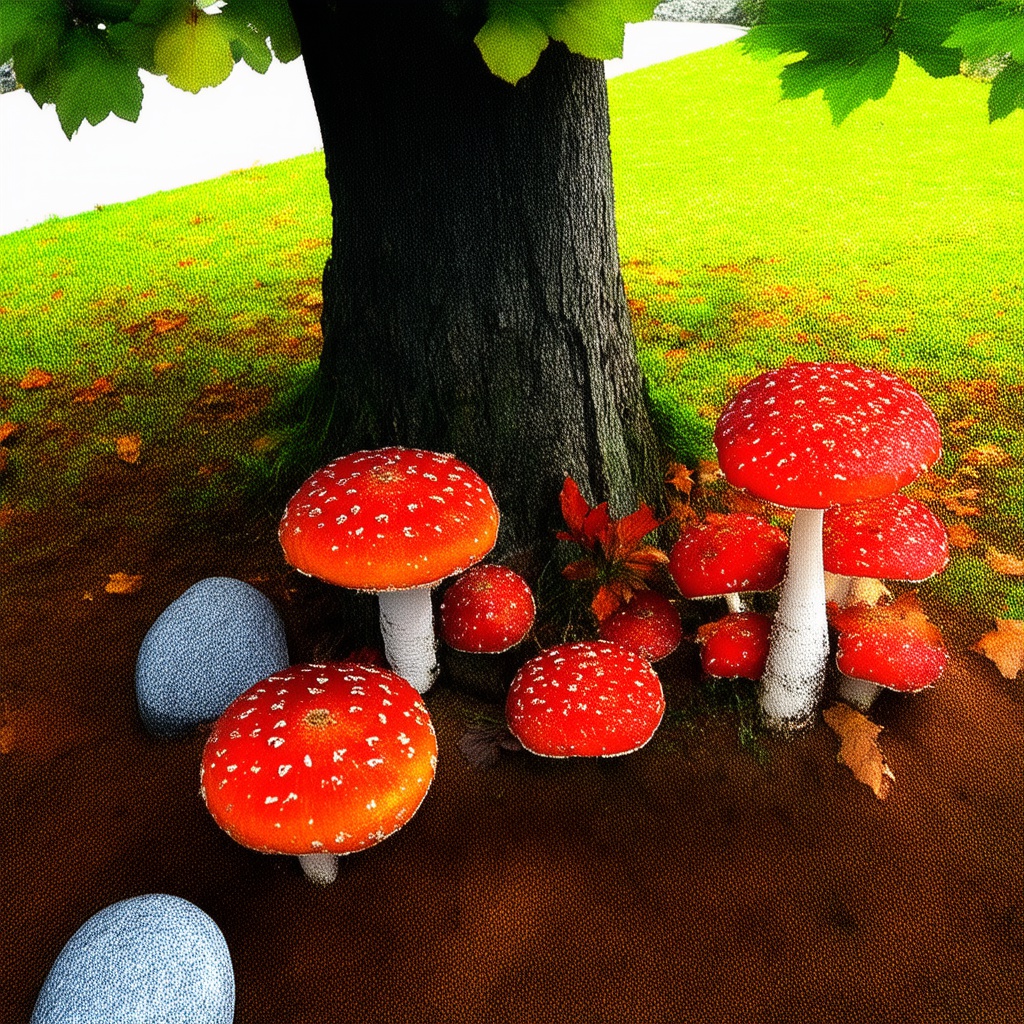} &
        \includegraphics[width=0.31\linewidth]{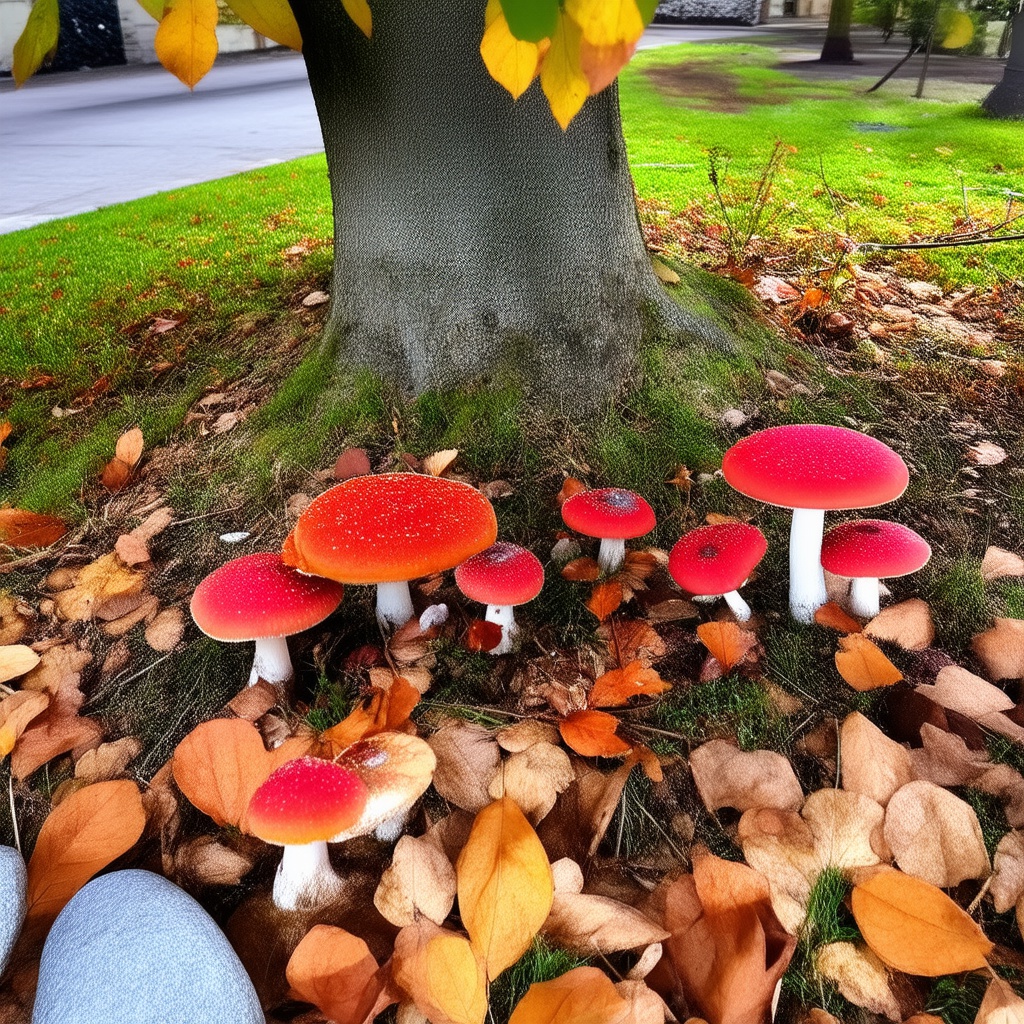} &
        \includegraphics[width=0.31\linewidth]{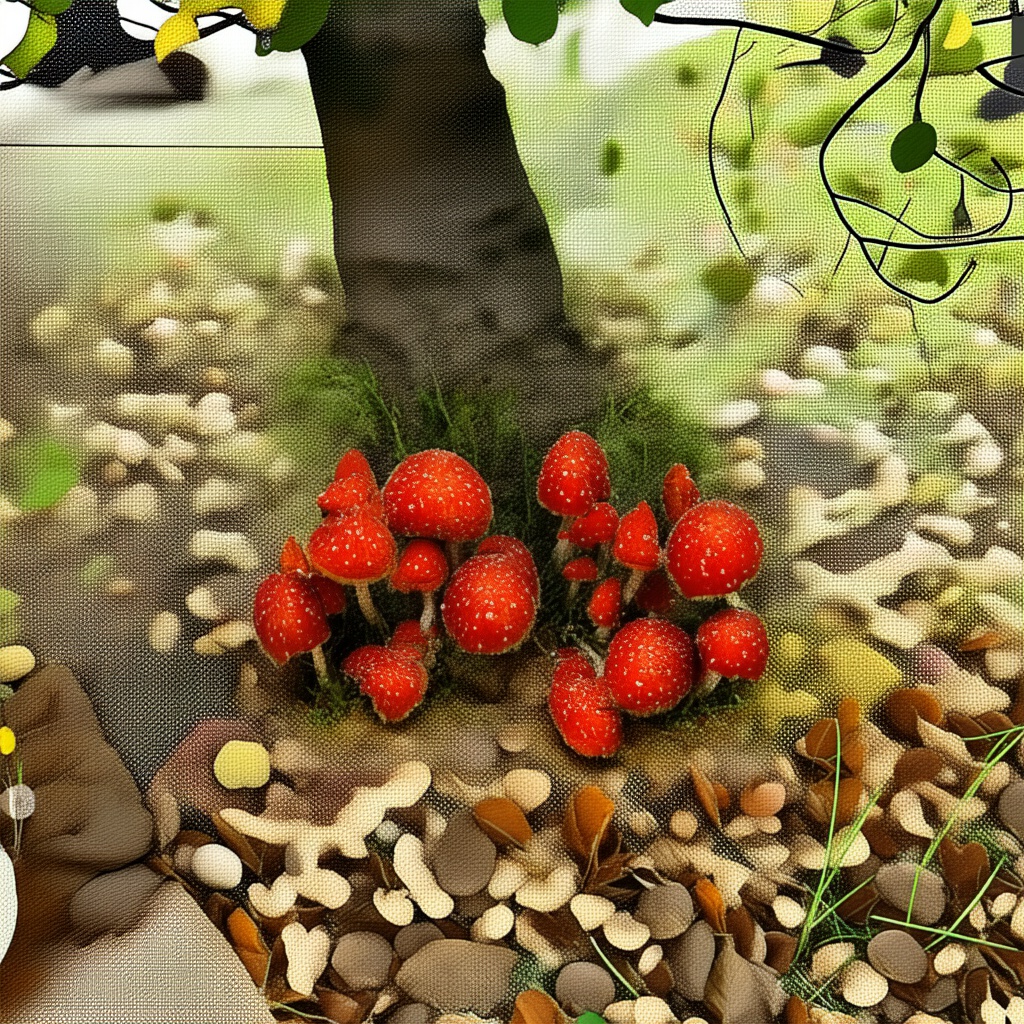} \\
        {\rotatebox[origin=l]{90}{\parbox[b]{\vtextheight}{\centering\tiny \bf w/ precond.}}} &
        \includegraphics[width=0.31\linewidth]{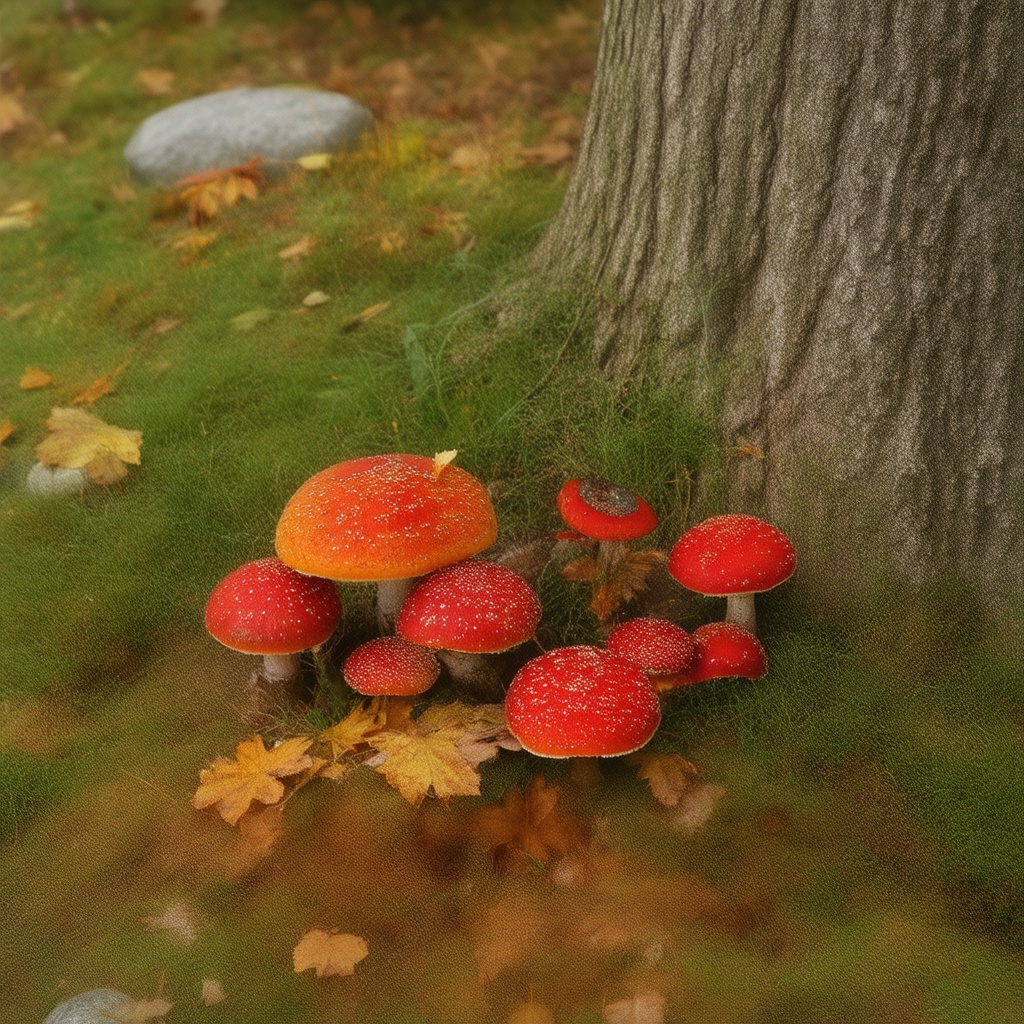} &
        \includegraphics[width=0.31\linewidth]{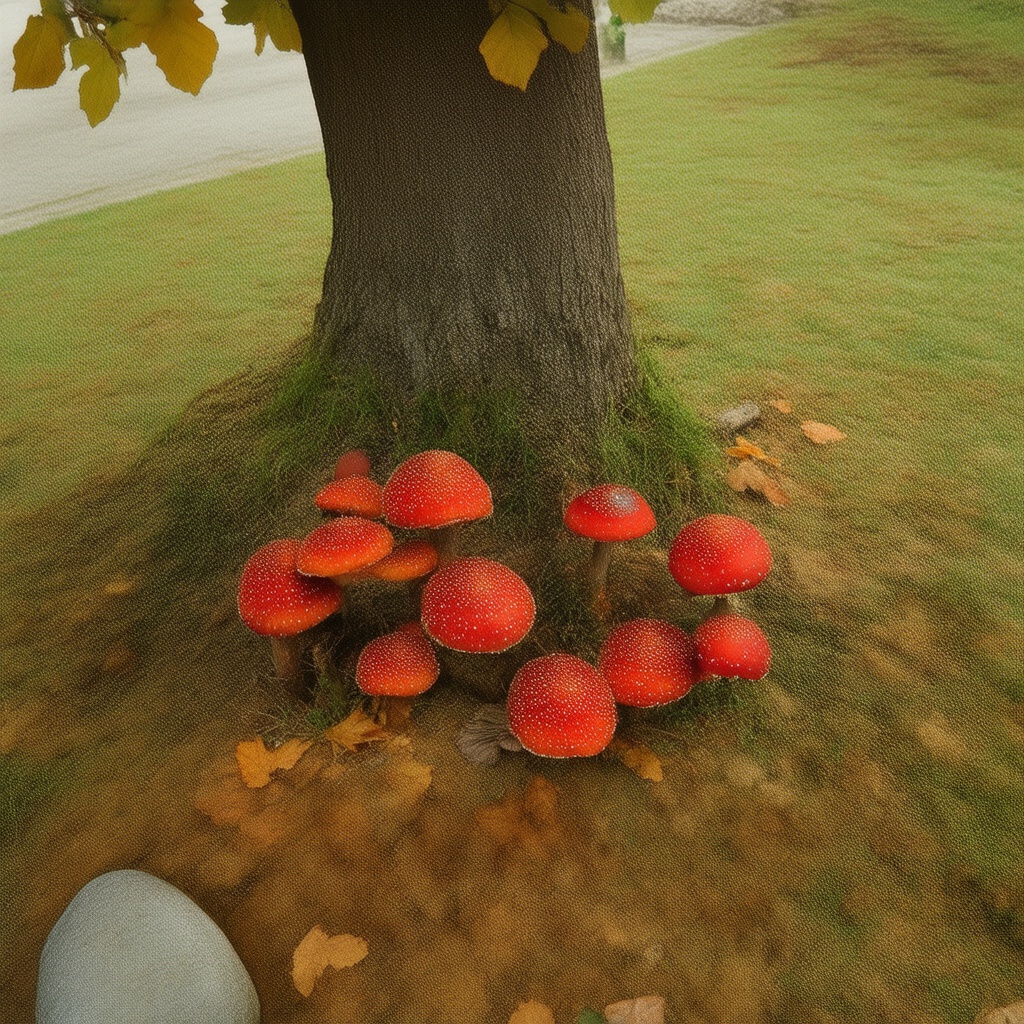} &
        \includegraphics[width=0.31\linewidth]{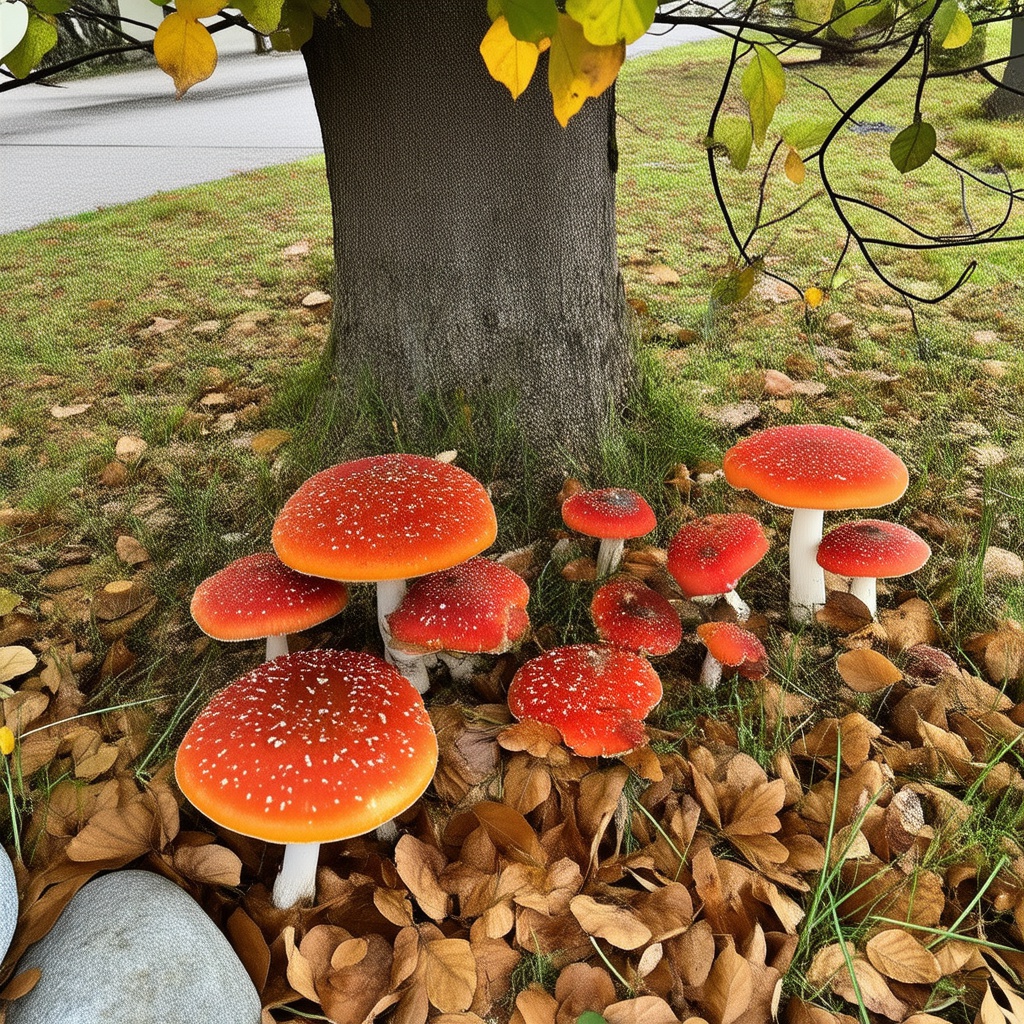} \\
        & {\scriptsize 5 iterations} & {\scriptsize 10 iterations} & {\scriptsize 20 iterations} \\
    \end{tabular}%
    }%
    \vspace{0.25em}
    }%

    \captionof{figure}{%
        {\bf Spectral preconditioning is important --}
        We show examples of how the optimized initial noise sample denoises during the optimization process, when optimized to match the non-masked regions in \cref{fig:init_noise}, starting from \textbf{seed A}.
        Naive optimization also guides toward the desired scene composition, but spectral preconditioning via spectral optimization provides a more stable and robust outcome.
    }
    \label{fig:freq_optimization}
\end{minipage}
\vspace{-2em}
\end{table}

To consider the initial noise sample while circumventing the back-propagation problem, various methods have been proposed.
Some choose the best initial noise sample among many~\cite{Li2025ReliableSeeds, kim2026model}, train a network to generate an ideal noise sample~\cite{ahn2026a, zhou2025golden}, or iteratively refine the noise to improve visual qualities of the output~\cite{wu2024freeinit}. Others formulate the problem as posterior sampling~\cite{flair, flowchef, flowdps, rout2023solvinglinearinverseproblems,Corneanu_2024_WACV, moufad2026efficient}, including a \emph{concurrent} work that drops the model Jacobian~\cite{Ronai2025FlowOpt} and performs inversion, starting from a point generated by another inversion method \cite{jiao2025uniedit}.%

In this work, we propose an inpainting method that \emph{optimizes} the initial noise sample with respect to the final generation outcome, based on two key ideas that allow \emph{practical} and \emph{stable} optimization.

\paragraph{Practical optimization through linearization.} To avoid back-propagating through the denoising chain, we propose to \emph{linearize} the entire denoising (flow) trajectory. 
A naive back-propagation through the denoising chain quickly becomes impractical due to memory and compute requirements.
Instead, we approximate the entire denoising trajectory as a linear path, allowing us to optimize the initial noise sample \emph{without back-propagating through the denoiser}.

\paragraph{Stable optimization through spectral preconditioning.}
Our second contribution is the spectral preconditioning of this optimization that allows for stable convergence.
Various works on spectral and time-based guidance scaling have hinted at this~\cite{yu2023freedomtrainingfreeenergyguidedconditional, sadat2025guidancefrequencydomainenables, Gao_2025_ICCV}---that different frequencies in the latent space have different preferences for the rate of change.
We thus precondition the optimization so that, effectively, each frequency can converge at its preferred pace.
Implementing this can be effectively done by moving the optimization into the spectral domain, together with the Adam~\cite{kingma2017adammethodstochasticoptimization} optimizer that would divide each parameter (now each frequency) by their second moments. This produces significantly more stable optimization and convergence; see \cref{fig:freq_optimization}.

Implementation of this idea, however, requires care.
We find that a naive implementation can lead to corruption of the initial noise sample, moving it away from the manifold of acceptable samples.
To prevent this, we constrain gradient updates to observed data points only, \ie, the unmasked pixels in inpainting.
We further find that the robustness of our method allows it to operate purely in latent space with simple nearest-neighbor fills for the masked regions when encoding masked images to the latent space.

To demonstrate the effectiveness of our method, we evaluate our method on three standard inpainting datasets: \ffhq, \divtwok, and \brushbench, each with different types of masks.
On all three datasets, we outperform the state of the art in SSIM~\cite{wang2004ssim} and LPIPS~\cite{zhang2018unreasonableeffectivenessdeepfeatures} perceptual metrics, and Fr\'echet Inception Distance (FID)~\cite{NIPS2017_8a1d6947}. 
Our method is also often preferable in human preference scores such as HPS v2~\cite{wu2023humanpreferencescorev2} in terms of aesthetics.

\section{Related Work}
While various methods have been proposed to inpaint images, recent works have focused on utilizing the powerful generative prior of diffusion models~\cite{wang2025towards, xie2025turbofill, kim2024radregionawarediffusionmodels,manukyan2024hdpainterhighresolutionpromptfaithfultextguided, xie2022smartbrushtextshapeguided, yang2022paint, Yang_2023, zhang2023coherentimageinpaintingusing}.
We provide a brief review of works that utilize diffusion models, both training-free (ones that utilize generic models) and those that are trained or fine-tuned to be specialized for inpainting; for a more comprehensive review, we refer the reader to \cite{Huang_2025}. 
We also review recent works that focus on the importance of the initial noise sample and its effect on the denoising process, and methods that aim to bypass back-propagation.

\paragraph{Training-free methods.}
One of the key benefits of denoising diffusion models~\cite{ho2020denoising, song2020denoising} is their ability to be `guided' by additional conditioning information ~\cite{ho2021classifierfree,bansal2023universalguidancediffusionmodels,he2023manifoldpreservingguideddiffusion}. The robustness of diffusion models under different manipulations naturally enables their use for inpainting.
Some methods~\cite{lugmayr2022repaint,Avrahami_2023} \emph{blend} the noise within the masked region with the unmasked ground-truth noised image to condition the generation process.
Others extend the Markov chain settings to be conditional~\cite{kawar2022denoising} or perform range-null space decomposition~\cite{wang2023zeroshot}.
Some more recent methods aim to sample from the posterior~\cite{flair, flowchef, flowdps, rout2023solvinglinearinverseproblems,Corneanu_2024_WACV, moufad2026efficient}, such that the observed data points can be treated as the likelihood, while the generic diffusion model serves as the prior; other recent works focus on manipulating the geometry of the sampling trajectory~\cite{shamsolmoali2025missing}.

A notable recent method along these lines is FLAIR~\cite{flair}, which formulates a variational method for posterior sampling.
Interestingly, in their work, the final algorithm involves keeping track of \emph{both} the final reconstruction and the \emph{initial noise}. Another very recent work, DiNG~\cite{moufad2026efficient}, performs guidance with closed-form solution for the posterior, evaluated through auxiliary random samples.

While these models all have shown improved performance beyond the original latent blending work~\cite{lugmayr2022repaint}, they still suffer from the same problem---their inpainting results often have an image structure that is inconsistent with the target image; see~\cref{fig:init_noise}.
Methods that manipulate the initial noise sample~\cite{flair,moufad2026efficient} suffer less from this, but as we show later in \cref{sec:results} with \flair, the effect is still present.
Thus, a proper intervention on the initial noise sample is important.

\paragraph{Importance of initial noise sample.}
Recent works~\cite{ahn2026a, Wang2025Seeds, lyu2025diff, Li2025ReliableSeeds, zhou2025golden} have revealed how much influence the initial noise has on the final generation.
Li \etal~\cite{Li2025ReliableSeeds} demonstrate that the initial seed largely determines the general layout of the generated image, and mine seeds that are more likely to produce better generations.
Similarly, Ahn \etal~\cite{ahn2026a} train a neural network to modify the initial noise sample to generate those that lead to better generations. Lyu \etal~\cite{lyu2025diff} generate a semantically meaningful initial seed with an automatic refinement pipeline. Very recently, an in-depth study~\cite{Wang2025Seeds} using Principal Component Analysis (PCA) revealed that the initial noise can reliably predict image layout.
All of these works hint at the same thing---that the initial noise sample must be considered if we are to match the structure of the image during inpainting.

In theory, this initial noise sample can be directly optimized to achieve a certain goal, which conventional methods have attempted with Generative Adversarial Nets~\cite{bora2017compressed, yeh2017semantic}, denoising diffusion models~\cite{Wallace_2023_ICCV, karunratanakul2024optimizing, tang2025inferencetime, guo2024gradient}, and flow models~\cite{ben2024d, guo2024initno, ReNO2024}. 
These methods, however, require back-propagating through the entire network, 
which is only suitable for smaller models or single-step generators.

\paragraph{Circumventing back-propagation.} 
With the back-propagation being costly, common workarounds include using methods such as the Straight-Through Estimator~\cite{bengio2013estimating}, adopted by many zero-order optimization works~\cite{tao2017zero, milanfarblackbox}. 
In the context of inpainting, recent works such as \flowchef and DiNG~\cite{moufad2026efficient} also suggest adopting the straight-through strategy, \ie, dropping the vector-Jacobian product, although they aim towards optimizing the intermediate noise estimates, not the initial noise sample.
A concurrent work~\cite{Ronai2025FlowOpt} also drops the vector-Jacobian and optimizes for the initial noise sample, but as we will show later (\cref{sec:ablation}), this alone does not provide stable optimization---our spectral preconditioning is required.
This may be why their method starts from an inverted noise from a conventional method~\cite{jiao2025uniedit} before beginning the optimization, and requires a carefully tuned learning rate---our method requires none of these.

\paragraph{Trained methods.}
Various methods have been proposed that \emph{train} or \emph{fine-tune} existing models.
Effectively, these models circumvent the initial noise problem by training inpainting models that ignore the structure imposed by the initial noise sample.
Common strategies include providing an inpainting mask and the encoded masked images as additional input and finetuning a pretrained model~\cite{rombach2022high} with randomly masked images~\cite{suvorov2022resolution}.
These can include having inpainting as a part of a multi-task adaptation via fine-tuning and prompt learning~\cite{zhuang2023task}.
Some separate masked-image features and noisy latents to make the task easier to learn~\cite{ju2024brushnet}, some utilize semantics~\cite{chen2024improving}, and others train inpainting adapters~\cite{xie2025turbofill,Corneanu_2024_WACV}. These methods, however, require additional training which can be costly, and as we show empirically later in \cref{sec:results}, can fail to generalize for mask types not seen during training.

\begin{figure*}[t]
    \centering
\includegraphics[width=1.0\linewidth]{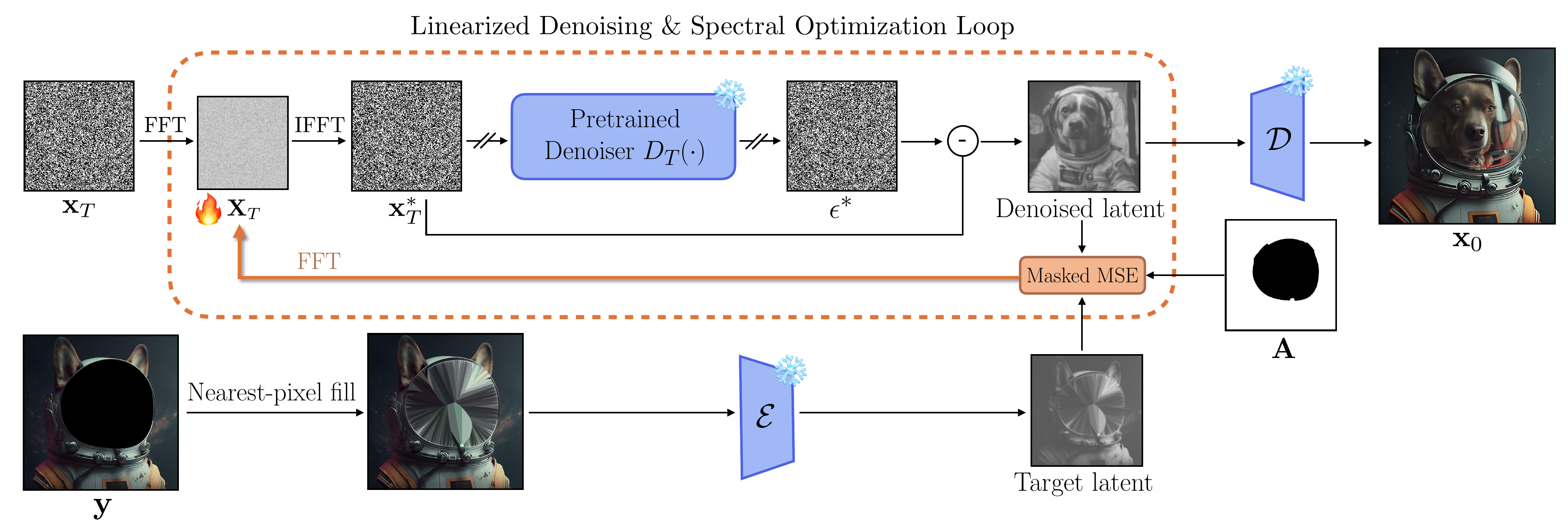}
    \vspace{-1.5em}
    \caption{
    {\bf Method overview -- } 
    We optimize the initial noise sample in the spectral domain $\bX_T$, starting from a random noise $\bx_T$, such that our denoised latent matches the masked observation $\by$ in the latent space.
    To allow partial observations to be encoded, we use nearest-pixel filling before passing it into the encoder.
    We then compute the masked mean square error in the latent space, comparing it with a \emph{fully denoised} latent and update $\bX_T$ accordingly.
    Importantly, we linearize the entire $T$ step denoising process, essentially disconnecting the gradient flow passing through it.
    This allows us to optimize the initial noise sample $\bX_T$ \emph{without back-propagating through the denoiser}.
    }
    \label{fig:overview}
    \vspace{-2em}
\end{figure*}

\section{Method}
\label{sec:method}

\subsection{Overview}
An overview of our method is provided in \cref{fig:overview}. 
Our method focuses solely on finding the initial noise sample that best fits a partial (masked) observation.
Our key innovation is a linearization strategy that allows optimization of the initial noise \emph{without back-propagating through the denoiser}.
We first formalize the problem, then explain our linear approximation.
We then introduce our spectral optimization, as well as subtle yet important implementation details.

\paragraph{Formalization.}
Let $\by$ be a corrupted (masked) image and $\bA$ be the corruption operator (mask), often denoted as a matrix in a slight abuse of notation.
Let $\bx_T$ be the initial noise of a $T$-step denoising process, such that we can write the denoising process as $\bx_0 = \denoise_T(\bx_T)$, where $\denoise_T$ is the $T$-step denoiser (\eg, a flow model~\cite{esser2024}) and $\bx_0$ is the fully denoised image.
We aim to find the initial noise $\bx_T^*$ such that, when denoised $\denoise_T(\bx_T^*)$, then corrupted, it matches the observed image $\by$.
We formulate this as an optimization problem in a least-squares sense:

\begin{equation}
    \label{eq:formalization}
\bx_T^* = \arg\min_{\bx_T} \; \big\| \by - \bA \denoise_T(\bx_T) \big\|_2^2
\end{equation}
In this work, we focus on inpainting, thus $\bA$ is simply a binary mask. Note that, contrary to other methods, we do not alter the denoising process itself, but \emph{only the initial noise $\bx_T$}.
While a noise that exactly satisfies $\by = \bA \denoise_T(\bx_T^*)$ may not exist, our method still provides an \emph{enhanced} starting point for \emph{any} denoising process. 

Naively optimizing $\bx_T$ would mean back-propagating through $\denoise_T(\cdot)$, an expensive iterative process requiring unrolling the entire sequence, costly both in memory and compute---in fact, with an NVidia RTX 5090, back-propagating through more than one denoising step for \sdthreefive is not feasible.
We thus propose a linear approximation to circumvent this issue.

\subsection{Optimizing without heavy back-propagation}
\label{sec:linear}

\paragraph{Denoising trajectory linearization.}
We approximate the denoising trajectory $\bx(t)$, with a linear equation of the form:  
\begin{equation}
    \label{eq:linearization}
    \bx(t) \approx \hat{\bx}(t) = \stopgrad{\denoise_T(\bx_T) - \bx_T}(1-\frac{t}{T}) + \bx_T
    ,
\end{equation}
where $t \in [0, T]$ and $\stopgrad{\cdot}$ is the stop-gradient operator, implying that we consider the term $\denoise_T(\bx_T) - \bx_T$ to be constant throughout the denoising process.
This is a reasonable assumption, given that modern flow models~\cite{liu2023rectifiedflow,lipman2023flow} are generally trained to produce linear trajectories.~\footnote{We provide an illustrative example in \supp.}

\paragraph{Optimization objective in the spatial domain.}
Given this linearization, we can now replace $\denoise_T(\bx_T)$ in \cref{eq:formalization} with $\hat{\bx}(0)$ from \cref{eq:linearization}, and formulate a loss that optimizes for $\bx_T^*$:
\begin{equation}
    \label{eq:spatial_loss}
    \loss{linear}=\left\| 
    \by - \bA \left(\stopgrad{\denoise_T(\bx_T) - \bx_T} + \bx_T\right) 
    \right\|_2^2
    .
\end{equation}
Note that \cref{eq:spatial_loss} is differentiable with respect to $\bx_T$, \emph{without} any need for back-propagation through $\denoise_T(\cdot)$.
\cref{eq:spatial_loss} is, in fact, the straight-through estimator~\cite{bengio2013estimating} that drops Jacobians, and a concurrent work~\cite{Ronai2025FlowOpt} that directly drops Jacobians also arrives at \cref{eq:spatial_loss}.
Direct application, however, as we show in \cref{fig:freq_optimization} and in \cref{sec:ablation}, does not produce stable optimization without the spectral optimization introduced next in \cref{sec:spectral}.
The concurrent work~\cite{Ronai2025FlowOpt} opts for a carefully tuned learning rate, which is not necessary for our method.

\subsection{Optimizing in the spectral domain}
\label{sec:spectral}

When directly optimizing for $\bx_T^*$ via \cref{eq:spatial_loss} we observe what appear to be regional instabilities in the \emph{spectra} of $\bx_T$---denoised image flickering between optimization steps, showing different levels of detail, \ie, \emph{spatial frequencies} converging at different paces. 
A recent work~\cite{guo2024initno} that optimizes initial noise without linearization, back-propagating through individual steps, also notes instability and opts for careful control of optimization steps.
A lower learning rate can avoid this instability, but drastically increases compute expense and leads to under-convergence, producing blurry results.
These observations, combined with recent works that treat different frequencies of the latent space differently during guidance~\cite{yu2023freedomtrainingfreeenergyguidedconditional, sadat2025guidancefrequencydomainenables, Gao_2025_ICCV},
motivate us to precondition the optimization based on frequency.

\paragraph{Spectral preconditioning with Adam~\cite{kingma2017adammethodstochasticoptimization}.}
Spectral preconditioning can be achieved by moving the optimization into the spectral domain and using the Adam optimizer~\cite{kingma2017adammethodstochasticoptimization}.
Instead of treating $\bx_T$ as a trainable parameter, we optimize its spectral representation $\bX_T$, defined as the Fourier transform of $\bx_T$, \ie, $\bx_T = \fourier^{-1}(\bX_T)$.
Then we use the Adam optimizer, which divides each dimension---now frequencies in the spectral domain---by their second moment, achieving \emph{spectrally preconditioned} optimization.
This spectral preconditioning allows
more stable convergence behavior when optimizing, as shown in \cref{fig:optimization_sweep}.
While this may seem like a small change, it has a significant impact on the quality of inpainting outcomes---to a degree where spatial optimization often results in inpainting failures.
We empirically ablate this choice in \cref{sec:ablation} and provide further theoretical justification in the \supp.

Thus, our final optimization objective is:
\begin{equation}
    \label{eq:spectral_loss}
    \loss{spectral}=\left\| 
    \by - \bA \left(\stopgrad{\denoise_T(\bx_T) - \fourier^{-1}(\bX_T)} + \fourier^{-1}(\bX_T)\right) 
    \right\|_2^2
    ,
\end{equation}
which we optimize for $\bX_T$ with Adam~\cite{kingma2017adammethodstochasticoptimization}.

\subsection{Implementation details}

\paragraph{Spatially constraining the optimization.}
\label{sec:masking}
A critical component to consider is that our optimized initial noise sample must remain within the manifold of acceptable initial noise samples.
A naive implementation of our method, even with spectral optimization, can still lead to corruption because of the continuous frequency formulation of the optimization target---where a single frequency component affects all pixels.
Thus, the latents corresponding to \emph{unobserved} regions, without observations to guide them, can easily diverge away from the Gaussian distribution that the denoising model expects the initial noise to follow.

\begin{figure}[t]
\centering
\begin{minipage}[c]{0.51\linewidth}
    \centering
    \includegraphics[width=0.95\linewidth]{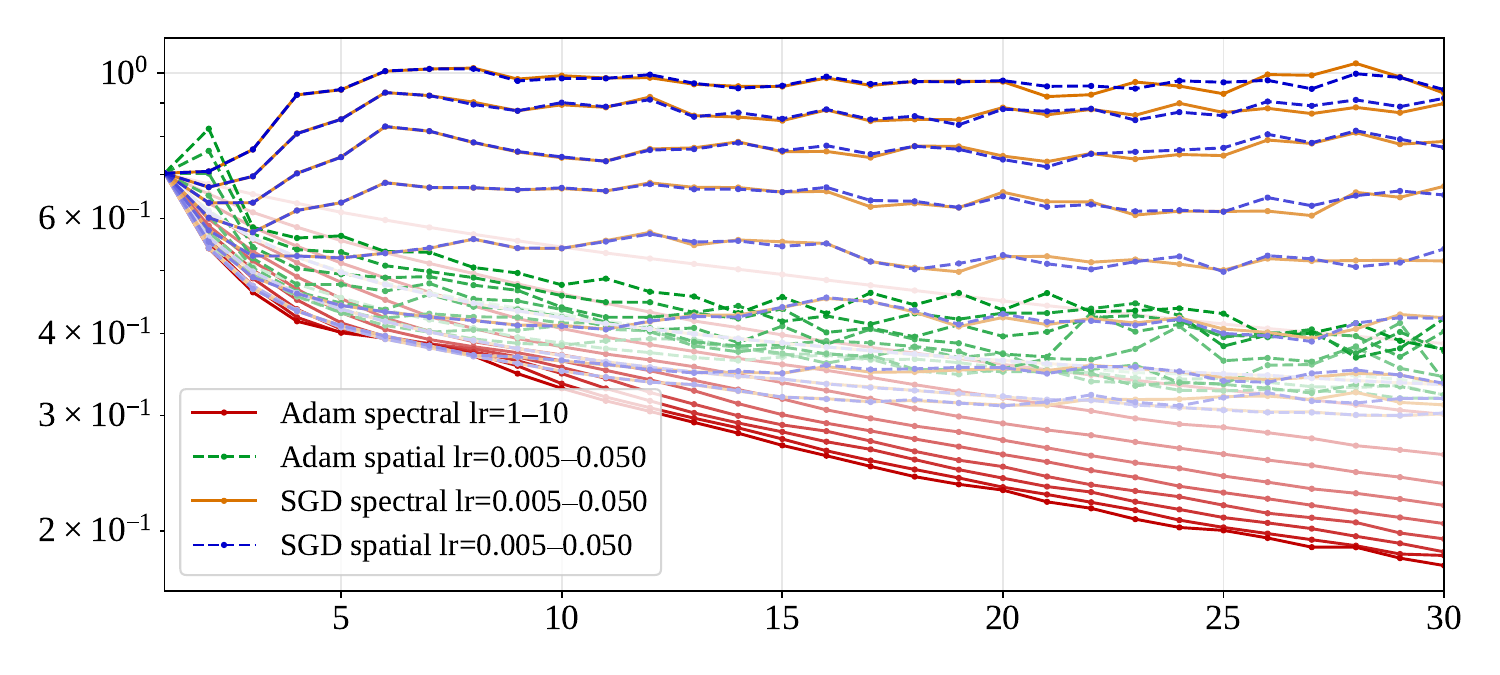}
    \vspace{-1em}
    \caption{%
      {\bf Spectral preconditioning --} 
      We plot the average convergence trajectory in terms of root mean-squared error (RMSE) when inverting 100 randomly generated images starting from a random noise sample (see \cref{sec:ablation} for details).
      Using the Adam optimizer in the spectral domain, \ie, with our spectral conditioning, converges well regardless of the learning rate, while Adam in the spatial domain performs significantly worse. 
      SGD performs the same regardless of the domain, as simply moving to the spectral domain, for SGD, is mathematically identical to being in the spatial domain.
    }
    \label{fig:optimization_sweep}
\end{minipage}
\hfill
\begin{minipage}[c]{0.46\linewidth}
    \centering
    {%
    \newcommand{\vtextheight}{1.75cm}
    \setlength{\tabcolsep}{2pt}
    \renewcommand{\arraystretch}{0.85}
    \resizebox{\linewidth}{!}{%
    \begin{tabular}{c ccc}
        {\rotatebox[origin=l]{90}{\parbox[b]{\vtextheight}{\centering\tiny w/o constraint}}} &
        \includegraphics[width=0.32\linewidth]{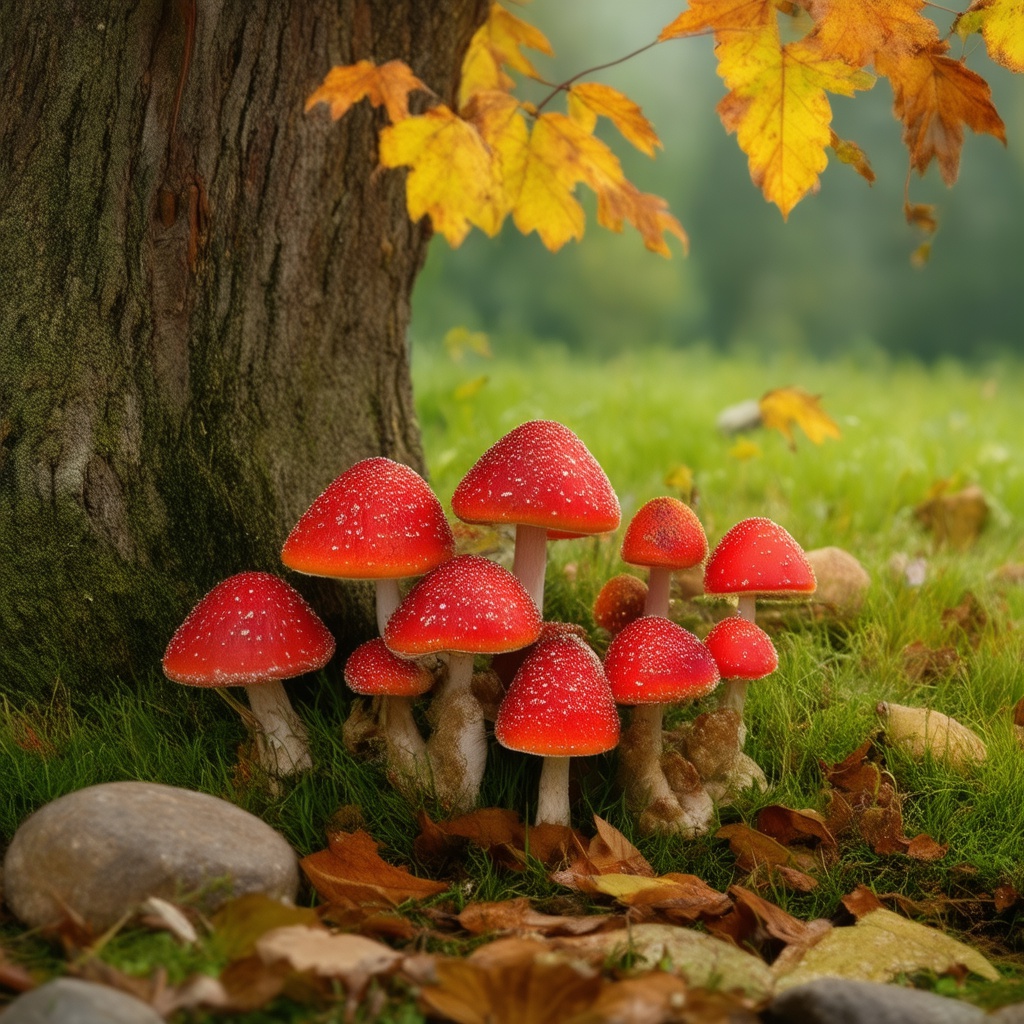} &
        \includegraphics[width=0.32\linewidth]{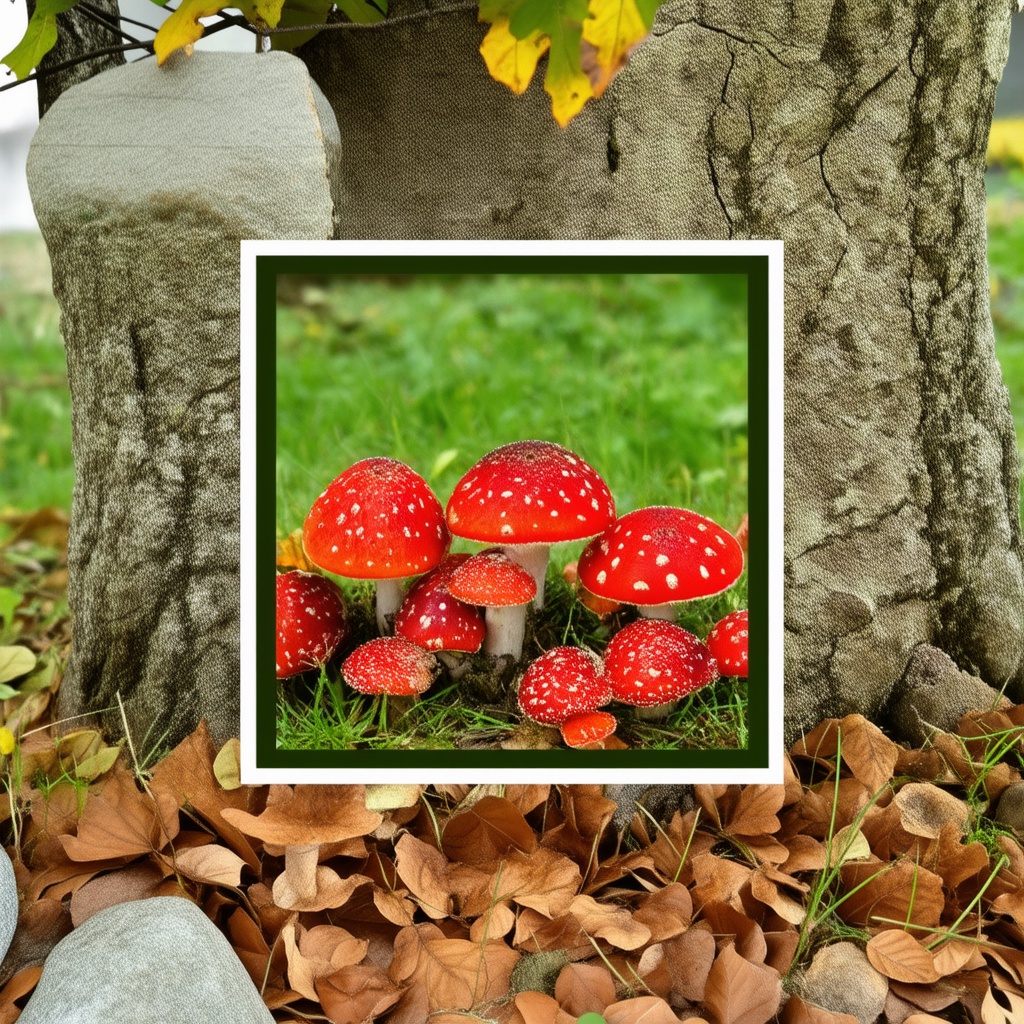} &
        \includegraphics[width=0.32\linewidth]{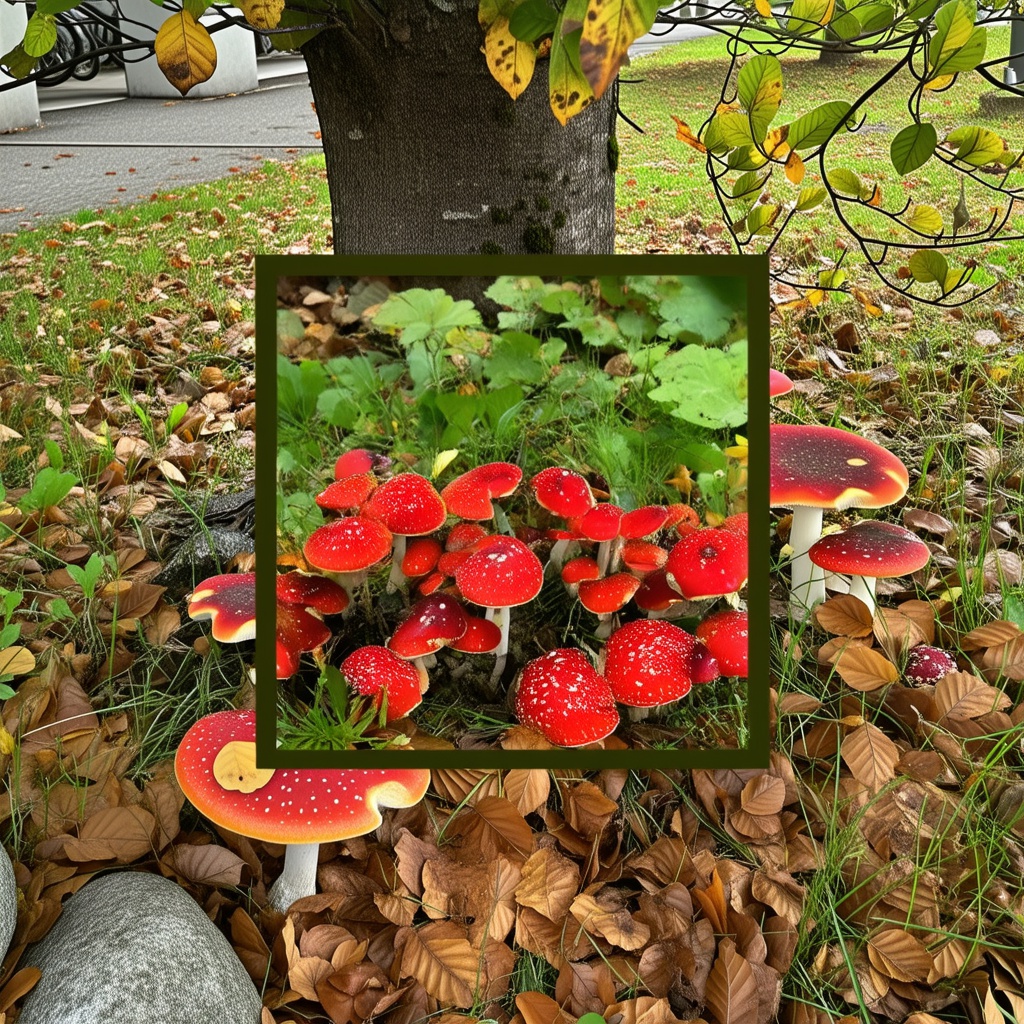} \\
        {\rotatebox[origin=l]{90}{\parbox[b]{\vtextheight}{\centering\tiny w/ constraint}}} &
        \includegraphics[width=0.32\linewidth]{figures/jpg_figs/freq_optimization/spectral_0.jpg} &
        \includegraphics[width=0.32\linewidth]{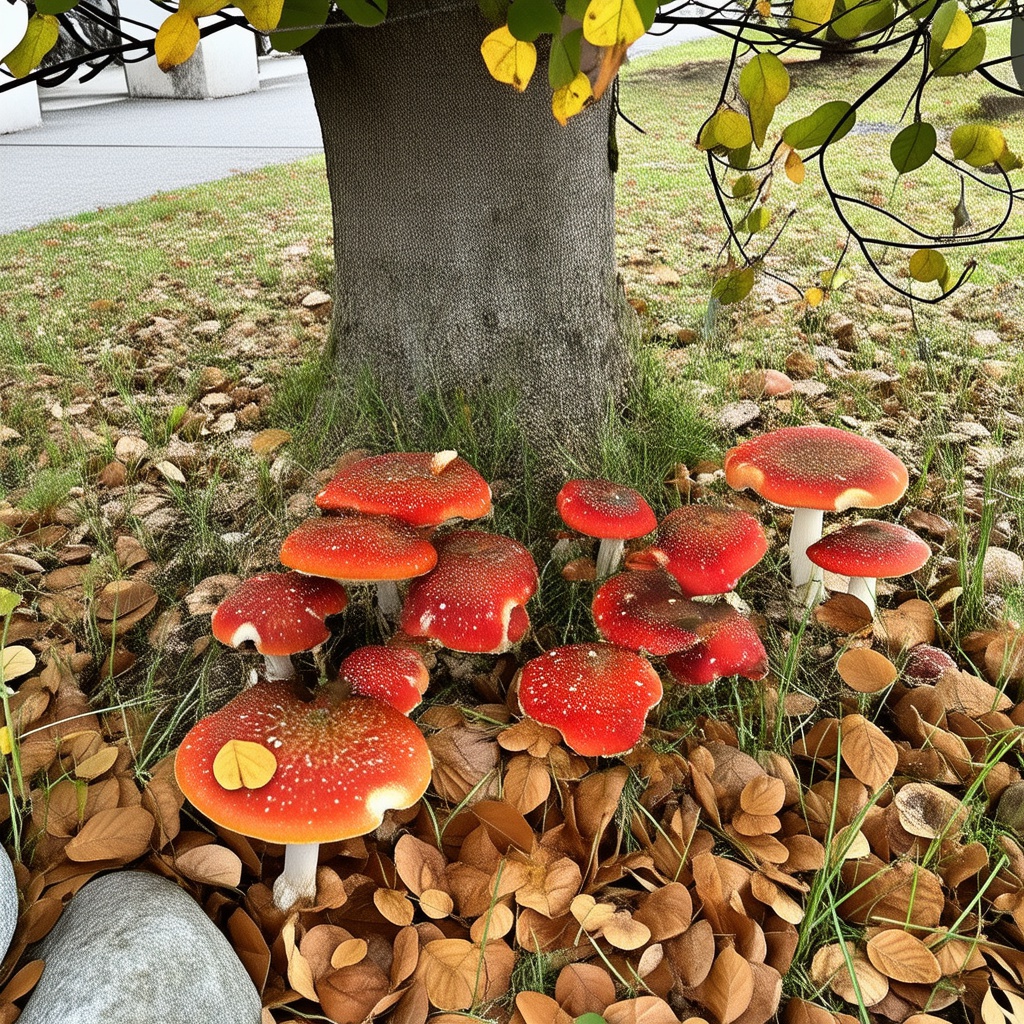} &
        \includegraphics[width=0.32\linewidth]{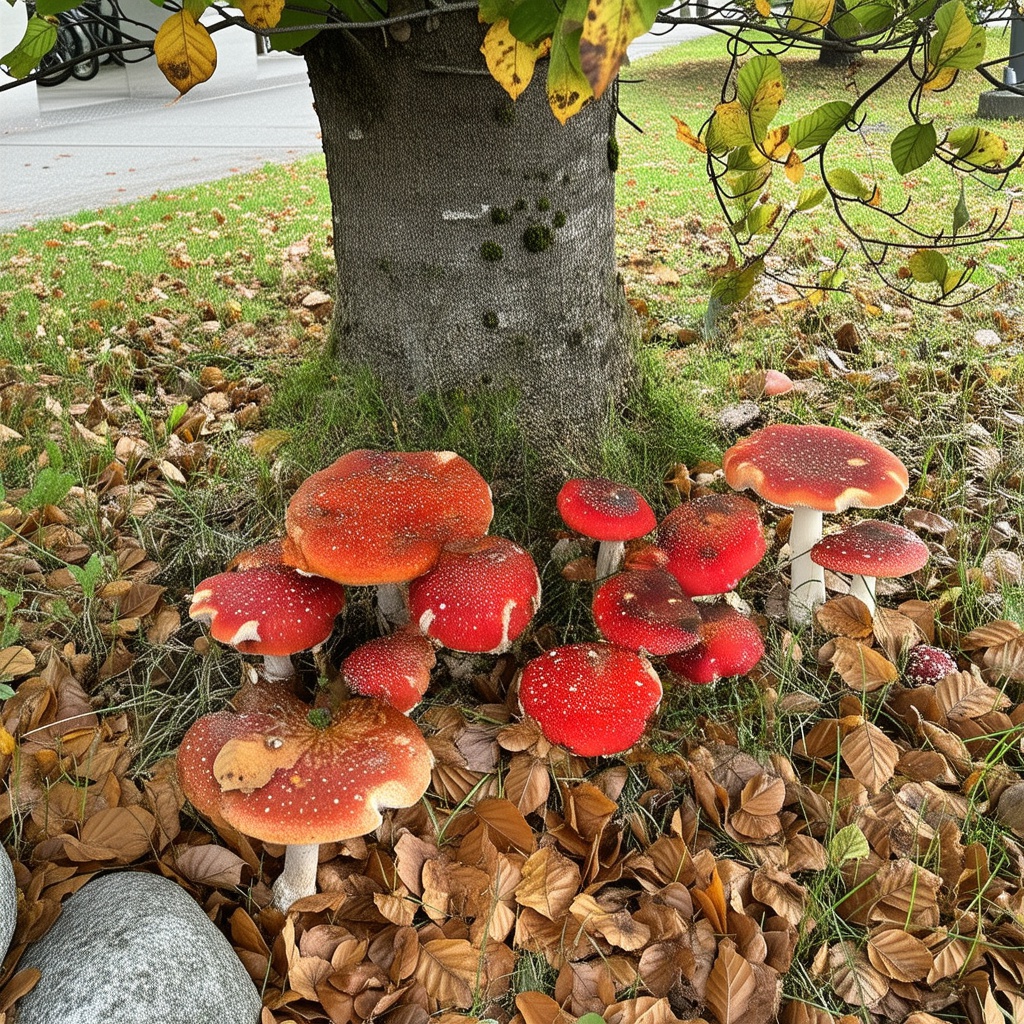} \\
        &
        {\scriptsize 0 iteration} & {\scriptsize 25 iterations} & {\scriptsize 40 iterations}
    \end{tabular}
    }%
    }%
    \caption{%
        {\bf Spatial constraining --}
        We show how the initial noise sample denoises, similarly to \cref{fig:freq_optimization}, but with and without the spatial constraint.
        As shown, without spatial constraining, the optimization diverges from the initial noise sample manifold, causing the visible colour shifts, artifacts, and mask borders as optimization progresses.
    }
    \label{fig:masking}
\end{minipage}
\vspace{-1em}
\end{figure}

We thus opt to \emph{freeze} latents in these regions, as they \emph{already} follow a Gaussian distribution, by masking their gradient updates.
As shown in \cref{fig:masking}, without such treatment, optimized noise may introduce visible artifacts, especially near the border.

\paragraph{Masking in the latent space.}
Another computational bottleneck, aside from back-propagation through the denoiser or the flow estimation, is the Variational Auto-Encoder (VAE) that maps images to the latent space.
Thanks to the robustness of our method, we find that we can safely perform all operations in the latent space, by simply infilling masked region pixels with their nearest-neighbor color; see \cref{fig:overview}.
While this does not look like a proper inpainted image, we empirically find this method sufficient.
We compare this simple infill strategy against the theoretical \emph{upper-bound} of ground-truth pixels in \cref{sec:ablation}.

\paragraph{Other details.}
We implement our method with \sdthreefive, and with \bld as the inpainter starting from our optimized initial noise sample.
To optimize the initial noise sample, we use the Adam~\cite{kingma2017adammethodstochasticoptimization} optimizer with a learning rate of 3.0, and with default parameters $\beta_1=0.9$ and $\beta_2=0.999$.
We use a classifier-free guidance~\cite{ho2021classifierfree} scale of 2.0, following FLAIR~\cite{flair}.
We use an image resolution of $1024 \times 1024$ for all our experiments to match the expected input size of SD~3.5.
We also use $T=20$ for our denoiser.

\section{Results}
\label{sec:results}

We first discuss our experimental setup, then present our results, and finally ablate our design choices.

\subsection{Experimental setup}
\label{sec:experimental_setup}

\vspace{-\customparskip}
\paragraph{Datasets and masks.}
We evaluate our method on three standard datasets, using different types of masks on each.
\begin{itemize}
    \item \textbf{\ffhq}:
    A standard dataset often used for evaluating image inpainting methods, containing 70,000 high-quality face photographs with diverse age, pose, and accessories.
    We follow the protocol in FLAIR~\cite{flair}, using the first 1000 images and applying a large, rectangular mask with a fixed size and position that covers approximately half of the subject’s face.
    \item \textbf{\divtwok}:
    A set of 1,000 2K high-resolution natural images.
    Again, we follow the protocol of \flair by taking the 800 images from the training set, and center-cropping a square region from the image. 
    Then, we apply the same masking pattern to all samples, consisting of six randomly positioned rectangular regions.\footnote{We use the exact same locations as \flair.}
    \item \textbf{\brushbench}:
    600 natural and artificial images with human-annotated masks based on segmentation and captions.
    We use the provided inpaint segmentation masks and prompts.
\end{itemize}

\paragraph{Prompts.}
As diffusion/flow-based inpainting methods require prompts, we systematically generate prompts through a Vision Language Model (VLM)~\cite{claude-sonnet-4-5} for \ffhq and \divtwok.

We provide the masked images and prompt the VLM to infer the contents of the image and describe them in two to three concise sentences. The exact prompt used for the VLM can be found in \supp. We use the generated prompts for \emph{all} methods, including ours.
For \brushbench we simply use the standard dataset-provided prompts.

\paragraph{Baselines.}
We compare our method against the following baselines:
\begin{itemize}
    \item \textbf{\bld}: Adopted by diffusers~\cite{von-platen-etal-2022-diffusers} as the default inpainting pipeline, it is a training-free method that `blends' the denoising process in the masked region with the ground-truth denoising trajectory of the unmasked region.
    \item \textbf{\flowchef}: Improves trajectories and stability for flow/diffusion models when solving inverse problems.
    \item \textbf{\flowdps}: A flow-based posterior sampling method for inverse problems with explicit guidance.
    Following \flair, we implement an inpainting solver for this method.
    \item \textbf{\flair}: A recent training-free inverse problem solver that uses a variational formula involving the update of both initial noise and final estimates.
    \item \textbf{\brushnet}: A trained method that specializes in inpainting.
    We use the Stable Diffusion 1.5~\cite{rombach2022high} checkpoint from the official implementation, and their standard classifier-free guidance scale of 7.5. 
    As BrushNet offers two model subtypes: one trained with random masks, and another with segmentation masks.
    We use the former for \ffhq and \divtwok while using the latter for \brushbench.
\end{itemize}
For fairness, as in \flair, we use the \emph{same} \sdthreefive (`medium' variant) base model for training-free methods (\bld, \flowchef, \flowdps, \flair, and ours).
We set the classifier-free guidance scale to 2.0 for these methods, as in \flair.
Additionally, as the baselines generate varying resolution outputs, we process images at their native resolution (varying from $512 \times 512$ to $768 \times 768$) and resize to $1024 \times 1024$ for evaluation.

For each posterior baseline, we report performance using their suggested number of function evaluations (NFE), and performance at 400 NFE, which is what we use for our method. Note, however, that as our method does not back-propagate through the denoiser, our runtime is comparable to methods with a lower NFE---taking approximately a minute to inpaint one image on a GeForce RTX 5090 GPU, roughly the same time as \flair using 50 NFE. 
We include the wall clock time table in the \supp.

\paragraph{Evaluation metrics.}
To quantify the quality of the inpainting results, we use the standard metrics: Peak Signal-to-Noise Ratio (PSNR), Structural Similarity Index Measure (SSIM) \ssim, Learned Perceptual Image Patch Similarity (LPIPS) \lpips, Fr\'echet Inception Distance (FID) \fid, as well as CLIP score (Contrastive Language–Image Pretraining)~\cite{radford2021clip}.
We also use metrics that are designed to mimic human preference: Image Reward (IR)~\cite{xu2023imagerewardlearningevaluatinghuman}, Human Preference Score v2 (HPS v2)~\cite{wu2023humanpreferencescorev2}, and Aesthetic Score (AS)~\cite{10.5555/3600270.3602103}.

For the task of inpainting, where many different answers can exist, we warn that the PSNR metric should be considered with care---\emph{lower PSNR does not necessarily mean worse performance}, as also discussed in~\cite{flair}.
Instead, all metrics should be considered altogether. 
Still, if a few metrics were to be weighed strongly, we argue that LPIPS and FID are the \emph{most important} for evaluating inpainting quality as they evaluate how similar the inpainted content is to the ground truth via `perceived' similarity.
This tolerates minor differences in exact structure and color, which are acceptable in the context of inpainting.
CLIP score gauges how well the inpainting results match the original content \emph{semantically}, but should also be read carefully, as the inpainted content may be \emph{geometrically} misaligned with the image.
IR, HPS v2, and AS are also imperfect metrics, as they are designed to measure image quality, not inpainting quality. 
We thus further conduct a single-blind two-alternative choice study with N=37 raters, where our method is preferred in 90\% of comparisons. We include detailed results in the \supp.

\begin{figure*}[t]
    \centering
    \newcommand{\imwidth}{0.187\linewidth}
    \setlength{\tabcolsep}{1pt}
    \renewcommand{\arraystretch}{0.85}
    \begin{tabular}{c ccccc}
        {\rotatebox[origin=l]{90}{\parbox[b]{2.2cm}{\centering\tiny \ffhq}}} &
        \includegraphics[width=\imwidth]{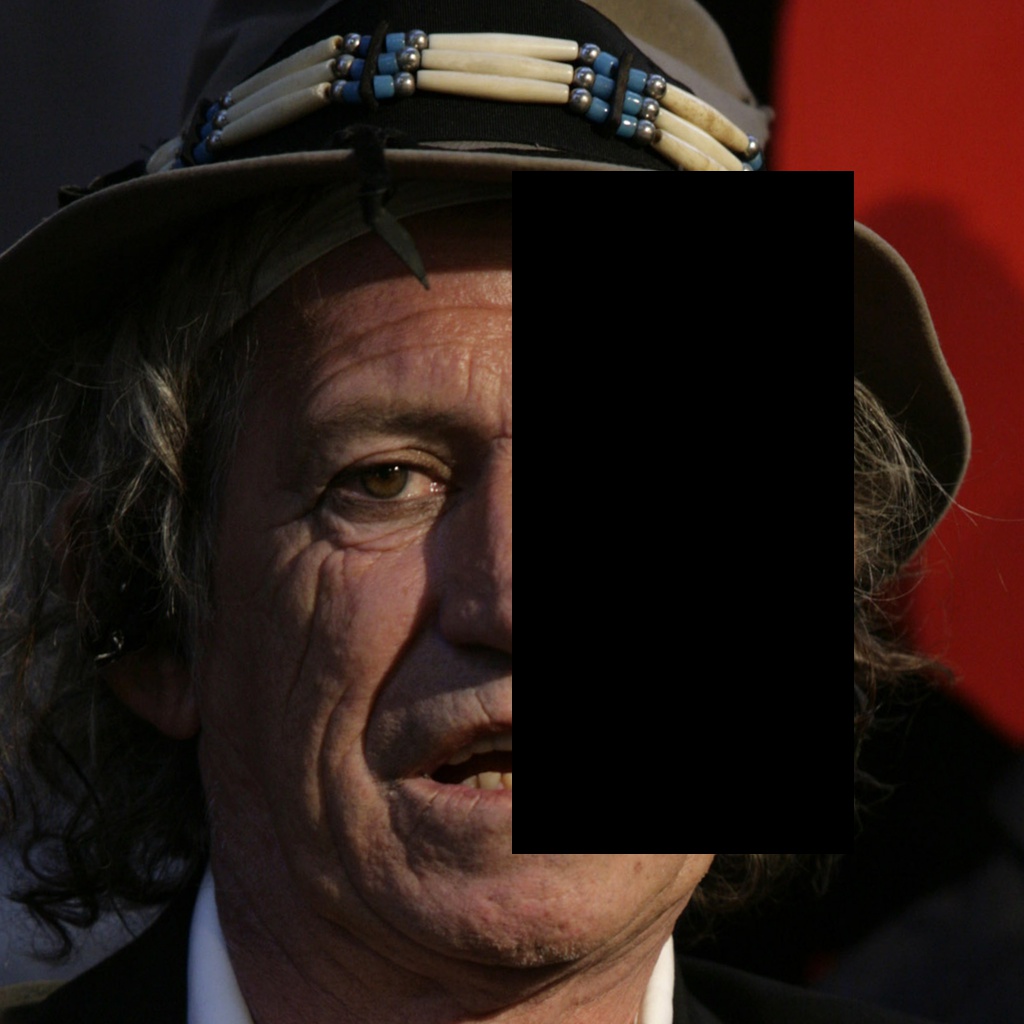} &
        \includegraphics[width=\imwidth]{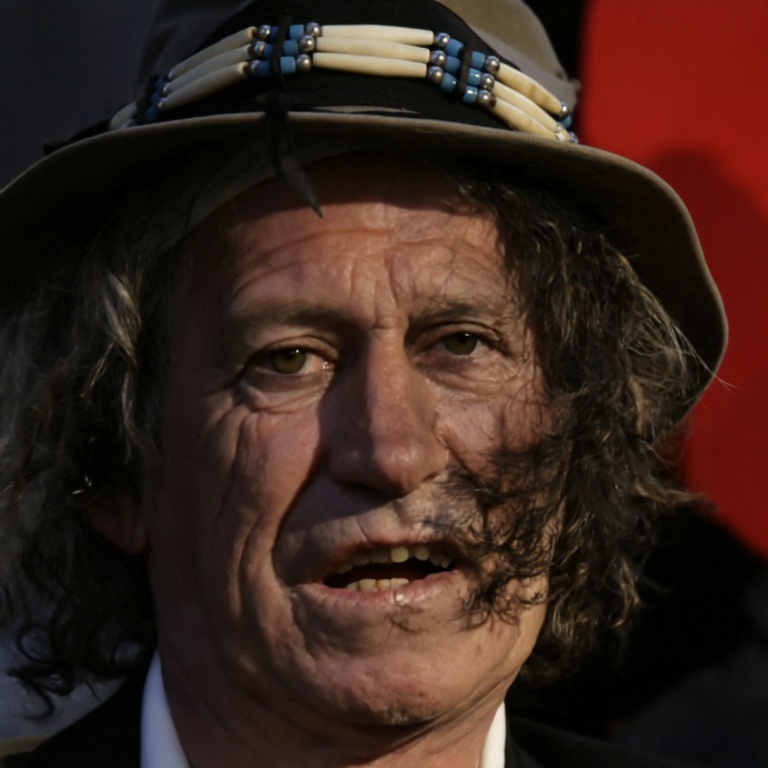} &
        \includegraphics[width=\imwidth]{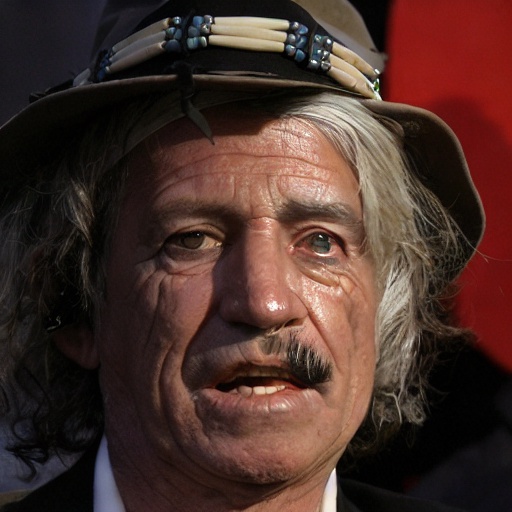} &
        \includegraphics[width=\imwidth]{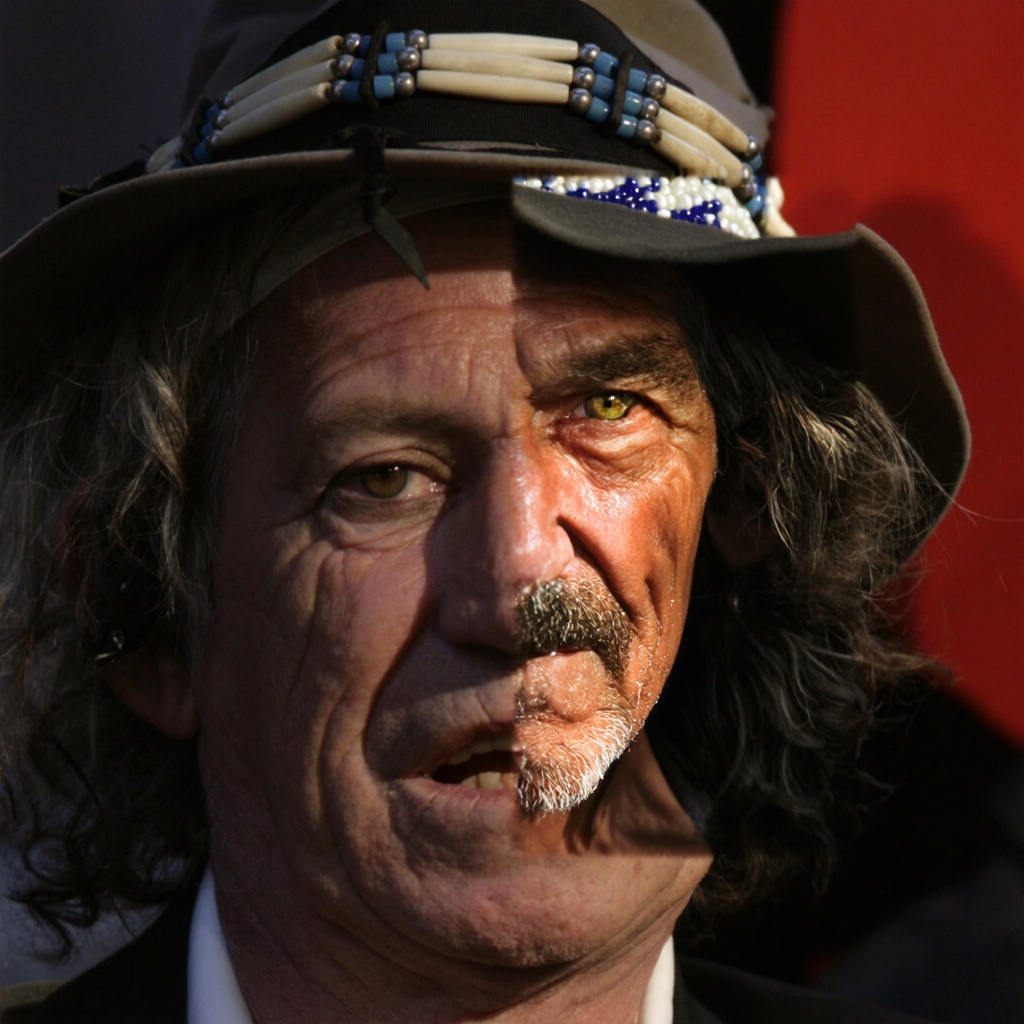} &
        \includegraphics[width=\imwidth]{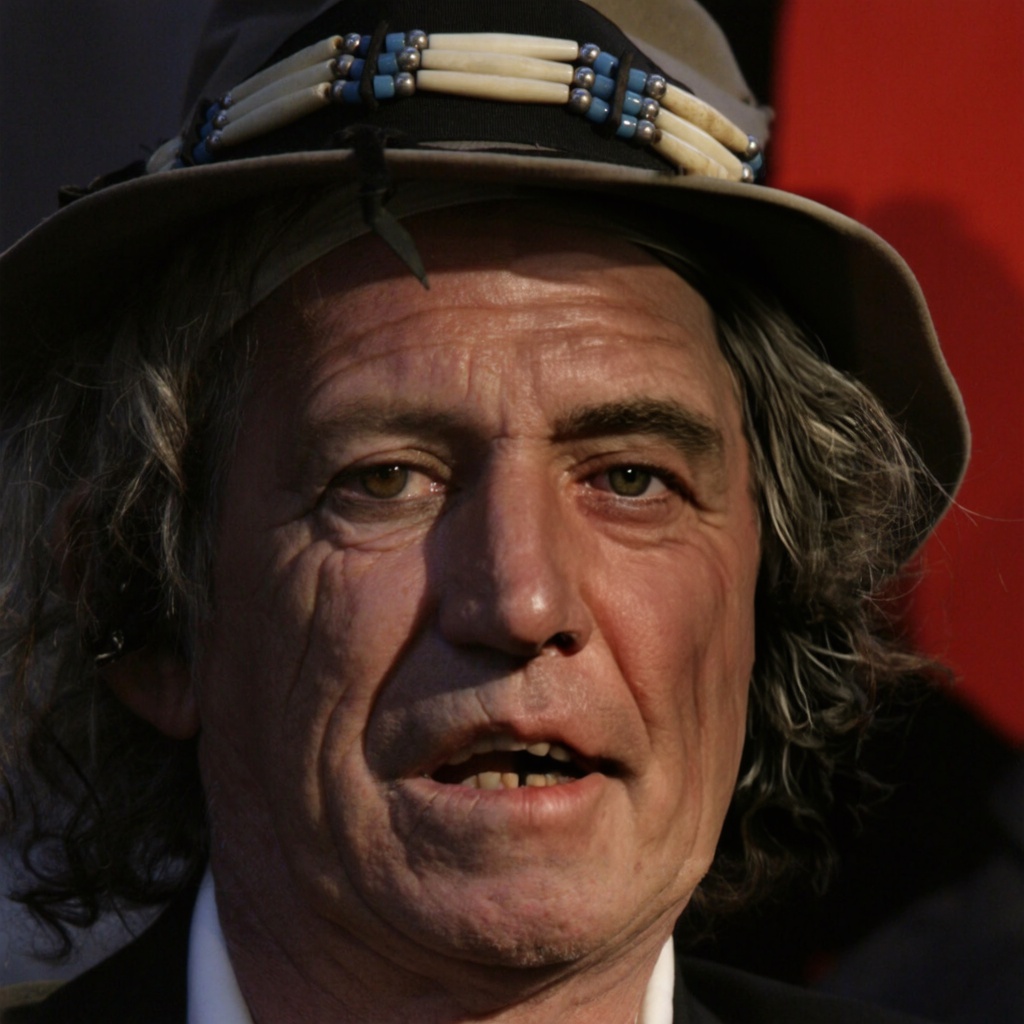} \\
        {\rotatebox[origin=l]{90}{\parbox[b]{2.2cm}{\centering\tiny \divtwok}}} &
        \includegraphics[width=\imwidth]{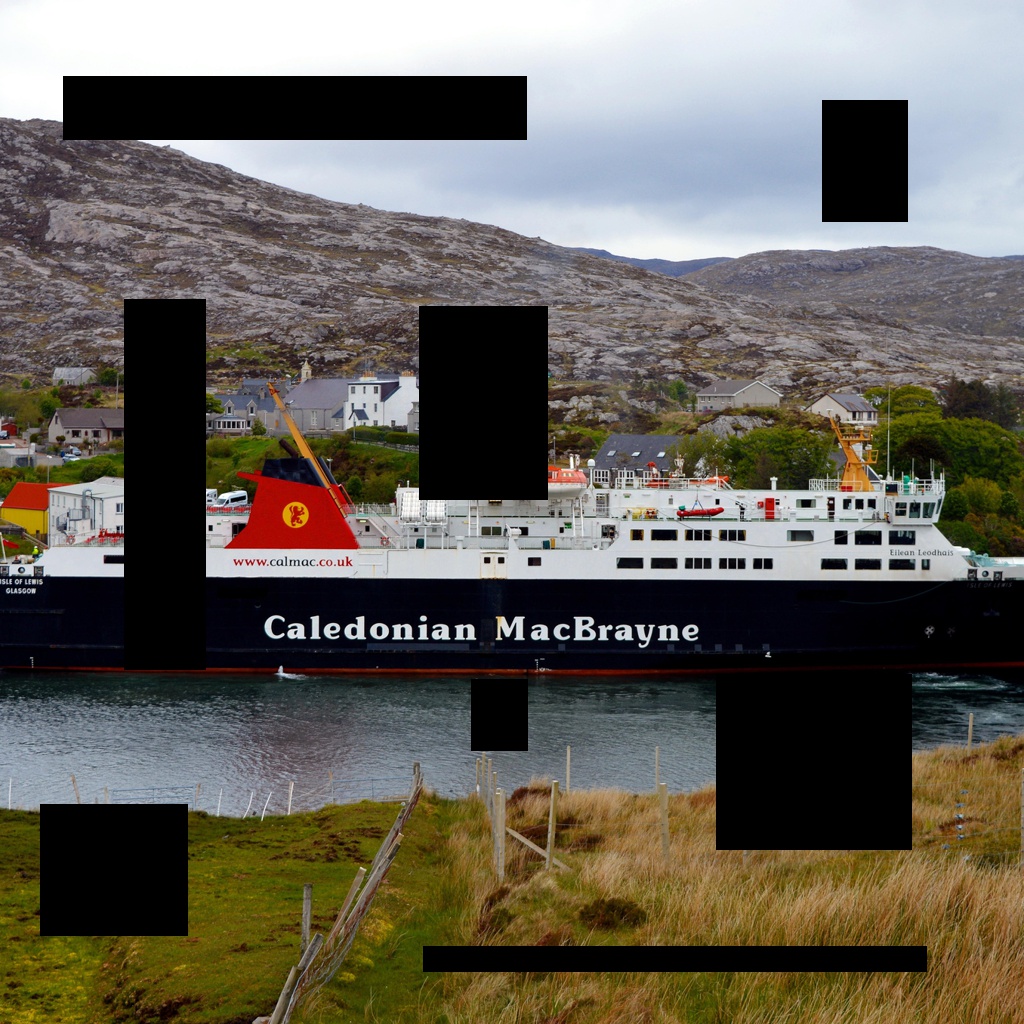} &
        \includegraphics[width=\imwidth]{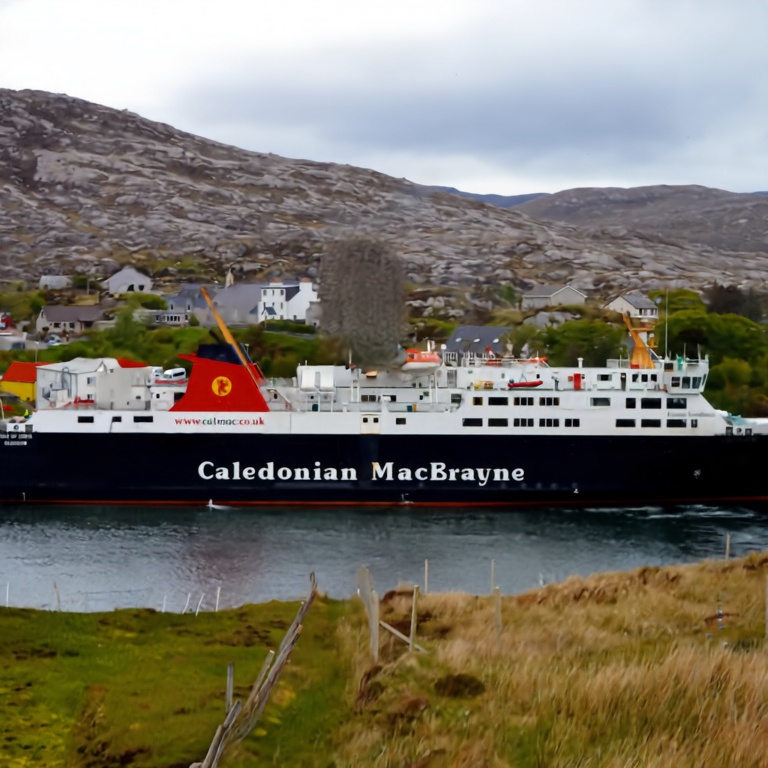} &
        \includegraphics[width=\imwidth]{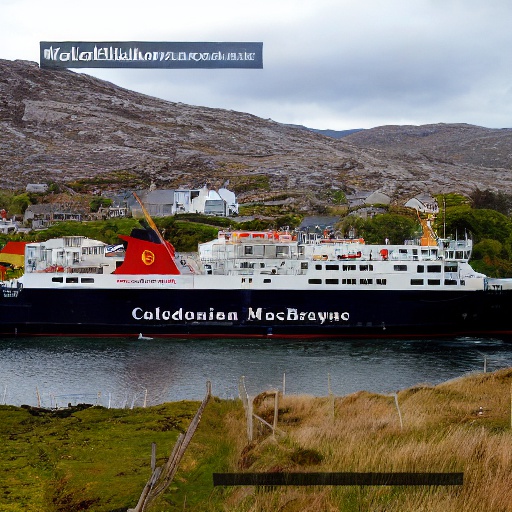} &
        \includegraphics[width=\imwidth]{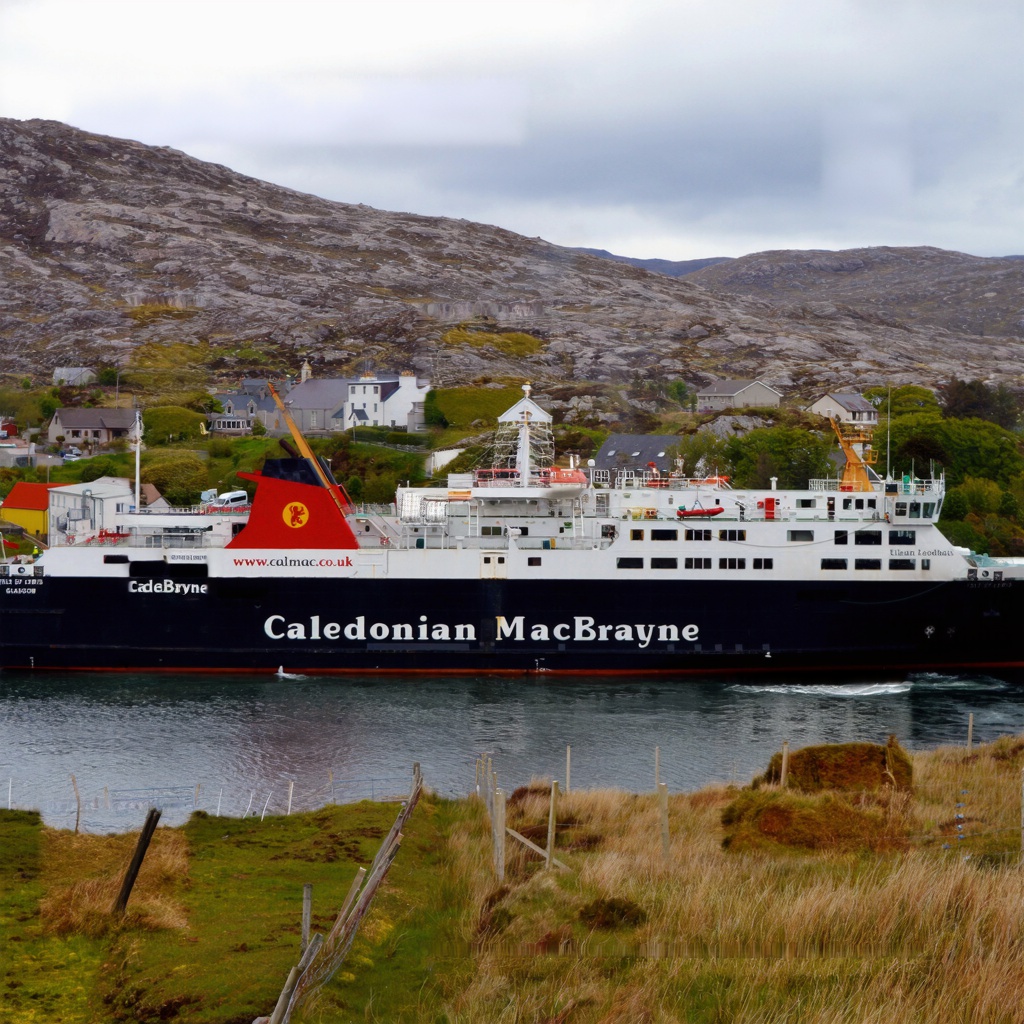} &
        \includegraphics[width=\imwidth]{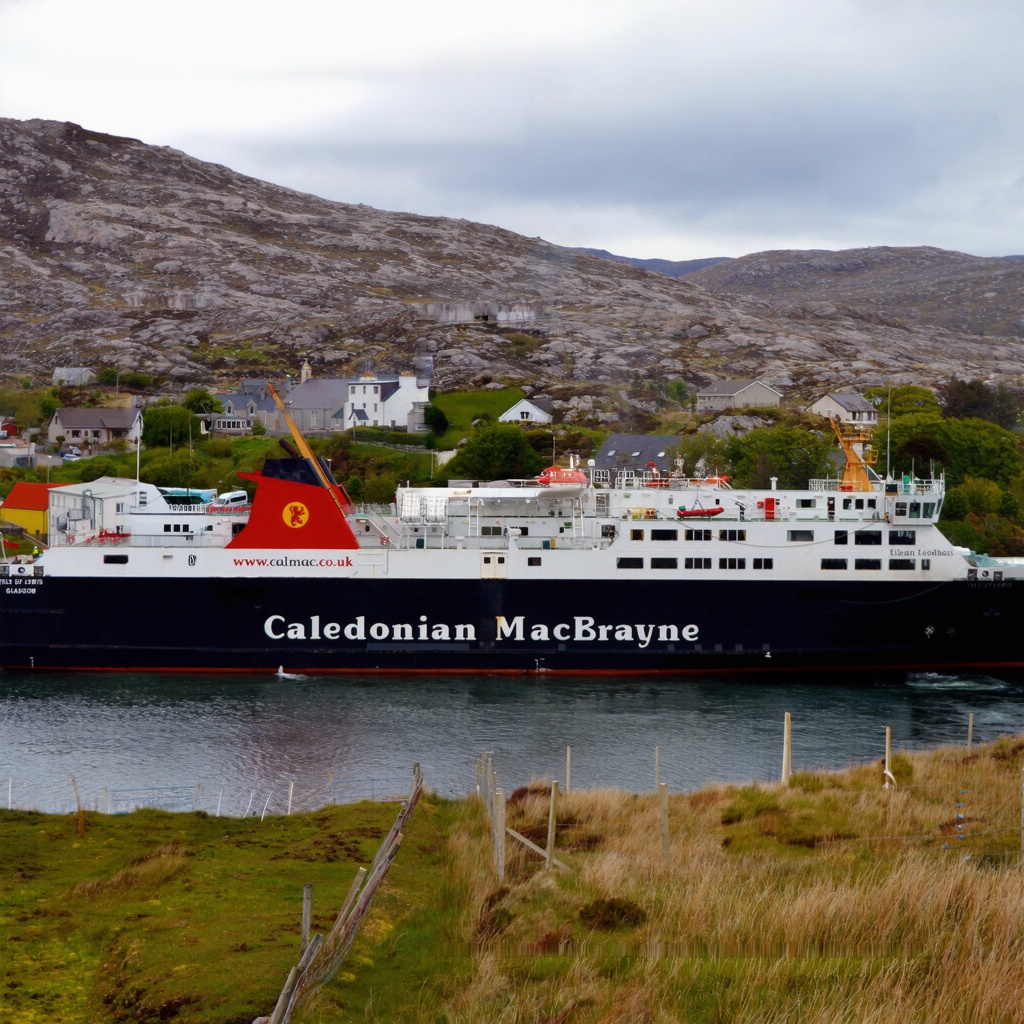} \\
        {\rotatebox[origin=l]{90}{\parbox[b]{2.2cm}{\centering\tiny \brushbench}}} &
        \includegraphics[width=\imwidth]{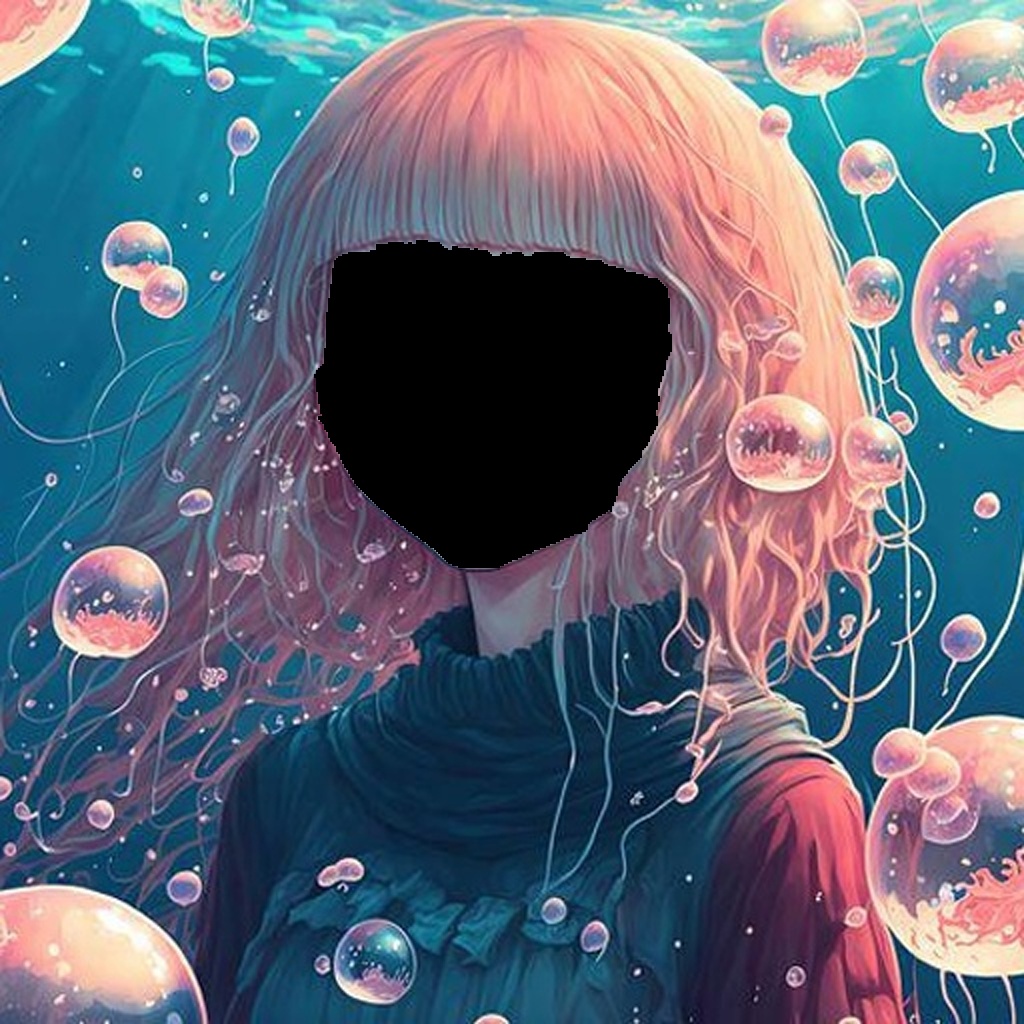} &
        \includegraphics[width=\imwidth]{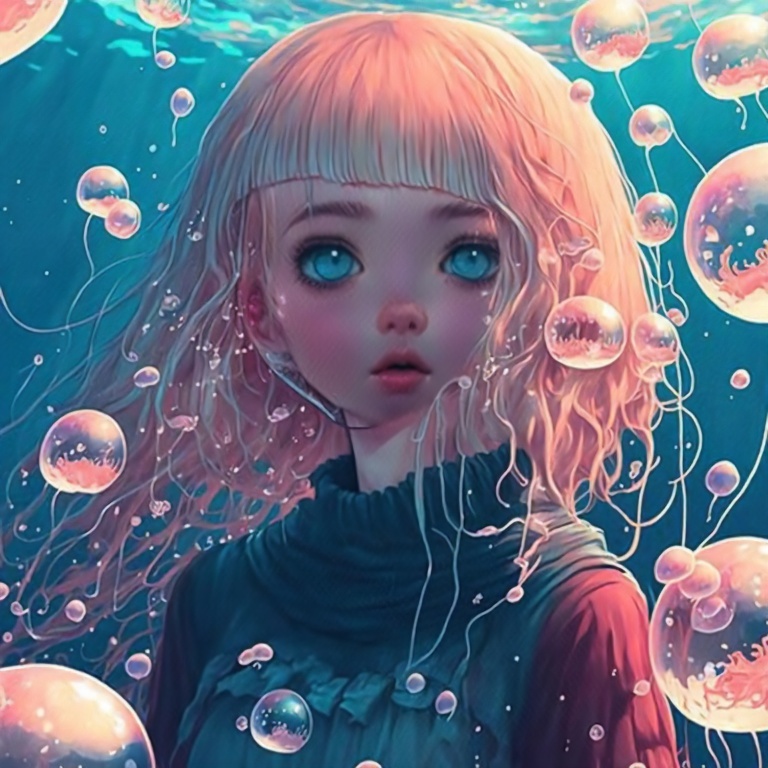} &
        \includegraphics[width=\imwidth]{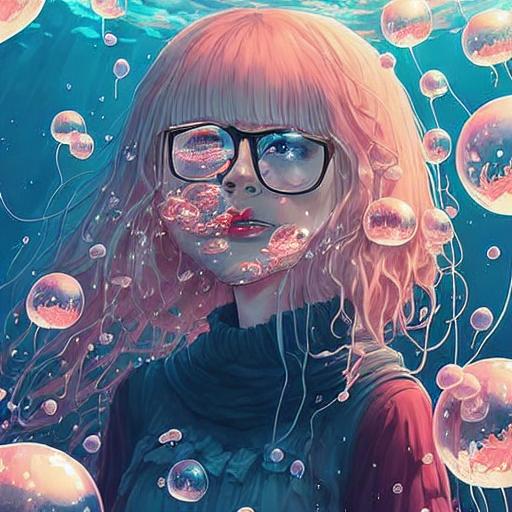} &
        \includegraphics[width=\imwidth]{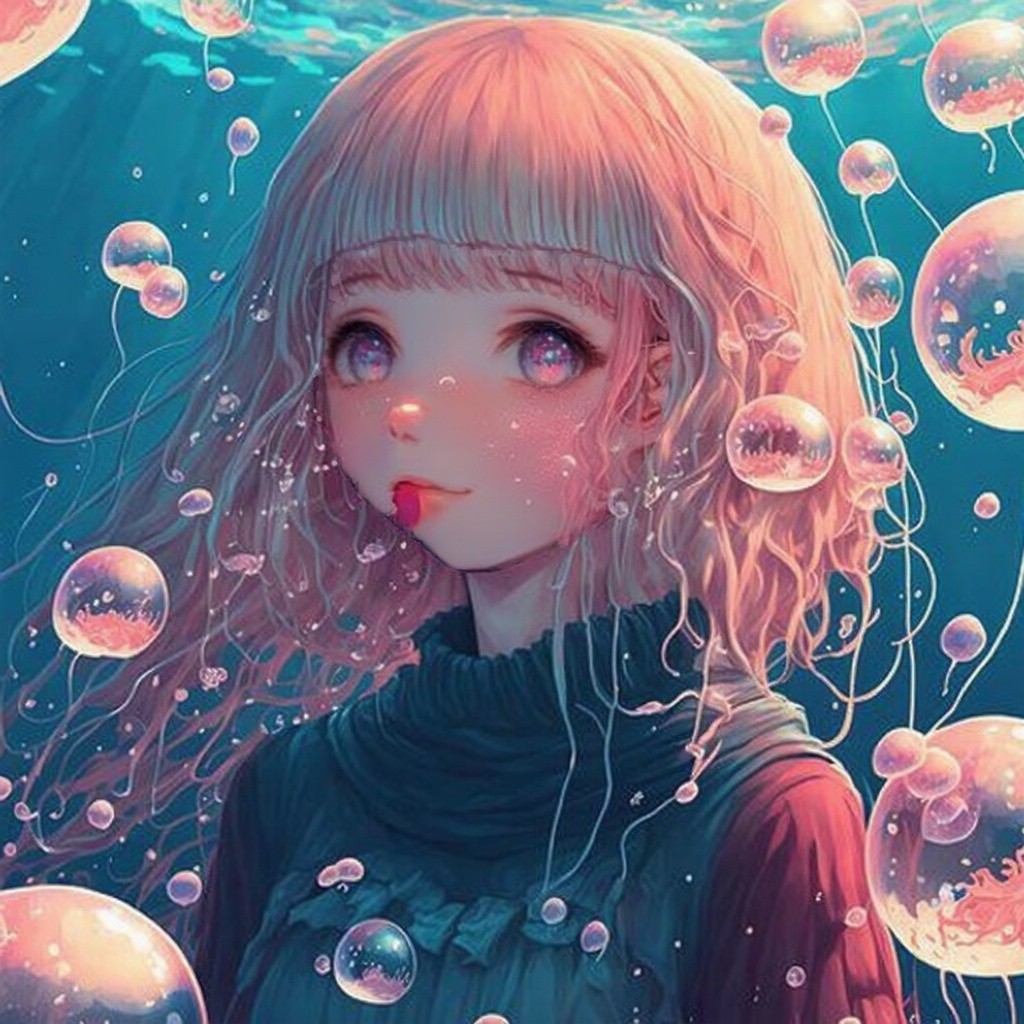} &
        \includegraphics[width=\imwidth]{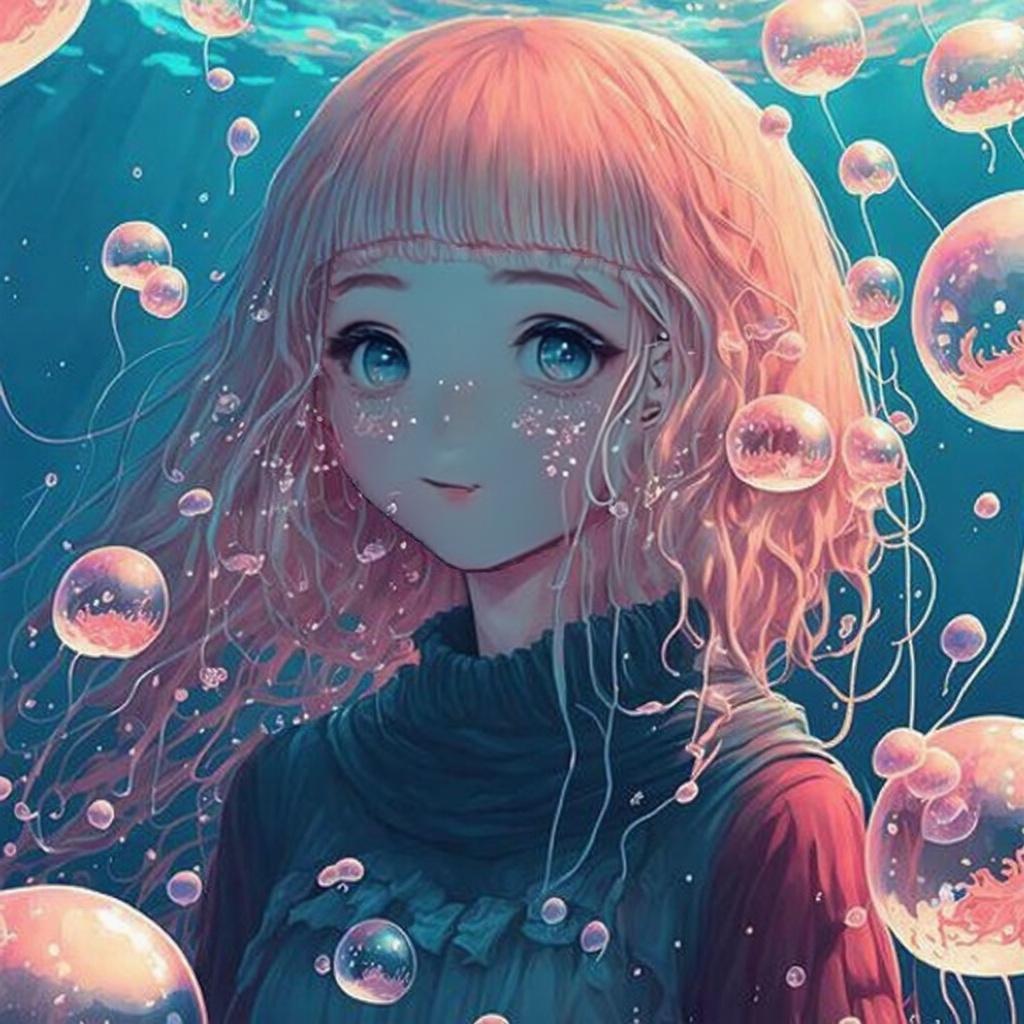} \\
        &
        {\tiny \makecell{Ground truth\\ and masks}} & {\tiny \flair} & {\tiny \brushnet} & {\tiny \bld} & {\tiny \bf Our method}
    \end{tabular}
    \renewcommand{\arraystretch}{1.0}
    \setlength{\tabcolsep}{6pt}
    \vspace{-0.5em}

    \caption{
        {\bf Qualitative results --} 
        We show results across different methods and datasets.
        We denote the masked areas with black rectangles on the ground truth images.
        Our method provides the best inpainting results, especially aligning well with the layout and also the lighting conditions of the input image. 
        Zoom in for better detail.
    }
    \label{fig:qualitative}
    \vspace{-1em}
\end{figure*}

\subsection{Results}

We present a qualitative comparison in \cref{fig:qualitative}.
We compare our method against \flair,
\brushnet, and \bld.
We provide additional qualitative results in the \supp.
We also provide per-dataset quantitative summaries in \cref{tab:quantitative_brushbench,tab:quantitative_div2k,tab:quantitative_ffhq}.

\begin{table}[t]
\centering
\caption{
    {\bf Quantitative results on \ffhq~-- }
    Our method outperforms all other methods in terms of SSIM, LPIPS, and FID by a large margin.
    Our method also performs best in terms of simulated human preference metrics, IR and HPS v2, and second best in AS.
    Note that while PSNR is lower, this does not mean our method provides worse results, as for inpainting, an image different from ground-truth may still be correct.
    \flowdps performs best in terms of CLIP scores, but notice the significantly higher LPIPS and FID scores, indicating poor inpainting quality.
}
\vspace{-0.5em}
\label{tab:quantitative_ffhq}
\setlength{\tabcolsep}{12pt}
\resizebox{\linewidth}{!}{%
\begin{tabular}{@{}lcccccccc@{}}
\toprule
Method & PSNR$\uparrow$ & SSIM$\uparrow$ & LPIPS$\downarrow$ & FID $\downarrow$ & IR $\uparrow$ & HPS v2$\uparrow$ & CLIP$\uparrow$ & AS$\uparrow$\\
\midrule
\brushnet & 21.929 & 0.759 & 0.237 & \second{17.950} & \second{0.275} & \second{0.244} & \third{24.804} & 5.713 \\
\bld & 17.592 & 0.824 & \second{0.180} & 25.842 & 0.038 & 0.213 & 22.456 & 5.136 \\
\flair & \third{23.100} & 0.823 & 0.266 & \third{18.982} & \third{0.220} & \third{0.240} & 24.122 & 5.735 \\
\flair (400 NFEs) & \best{24.099} & \second{0.829} & 0.292 & 19.873 & 0.154 & 0.234 & 23.765 & 5.537 \\
\flowchef & 19.624 & \third{0.828} & \third{0.193} & 20.706 & -0.021 & 0.214 & 22.178 & 5.074 \\
\flowchef (400 NFEs) & 19.336 & 0.824 & 0.197 & 21.191 & -0.015 & 0.214 & 22.066 & 5.029 \\
\flowdps & 20.215 & 0.760 & 0.374 & 48.333 & 0.025 & 0.234 & \best{25.076} & \third{5.792} \\
\flowdps (400 NFEs) & \second{23.243} & 0.786 & 0.389 & 68.644 & 0.008 & 0.232 & \second{25.014} & \best{5.924} \\
\midrule
\textbf{Our method} & 22.562 & \best{0.857} & \best{0.121} & \best{12.883} & \best{0.280} & \best{0.245} & 24.042 & \second{5.848} \\
\bottomrule
\end{tabular}
}
\setlength{\tabcolsep}{6pt}
\vspace{-1em}
\end{table}

\paragraph{\ffhq~-- \cref{fig:qualitative,tab:quantitative_ffhq}.}
As seen in \cref{fig:qualitative}, our method significantly outperforms other baselines, followed by \flair.
\brushnet is great at hiding seams, as it is a trained method, but the content it inpaints can misalign with the full image.
In the case of \bld, it completely fails to align the global structure, resulting in an inpainting failure.
These findings are also reflected in the quantitative metrics presented in \cref{tab:quantitative_ffhq}, with our method outperforming all other methods in terms of SSIM, LPIPS, FID, IR, and HPS~v2.
In terms of FID, our method outperforms all other methods by a large margin, with \brushnet being the second best and \flair being the third best.
As \brushnet does not have seams, it shows up favorably in FID, and other aesthetic metrics.
But in terms of SSIM and LPIPS, where structure is considered, the method performs significantly worse.
While we perform worse in terms of PSNR and CLIP score, we emphasize again that these in isolation \emph{do not} directly correlate with inpainting performance.

\begin{table}[t]
\centering
\caption{
    {\bf Quantitative results on \divtwok~-- }
    Our method performs best in terms of SSIM, LPIPS, and FID by a large margin.
    While \flair shows high PSNR, as shown in \cref{fig:qualitative}, this could be attributed to the fact that it produces blurry patches.
    \flair also has high IR and AS, and \bld and \brushnet show higher CLIP scores, but they all perform significantly worse in terms of LPIPS and FID, indicating poor inpainting quality. 
}
\vspace{-0.5em}
\label{tab:quantitative_div2k}
\setlength{\tabcolsep}{12pt}
\resizebox{\linewidth}{!}{%
\begin{tabular}{@{}lcccccccc@{}}
\toprule
Method & PSNR$\uparrow$ & SSIM$\uparrow$ & LPIPS$\downarrow$ & FID $\downarrow$ & IR $\uparrow$ & HPS v2$\uparrow$ & CLIP$\uparrow$ & AS$\uparrow$ \\
\midrule
\brushnet & 19.414 & 0.575 & 0.308 & 29.004 & 0.255 & 0.249 & \second{27.666} & 5.406 \\
\bld & 19.768 & \second{0.789} & \second{0.144} & 25.328 & \best{0.389} & \best{0.255} & \best{27.752} & 5.883 \\
\flair & \second{23.128} & 0.760 & 0.298 & \third{22.421} & 0.290 & 0.248 & 27.586 & \third{5.909} \\
\flair (400 NFEs) & \best{23.250} & \third{0.768} & 0.294 & \second{22.147} & \second{0.339} & \third{0.251} & 27.555 & \best{5.929} \\
\flowchef & 20.067 & 0.748 & \third{0.237} & 30.481 & 0.169 & 0.240 & 27.277 & 5.062 \\
\flowchef (400 NFEs) & 19.938 & 0.747 & 0.238 & 30.769 & 0.177 & 0.241 & 27.322 & 5.084 \\
\flowdps & 20.596 & 0.628 & 0.424 & 46.611 & 0.179 & 0.237 & 27.583 & 5.802 \\
\flowdps (400 NFEs) & \third{21.728} & 0.650 & 0.420 & 67.893 & 0.041 & 0.231 & 27.522 & 5.514 \\
\midrule
\textbf{Our method} & 21.284 & \best{0.803} & \best{0.115} & \best{17.699} & \third{0.309} & \second{0.253} & \third{27.613} & \second{5.927} \\
\bottomrule
\end{tabular}
}
\setlength{\tabcolsep}{6pt}
\vspace{-1em}
\end{table}

\paragraph{\divtwok~-- \cref{fig:qualitative,tab:quantitative_div2k}.}
Our method performs best for this dataset as well.
\flair struggles and produces blurry patches, which may be why it shows high PSNR in \cref{tab:quantitative_div2k}, but lower SSIM and LPIPS scores.
\brushnet, in this example, produces caption-like inpainting results in horizontal stripe masks, hinting that its inpainting outcomes are strongly correlated with the shape of the mask.
The quantitative results in \cref{tab:quantitative_div2k} further show how this method performs significantly worse than other methods for this dataset.
\bld does not show apparent artifacts, but as can be seen in the square mask in the bottom right, or the vertical mask on the left, its inpainting results do not align well with the full image.
Again, for this dataset, our method outperforms all other methods in terms of SSIM, LPIPS, FID, and HPS v2.

\begin{table}[t]
\centering
\caption{
    {\bf Quantitative results on \brushbench~-- }
    Our method outperforms all baselines in terms of SSIM, LPIPS, and FID.
    For other metrics, ours performs third-best.
    Unsurprisingly, \brushnet performs second best in terms of FID, and best in terms of IR, HPS v2, and AS, as this dataset was developed by \brushnet.
    But in terms of LPIPS and FID, they are worse than ours, in line with \cref{fig:qualitative}.
    \flair again shows high PSNR and SSIM, but performs worse in all other metrics.
}
\vspace{-0.5em}
\label{tab:quantitative_brushbench}
\setlength{\tabcolsep}{12pt}
\resizebox{\linewidth}{!}{%
\begin{tabular}{@{}lcccccccc@{}}
\toprule
Method & PSNR$\uparrow$ & SSIM$\uparrow$ & LPIPS$\downarrow$ & FID $\downarrow$ & IR $\uparrow$ & HPS v2$\uparrow$ & CLIP$\uparrow$ & AS$\uparrow$\\
\midrule
\brushnet & 18.668 & 0.740 & 0.199 & \second{50.082} & \best{1.246} & \best{0.271} & 26.551 & \best{6.366} \\
\bld & 18.234 & 0.854 & 0.169 & 51.661 & \second{1.211} & \second{0.267} & \best{27.262} & \second{6.072} \\
\flair & \second{19.986} & \third{0.855} & 0.201 & 51.644 & 1.007 & 0.255 & 26.655 & 5.978 \\
\flair (400 NFEs) & \best{20.154} & \third{0.855} & 0.190 & \third{51.535} & 1.025 & 0.255 & 26.455 & 5.878  \\
\flowchef & 18.711 & \second{0.856} & \second{0.161} & 53.568 & 0.975 & 0.253 & 26.680 & 5.826  \\
\flowchef (400 NFEs) & 18.688 & 0.853 & \second{0.161} & 52.281 & 1.025 & 0.256 & 26.767 & 5.877 \\
\flowdps & 18.269 & 0.775 & 0.360 & 67.731 & 0.895 & 0.248 & \second{27.045} & 5.909 \\
\flowdps (400 NFEs) & 18.558 & 0.795 & 0.338 & 80.544 & 0.832 & 0.241 & 26.742 & 5.813 \\
\midrule
\textbf{Our Method} & \third{19.014} & \best{0.861} & \best{0.153} & \best{48.072} & \third{1.188} & \third{0.263} & \third{26.890} & \third{6.061} \\
\bottomrule
\end{tabular}
}
\setlength{\tabcolsep}{6pt}
\vspace{-1em}
\end{table}

\paragraph{\brushbench~-- \cref{fig:qualitative,tab:quantitative_brushbench}.}
Again, our method performs best.
\flair, for this dataset, is unable to match the global structure of the image well and often produces a disjoint inpainting outcome.
\brushnet, which is the method developed for this dataset, provides results that seamlessly align with the local surroundings, but often misaligns the rough global structure.
\bld, for this example, produces arguably the second best results, which aligns with the FID-based evaluation in \cref{tab:quantitative_brushbench}.
Our method, quantitatively, outperforms all methods in terms of SSIM, LPIPS, FID, and performs third-best in terms of other metrics.
We emphasize again, that each metric alone does not paint a complete picture of the inpainting quality, and all metrics should be considered together.
Among them, we believe LPIPS and FID \emph{best} represent the performance of each method, as they compare the perceived similarity between the inpainting results and the ground truth, tolerating minor differences in exact structure and color.

\subsection{Ablation study}
\label{sec:ablation}

We perform ablation studies to motivate our design choices. 
We summarize the quantitative results in \cref{tab:ablation}, where we compare the inpainting performance without spectral conditioning and spatial constraining.
When removing spectral conditioning, \ie, optimizing in the spatial domain, we use a learning rate of 0.05 for Adam~\cite{kingma2017adammethodstochasticoptimization}.
Observe that both spectral preconditioning and spatial constraining are critical to the performance of the method.
Also, optimizing with the nearest-neighbor fill-in produces similar metrics as optimizing with the ground truth image, which would be the upper-bound performance.
This confirms the robustness of our method and that the simple fill-in strategy is sufficient.

For further analysis on the convergence enhancement of spectral preconditioning, we synthesize images whom we \emph{know} their initial noise sample of, and plot their convergence in \cref{fig:optimization_sweep}.
Specifically, we use the first 100 prompts from \divtwok, with various learning rates, using \sdthreefive with CFG${=}$2.0, T=20.
For all methods, we use the same random starting point, drawn from $\mathcal{N}(0, 1)$.
As shown, Adam~\cite{kingma2017adammethodstochasticoptimization} in spectral domain converges significantly faster, allows larger learning rates thanks to the preconditioning.
Adam in spatial domain converges much slower, and becomes unstable if too large learning rates are used. 
As Fourier transform is linear, SGD in spectral domain has no impact on optimization---they both perform worse.

\paragraph{More ablations.}
In the \supp, we provide qualitative comparisons with randomly selected samples for objective comparison; and ablation studies on the number of optimization steps used, which show diminishing gains as we increase the number of optimization iterations.

\begin{table}[t]
\centering
\caption{
    {\bf Ablation study -- }
    We report the evaluation metrics for \ffhq dataset with variants of our method: using spatial optimization, \ie, without spectral preconditioning; removing spatial constraining of gradient updates; and using ground-truth image as input to the encoder, which shows the upper-bound performance of operating in the latent space.
    Both spectral preconditioning and spatial constraining are critical. 
    Using ground-truth image for the encoder shows similar performance as our simple nearest-neighbor fill-ins, showing that this simple strategy is sufficient.
}
\vspace{-0.5em}
\label{tab:ablation}
\setlength{\tabcolsep}{12pt}
\resizebox{\linewidth}{!}{%
\begin{tabular}{@{}lcccccccc@{}}
\toprule
Variant & PSNR$\uparrow$ & SSIM$\uparrow$ & LPIPS$\downarrow$ & FID$\downarrow$ & IR$\uparrow$ & HPS v2$\uparrow$ & CLIP$\uparrow$ & AS$\uparrow$ \\
\midrule
w/o spectral preconditioning & 22.120 & 0.853 & 0.127 & 14.579 & 0.216 & 0.239 & 23.654 & 5.760 \\
w/o spatial constraining & 18.863 & 0.833 & 0.165 & 33.269 & -0.259 & 0.223 & 22.676 & 5.365 \\
Optimizing with the GT & \best{22.569} & \second{0.856} & \best{0.121} & \best{12.773} & \best{0.280} & \best{0.245} & \best{24.074} & \best{5.850} \\
\midrule
Our method & \second{22.562} & \best{0.857} & \best{0.121} & \second{12.883} & \best{0.280} & \best{0.245} & \second{24.042} & \second{5.848} \\
\bottomrule
\end{tabular}
}
\setlength{\tabcolsep}{6pt}
\vspace{-1em}
\end{table}

\section{Conclusion}
\label{sec:conclusion}

We have presented a novel training-free method for inpainting that optimizes the initial noise sample.
We have shown that by finding the right initial noise with our method, we can achieve inpainting results that outperform the state of the art.
The key ideas that enable this are the linear approximation strategy that allows optimization of the initial noise without back-propagation through the denoiser, and spectral preconditioning, realized by optimizing in the spectral domain, ensuring fast and stable convergence.
Spatially constraining the updates further ensures that the optimization does not stray from the initial noise sample manifold.
We conduct experiments on three standard inpainting datasets \ffhq, \divtwok, and \brushbench, demonstrating the effectiveness of our method.

\paragraph{Limitations and future work.} 
While we have demonstrated our method purely with an inpainting task, we believe our method of linearization and spectral preconditioning could, in theory, be applied to other inverse problems such as super-resolution, deblurring, and video inpainting.\footnote{We show preliminary video inpainting results in the \supp.} We hope our work can serve as a starting point for future research that directly utilizes generic flow models for other inverse problems.

\section*{Acknowledgements}
This work was supported in part by the Natural Sciences and Engineering Research Council of Canada (NSERC) Discovery Grant, NSERC Alliance Advantage Grant, Google, the Digital Research Alliance of Canada, and the Advanced Research Computing and the Computer Science department at the University of British Columbia.

{\small
\bibliographystyle{splncs04}
\bibliography{macros,main}

@STRING{AAAI     = "American Association for Artificial Intelligence Conference"}

@STRING{AI       = "Artificial Intelligence"}

@STRING{ARXIV    = "arXiv Preprint"}

@STRING{CVPR     = "Conference on Computer Vision and Pattern Recognition"}

@STRING{CVPRW    = "Conference on Computer Vision and Pattern Recognition Workshops"}

@STRING{ECCV     = "European Conference on Computer Vision"}

@STRING{ICCV     = "International Conference on Computer Vision"}

@STRING{ICLR     = "International Conference on Learning Representations"}

@STRING{ICML     = "International Conference on Machine Learning"}

@STRING{JI       = "Journal of Imaging"}

@STRING{NIPS     = "Advances in Neural Information Processing Systems"}

@STRING{TIP      = "IEEE Transactions on Image Processing"}

@STRING{TOG      = "ACM Transactions on Graphics"}

@STRING{WACV     = "IEEE Winter Conference on Applications of Computer Vision"}

@String(TPAMI = {IEEE Trans. Pattern Anal. Mach. Intell.})

@String(CVPR= {IEEE Conf. Comput. Vis. Pattern Recog.})

@String(ICCV= {Int. Conf. Comput. Vis.})

@String(ECCV= {Eur. Conf. Comput. Vis.})

@String(NeurIPS= {Adv. Neural Inform. Process. Syst.})

@String(NIPS= {Adv. Neural Inform. Process. Syst.})

@String(TOG= {ACM Trans. Graph.})

@String(TIP  = {IEEE Trans. Image Process.})

@String(ACMMM= {ACM Int. Conf. Multimedia})

@String(ICLR = {Int. Conf. Learn. Represent.})

@String(AAAI = {AAAI Conf. Artificial Intell.})

@String(CVPRW= {IEEE Conf. Comput. Vis. Pattern Recog. Worksh.})

@String(WACV = {IEEE Winter Conf. Appl. Comput. Vis.})

@String(arxiv = {arXiv})

@article{ho2020denoising,
  title={Denoising diffusion probabilistic models},
  author={Ho, Jonathan and Jain, Ajay and Abbeel, Pieter},
  journal={Advances in neural information processing systems},
  volume={33},
  pages={6840--6851},
  year={2020}
}

@inproceedings{
song2020denoising,
title={Denoising Diffusion Implicit Models},
author={Jiaming Song and Chenlin Meng and Stefano Ermon},
booktitle={International Conference on Learning Representations},
year={2021},
url={https://openreview.net/forum?id=St1giarCHLP}
}

@inproceedings{rombach2022high,
  title={High-resolution image synthesis with latent diffusion models},
  author={Rombach, Robin and Blattmann, Andreas and Lorenz, Dominik and Esser, Patrick and Ommer, Bj{\"o}rn},
  booktitle={Proceedings of the IEEE/CVF conference on computer vision and pattern recognition},
  pages={10684--10695},
  year={2022}
}

@inproceedings{
lipman2023flow,
title={Flow Matching for Generative Modeling},
author={Yaron Lipman and Ricky T. Q. Chen and Heli Ben-Hamu and Maximilian Nickel and Matthew Le},
booktitle={The Eleventh International Conference on Learning Representations },
year={2023},
}

@inproceedings{liu2023rectifiedflow,
  title     = {Flow Straight and Fast: Learning to Generate and Transfer Data with Rectified Flow},
  author    = {Liu, Xingchao and Gong, Chengyue and Liu, Qiang},
  booktitle = ICLR,
  year      = {2023},
}

@inproceedings{esser2024,
author = {Esser, Patrick and Kulal, Sumith and Blattmann, Andreas and Entezari, Rahim and M\"{u}ller, Jonas and Saini, Harry and Levi, Yam and Lorenz, Dominik and Sauer, Axel and Boesel, Frederic and Podell, Dustin and Dockhorn, Tim and English, Zion and Rombach, Robin},
title = {Scaling rectified flow transformers for high-resolution image synthesis},
year = {2024},
booktitle = ICML,
articleno = {503}
}

@inproceedings{bora2017compressed,
  title={Compressed sensing using generative models},
  author={Bora, Ashish and Jalal, Ajil and Price, Eric and Dimakis, Alexandros G},
  booktitle={International conference on machine learning},
  pages={537--546},
  year={2017},
  organization={PMLR}
}

@inproceedings{yeh2017semantic,
  title={Semantic image inpainting with deep generative models},
  author={Yeh, Raymond A and Chen, Chen and Yian Lim, Teck and Schwing, Alexander G and Hasegawa-Johnson, Mark and Do, Minh N},
  booktitle={Proceedings of the IEEE conference on computer vision and pattern recognition},
  pages={5485--5493},
  year={2017}
}

@InProceedings{Wallace_2023_ICCV,
    author    = {Wallace, Bram and Gokul, Akash and Ermon, Stefano and Naik, Nikhil},
    title     = {End-to-End Diffusion Latent Optimization Improves Classifier Guidance},
    booktitle = {Proceedings of the IEEE/CVF International Conference on Computer Vision (ICCV)},
    month     = {October},
    year      = {2023},
    pages     = {7280-7290}
}

@inproceedings{karunratanakul2024optimizing,
  title={Optimizing diffusion noise can serve as universal motion priors},
  author={Karunratanakul, Korrawe and Preechakul, Konpat and Aksan, Emre and Beeler, Thabo and Suwajanakorn, Supasorn and Tang, Siyu},
  booktitle={Proceedings of the IEEE/CVF Conference on Computer Vision and Pattern Recognition},
  pages={1334--1345},
  year={2024}
}

@inproceedings{
tang2025inferencetime,
title={Inference-Time Alignment of Diffusion Models with Direct Noise Optimization},
author={Zhiwei Tang and Jiangweizhi Peng and Jiasheng Tang and Mingyi Hong and Fan Wang and Tsung-Hui Chang},
booktitle={Forty-second International Conference on Machine Learning},
year={2025},
}

@InProceedings{ben2024d,
  title = 	 {D-Flow: Differentiating through Flows for Controlled Generation},
  author =       {Ben-Hamu, Heli and Puny, Omri and Gat, Itai and Karrer, Brian and Singer, Uriel and Lipman, Yaron},
  booktitle = 	 {Proceedings of the 41st International Conference on Machine Learning},
    year = 	 {2024}
}

@inproceedings{guo2024initno,
  title={Initno: Boosting text-to-image diffusion models via initial noise optimization},
  author={Guo, Xiefan and Liu, Jinlin and Cui, Miaomiao and Li, Jiankai and Yang, Hongyu and Huang, Di},
  booktitle={Proceedings of the IEEE/CVF Conference on Computer Vision and Pattern Recognition},
  pages={9380--9389},
  year={2024}
}

@inproceedings{ReNO2024,
 author = {Eyring, Luca and Karthik, Shyamgopal and Roth, Karsten and Dosovitskiy, Alexey and Akata, Zeynep},
 booktitle = {Advances in Neural Information Processing Systems},
 editor = {A. Globerson and L. Mackey and D. Belgrave and A. Fan and U. Paquet and J. Tomczak and C. Zhang},
 pages = {125487--125519},
 publisher = {Curran Associates, Inc.},
 title = {ReNO: Enhancing One-step Text-to-Image Models through Reward-based Noise Optimization},
 volume = {37},
 year = {2024}
}

@inproceedings{ho2021classifierfree,
title={Classifier-Free Diffusion Guidance},
author={Jonathan Ho and Tim Salimans},
booktitle={NeurIPS 2021 Workshop on Deep Generative Models and Downstream Applications},
year={2021},
}

@inproceedings{bansal2023universalguidancediffusionmodels,
      title={Universal Guidance for Diffusion Models}, 
      author={Arpit Bansal and Hong-Min Chu and Avi Schwarzschild and Soumyadip Sengupta and Micah Goldblum and Jonas Geiping and Tom Goldstein},
      year={2024},
      booktitle=ICLR,
}

@inproceedings{he2023manifoldpreservingguideddiffusion,
      title={Manifold Preserving Guided Diffusion}, 
      author={Yutong He and Naoki Murata and Chieh-Hsin Lai and Yuhta Takida and Toshimitsu Uesaka and Dongjun Kim and Wei-Hsiang Liao and Yuki Mitsufuji and J. Zico Kolter and Ruslan Salakhutdinov and Stefano Ermon},
      year={2024},
booktitle=ICLR,
}

@inproceedings{luo2024readoutguidance,
    title={Readout Guidance: Learning Control from Diffusion Features},
    author={Grace Luo and Trevor Darrell and Oliver Wang and Dan B Goldman and Aleksander Holynski},
    booktitle=CVPR,
    year={2024}
}

@inproceedings{
guo2024gradient,
title={Gradient Guidance for Diffusion Models: An Optimization Perspective},
author={Yingqing Guo and Hui Yuan and Yukang Yang and Minshuo Chen and Mengdi Wang},
booktitle={The Thirty-eighth Annual Conference on Neural Information Processing Systems},
year={2024},
}

@article{bengio2013estimating,
  title={Estimating or propagating gradients through stochastic neurons for conditional computation},
  author={Bengio, Yoshua and L{\'e}onard, Nicholas and Courville, Aaron},
  journal={arXiv preprint arXiv:1308.3432},
  year={2013}
}

@inproceedings{
moufad2026efficient,
title={Efficient Zero-shot Inpainting with Decoupled Diffusion Guidance},
author={Badr MOUFAD and Yazid Janati and Navid Bagheri Shouraki and Alain Oliviero Durmus and Thomas Hirtz and Eric Moulines and Jimmy Olsson},
booktitle={The Fourteenth International Conference on Learning Representations},
year={2026},
}

@inproceedings{tao2017zero,
  title={Zero-order reverse filtering},
  author={Tao, Xin and Zhou, Chao and Shen, Xiaoyong and Wang, Jue and Jia, Jiaya},
  booktitle={Proceedings of the IEEE International Conference on Computer Vision},
  pages={222--230},
  year={2017}
}

@article{milanfarblackbox,
author = {Milanfar, Peyman},
title = {Rendition: Reclaiming What a Black Box Takes Away},
journal = {SIAM Journal on Imaging Sciences},
volume = {11},
number = {4},
pages = {2722-2756},
year = {2018},
}

@inproceedings{
ahn2026a,
title={A Noise is Worth Diffusion Guidance},
author={Donghoon Ahn and Jiwon Kang and Sanghyun Lee and Jaewon Min and Minjae Kim and Wooseok Jang and Hyoungwon Cho and Sayak Paul and SeonHwa Kim and Eunju Cha and Kyong Hwan Jin and Seungryong Kim},
booktitle={The Fourteenth International Conference on Learning Representations},
year={2026},
}

@inproceedings{wu2024freeinit,
  title={Freeinit: Bridging initialization gap in video diffusion models},
  author={Wu, Tianxing and Si, Chenyang and Jiang, Yuming and Huang, Ziqi and Liu, Ziwei},
  booktitle={European conference on computer vision},
  pages={378--394},
  year={2024},
  organization={Springer}
}

@inproceedings{Wang2025Seeds,
  title        = {Seeds of Structure: Patch PCA Reveals Universal Compositional Cues in Diffusion Models},
  author       = {Qingsong Wang and Zhengchao Wan and Mikhail Belkin and Yusu Wang},
  booktitle    = NIPS,
  year         = {2025},
}

@inproceedings{Li2025ReliableSeeds,
  title        = {All Seeds Are Not Equal: Enhancing Compositional Text-to-Image Generation with Reliable Random Seeds},
  author       = {Li, Shuangqi and Le, Hieu and Xu, Jingyi and Salzmann, Mathieu},
  booktitle    = ICLR,
  year         = {2025},
}

@inproceedings{zhou2025golden,
  title={Golden noise for diffusion models: A learning framework},
  author={Zhou, Zikai and Shao, Shitong and Bai, Lichen and Zhang, Shufei and Xu, Zhiqiang and Han, Bo and Xie, Zeke},
  booktitle={Proceedings of the IEEE/CVF International Conference on Computer Vision},
  pages={17688--17697},
  year={2025}
}

@inproceedings{
kim2026model,
title={Model Already Knows the Best Noise: Bayesian Active Noise Selection via Attention  in Video Diffusion Model},
author={Kwanyoung Kim and Sanghyun Kim},
booktitle={The Fourteenth International Conference on Learning Representations},
year={2026},
}

@inproceedings{
jiao2025uniedit,
title={UniEdit-Flow: Unleashing Inversion and Editing in the Era of Flow Models},
author={Guanlong Jiao and Biqing Huang and Kuan-Chieh Jackson Wang and Renjie Liao},
booktitle={The Fourteenth International Conference on Learning Representations},
year={2026}
}

@inproceedings{karras2019stylegan,
  title={A Style-Based Generator Architecture for Generative Adversarial Networks},
  author={Karras, Tero and Laine, Samuli and Aila, Timo},
  booktitle=CVPR,
  year={2019}
}

@inproceedings{div2k,
  title={{NTIRE} 2017 Challenge on Single Image Super-Resolution: Dataset and Study},
  author={Agustsson, Eirikur and Timofte, Radu},
  booktitle=CVPRW,
  year={2017}
}

@inproceedings{ju2024brushnet,
  title={BrushNet: A Plug-and-Play Image Inpainting Model with Decomposed Dual-Branch Diffusion}, 
  author={Xuan Ju and Xian Liu and Xintao Wang and Yuxuan Bian and Ying Shan and Qiang Xu},
  year={2024},
  booktitle=ECCV,
}

@inproceedings{wang2025towards,
  title={Towards Enhanced Image Inpainting: Mitigating Unwanted Object Insertion and Preserving Color Consistency},
  author={Wang, Yikai and Cao, Chenjie and Yu, Junqiu and Fan, Ke and Xue, Xiangyang and Fu, Yanwei},
  booktitle=CVPR,
  year={2025}
}

@inproceedings{flair,
      title={Solving Inverse Problems with FLAIR}, 
      author={Julius Erbach and Dominik Narnhofer and Andreas Dombos and Bernt Schiele and Jan Eric Lenssen and Konrad Schindler},
      year={2025},
booktitle=NIPS,
}

@article{Avrahami_2023,
   title={Blended Latent Diffusion},
   volume={42},
   number={4},
   journal=TOG,
   author={Avrahami, Omri and Fried, Ohad and Lischinski, Dani},
   year={2023},
   month=jul}

@inproceedings{controlnet,
  title     = {Adding Conditional Control to Text-to-Image Diffusion Models},
  author    = {Zhang, Lvmin and Rao, Anyi and Agrawala, Maneesh},
  booktitle = ICCV,
  year      = {2023},
}

@inproceedings{flowdps,
  title={FlowDPS: Flow-driven posterior sampling for inverse problems},
  author={Kim, Jeongsol and Kim, Bryan Sangwoo and Ye, Jong Chul},
  booktitle=ICCV,
  year={2025},
}

@inproceedings{flowchef,
  title={FlowChef: Steering of Rectified Flow Models for Controlled Generations},
  author={Patel, Maitreya and Wen, Song and Metaxas, Dimitris N. and Yang, Yezhou},
  booktitle=ICCV,
  year={2025},
}

@inproceedings{lugmayr2022repaint,
  title={Repaint: Inpainting using denoising diffusion probabilistic models},
  author={Lugmayr, Andreas and Danelljan, Martin and Romero, Andres and Yu, Fisher and Timofte, Radu and Van Gool, Luc},
  booktitle=CVPR,
  year={2022}
}

@inproceedings{kawar2022denoising,
  title={Denoising diffusion restoration models},
  author={Kawar, Bahjat and Elad, Michael and Ermon, Stefano and Song, Jiaming},
  booktitle=NIPS,
  year={2022}
}

@inproceedings{suvorov2022resolution,
  title={Resolution-robust large mask inpainting with fourier convolutions},
  author={Suvorov, Roman and Logacheva, Elizaveta and Mashikhin, Anton and Remizova, Anastasia and Ashukha, Arsenii and Silvestrov, Aleksei and Kong, Naejin and Goka, Harshith and Park, Kiwoong and Lempitsky, Victor},
  booktitle=WACV,
  year={2022}
}

@article{Huang_2025,
   title={Diffusion Model-Based Image Editing: A Survey},
   volume={47},
   number={6},
   journal=TPAMI,
   author={Huang, Yi and Huang, Jiancheng and Liu, Yifan and Yan, Mingfu and Lv, Jiaxi and Liu, Jianzhuang and Xiong, Wei and Zhang, He and Cao, Liangliang and Chen, Shifeng},
   year={2025},
   month=jun}

@article{daras2024surveydiffusionmodelsinverse,
      title={A Survey on Diffusion Models for Inverse Problems}, 
      author={Giannis Daras and Hyungjin Chung and Chieh-Hsin Lai and Yuki Mitsufuji and Jong Chul Ye and Peyman Milanfar and Alexandros G. Dimakis and Mauricio Delbracio},
      year={2024},
journal=arxiv,
}

@inproceedings{zhuang2023task,
  title={A task is worth one word: Learning with task prompts for high-quality versatile image inpainting},
  author={Zhuang, Junhao and Zeng, Yanhong and Liu, Wenran and Yuan, Chun and Chen, Kai},
  booktitle=ECCV,
  year={2024},
}

@inproceedings{xie2025turbofill,
  title={TurboFill: Adapting Few-step Text-to-image Model for Fast Image Inpainting},
  author={Liangbin Xie and Daniil Pakhomov and Zhonghao Wang and Zongze Wu and Ziyan Chen and Yuqian Zhou and Haitian Zheng and Zhifei Zhang and Zhe Lin and Jiantao Zhou and Chao Dong},
  booktitle=CVPR,
  year={2025}
}

@inproceedings{Corneanu_2024_WACV,
    author    = {Corneanu, Ciprian and Gadde, Raghudeep and Martinez, Aleix M.},
    title     = {LatentPaint: Image Inpainting in Latent Space With Diffusion Models},
    booktitle = WACV,
    year      = {2024},
}

@inproceedings{kim2024radregionawarediffusionmodels,
      title={RAD: Region-Aware Diffusion Models for Image Inpainting}, 
      author={Sora Kim and Sungho Suh and Minsik Lee},
      year={2025},
booktitle=CVPR,
}

@inproceedings{manukyan2024hdpainterhighresolutionpromptfaithfultextguided,
      title={HD-Painter: High-Resolution and Prompt-Faithful Text-Guided Image Inpainting with Diffusion Models}, 
      author={Hayk Manukyan and Andranik Sargsyan and Barsegh Atanyan and Zhangyang Wang and Shant Navasardyan and Humphrey Shi},
      year={2025},
booktitle=ICLR,
}

@inproceedings{xie2022smartbrushtextshapeguided,
      title={SmartBrush: Text and Shape Guided Object Inpainting with Diffusion Model},
      author={Shaoan Xie and Zhifei Zhang and Zhe Lin and Tobias Hinz and Kun Zhang},
      year={2023},
      booktitle=CVPR,
}

@inproceedings{yang2022paint,
  title={Paint by Example: Exemplar-based Image Editing with Diffusion Models},
  author={Binxin Yang and Shuyang Gu and Bo Zhang and Ting Zhang and Xuejin Chen and Xiaoyan Sun and Dong Chen and Fang Wen},
  booktitle=CVPR,
  year={2023}
}

@inproceedings{Yang_2023,
   title={Uni-paint: A Unified Framework for Multimodal Image Inpainting with Pretrained Diffusion Model},
   booktitle=ACMMM,
   author={Yang, Shiyuan and Chen, Xiaodong and Liao, Jing},
   year={2023},
}

@inproceedings{zhang2023coherentimageinpaintingusing,
      title={Towards Coherent Image Inpainting Using Denoising Diffusion Implicit Models},
      author={Guanhua Zhang and Jiabao Ji and Yang Zhang and Mo Yu and Tommi Jaakkola and Shiyu Chang},
      year={2023},
      booktitle=ICML,
}

@inproceedings{rout2023solvinglinearinverseproblems,
      title={Solving Linear Inverse Problems Provably via Posterior Sampling with Latent Diffusion Models}, 
      author={Litu Rout and Negin Raoof and Giannis Daras and Constantine Caramanis and Alexandros G. Dimakis and Sanjay Shakkottai},
booktitle=NIPS,
      year={2023},
}

@inproceedings{
wang2023zeroshot,
title={Zero-Shot Image Restoration Using Denoising Diffusion Null-Space Model},
author={Yinhuai Wang and Jiwen Yu and Jian Zhang},
booktitle=ICLR,
year={2023},
}

@article{sadat2025guidancefrequencydomainenables,
      title={Guidance in the Frequency Domain Enables High-Fidelity Sampling at Low CFG Scales}, 
      author={Seyedmorteza Sadat and Tobias Vontobel and Farnood Salehi and Romann M. Weber},
      year={2025},
journal=arxiv,
}

@InProceedings{Gao_2025_ICCV,
    author    = {Gao, Zheng and Song, Jifei and Zhang, Zhensong and Deng, Jiankang and Patras, Ioannis},
    title     = {Frequency-Guided Diffusion for Training-Free Text-Driven Image Translation},
    booktitle = ICCV,
    year      = {2025},
}

@inproceedings{yu2023freedomtrainingfreeenergyguidedconditional,
      title={FreeDoM: Training-Free Energy-Guided Conditional Diffusion Model}, 
      author={Jiwen Yu and Yinhuai Wang and Chen Zhao and Bernard Ghanem and Jian Zhang},
      year={2023},
      booktitle=ICCV,
}

@article{wang2004ssim,
author = {Wang, Zhou and Bovik, Alan and Sheikh, Hamid and Simoncelli, Eero},
year = {2004},
month = {05},
title = {Image Quality Assessment: From Error Visibility to Structural Similarity},
volume = {13},
journal = TIP,
}

@inproceedings{zhang2018unreasonableeffectivenessdeepfeatures,
  title={The unreasonable effectiveness of deep features as a perceptual metric},
  author={Zhang, Richard and Isola, Phillip and Efros, Alexei A and Shechtman, Eli and Wang, Oliver},
  booktitle=CVPR,
  year={2018}
}

@inproceedings{NIPS2017_8a1d6947,
 author = {Heusel, Martin and Ramsauer, Hubert and Unterthiner, Thomas and Nessler, Bernhard and Hochreiter, Sepp},
 booktitle = NIPS,
 title = {GANs Trained by a Two Time-Scale Update Rule Converge to a Local Nash Equilibrium},
 year = {2017}
}

@article{wu2023humanpreferencescorev2,
      title={Human Preference Score v2: A Solid Benchmark for Evaluating Human Preferences of Text-to-Image Synthesis}, 
      author={Xiaoshi Wu and Yiming Hao and Keqiang Sun and Yixiong Chen and Feng Zhu and Rui Zhao and Hongsheng Li},
      year={2023},
journal=arxiv,
}

@inproceedings{radford2021clip,
  title={Learning Transferable Visual Models From Natural Language Supervision},
  author={Radford, Alec and Kim, Jong Wook and Hallacy, Chris and Ramesh, Aditya and Goh, Gabriel and Agarwal, Sandhini and Sastry, Girish and Askell, Amanda and Mishkin, Pamela and Clark, Jack and Krueger, Gretchen and Sutskever, Ilya},
  booktitle=ICML,
  year={2021}
}

@misc{claude-sonnet-4-5,
  title = {Claude Sonnet 4.5},
  author = {{Anthropic}},
  year = {2025},
  howpublished = {\url{https://www.anthropic.com/news/claude-sonnet-4-5}},
  note = {Accessed: 2026-06-30},
}

@inproceedings{kingma2017adammethodstochasticoptimization,
      title={Adam: A Method for Stochastic Optimization},
      author={Diederik P. Kingma and Jimmy Ba},
      year={2015},
      booktitle=ICLR,
}

@inproceedings{
podell2024sdxl,
title={{SDXL}: Improving Latent Diffusion Models for High-Resolution Image Synthesis},
author={Dustin Podell and Zion English and Kyle Lacey and Andreas Blattmann and Tim Dockhorn and Jonas M{\"u}ller and Joe Penna and Robin Rombach},
booktitle=ICLR,
year={2024}
}

@inproceedings{gong2026freeinpaint,
  title={FreeInpaint: Tuning-free Prompt Alignment and Visual Rationality Enhancement in Image Inpainting},
  author={Gong, Chao and Li, Dong and Pan, Yingwei and Chen, Jingjing and Yao, Ting and Mei, Tao},
  booktitle={Proceedings of the AAAI Conference on Artificial Intelligence},
  volume={40},
  number={6},
  pages={4239--4247},
  year={2026}
}

@misc{alimama2024sd3controlnet,
  author = {{Alimama Creative}},
  title = {{SD3-ControlNet-Inpainting}},
  year = {2024},
  howpublished = {\url{https://huggingface.co/alimama-creative/SD3-Controlnet-Inpainting}},
  note = {Accessed: 2026-06-29},
}

@inproceedings{xu2023imagerewardlearningevaluatinghuman,
      title={ImageReward: Learning and Evaluating Human Preferences for Text-to-Image Generation}, 
      author={Jiazheng Xu and Xiao Liu and Yuchen Wu and Yuxuan Tong and Qinkai Li and Ming Ding and Jie Tang and Yuxiao Dong},
      year={2023},
booktitle=NIPS,
}

@inproceedings{10.5555/3600270.3602103,
author = {Schuhmann, Christoph and Beaumont, Romain and Vencu, Richard and Gordon, Cade and Wightman, Ross and Cherti, Mehdi and Coombes, Theo and Katta, Aarush and Mullis, Clayton and Wortsman, Mitchell and Schramowski, Patrick and Kundurthy, Srivatsa and Crowson, Katherine and Schmidt, Ludwig and Kaczmarczyk, Robert and Jitsev, Jenia},
title = {LAION-5B: an open large-scale dataset for training next generation image-text models},
year = {2022},
isbn = {9781713871088},
booktitle = NeurIPS,
articleno = {1833}
}

@article{lyu2025diff,
  title={IS-Diff: Improving Diffusion-Based Inpainting with Better Initial Seed},
  author={Lyu, Yongzhe and Wu, Yu and Lin, Yutian and Du, Bo},
  journal=arxiv,
  year={2025}
}

@inproceedings{chen2024improving,
  title={Improving text-guided object inpainting with semantic pre-inpainting},
  author={Chen, Yifu and Chen, Jingwen and Pan, Yingwei and Li, Yehao and Yao, Ting and Chen, Zhineng and Mei, Tao},
  booktitle=ECCV,
  year={2024},
}

@misc{von-platen-etal-2022-diffusers,
  author    = {Patrick von Platen and Suraj Patil and Anton Lozhkov and Pedro Cuenca and Nathan Lambert and Kashif Rasul and Mishig Davaadorj and Dhruv Nair and Sayak Paul and William Berman and Yiyi Xu and Steven Liu and Thomas Wolf},
  title     = {Diffusers: State-of-the-art diffusion models},
  year      = {2022},
  publisher = {GitHub},
  journal   = {GitHub repository},
  howpublished = {\url{https://github.com/huggingface/diffusers}},
  note      = {Accessed: 2026-06-30}
}

@article{shamsolmoali2025missing,
  title={From Missing Pieces to Masterpieces: Image Completion with Context-Adaptive Diffusion},
  author={Shamsolmoali, Pourya and Zareapoor, Masoumeh and Zhou, Huiyu and Felsberg, Michael and Tao, Dacheng and Li, Xuelong},
  journal=TPAMI,
  year={2025},
}

@inproceedings{liu2025corrfillenhancingfaithfulnessreferencebased,
      title={CorrFill: Enhancing Faithfulness in Reference-based Inpainting with Correspondence Guidance in Diffusion Models}, 
      author={Kuan-Hung Liu and Cheng-Kun Yang and Min-Hung Chen and Yu-Lun Liu and Yen-Yu Lin},
      year={2025},
booktitle=WACV,
}

@article{Ronai2025FlowOpt,
  author       = {Or Ronai and Vladimir Kulikov and Tomer Michaeli},
  title        = {FlowOpt: Fast Optimization Through Whole Flow Processes for Training-Free Editing},
  journal = arxiv,
  year         = {2025}}

@article{wan2025,
      title={Wan: Open and Advanced Large-Scale Video Generative Models},
      author={{Team Wan} and Ang Wang and Baole Ai and Bin Wen and Chaojie Mao and Chen-Wei Xie and Di Chen and Feiwu Yu and Haiming Zhao and Jianxiao Yang and Jianyuan Zeng and Jiayu Wang and Jingfeng Zhang and Jingren Zhou and Jinkai Wang and Jixuan Chen and Kai Zhu and Kang Zhao and Keyu Yan and Lianghua Huang and Mengyang Feng and Ningyi Zhang and Pandeng Li and Pingyu Wu and Ruihang Chu and Ruili Feng and Shiwei Zhang and Siyang Sun and Tao Fang and Tianxing Wang and Tianyi Gui and Tingyu Weng and Tong Shen and Wei Lin and Wei Wang and Wei Wang and Wenmeng Zhou and Wente Wang and Wenting Shen and Wenyuan Yu and Xianzhong Shi and Xiaoming Huang and Xin Xu and Yan Kou and Yangyu Lv and Yifei Li and Yijing Liu and Yiming Wang and Yingya Zhang and Yitong Huang and Yong Li and You Wu and Yu Liu and Yulin Pan and Yun Zheng and Yuntao Hong and Yupeng Shi and Yutong Feng and Zeyinzi Jiang and Zhen Han and Zhi-Fan Wu and Ziyu Liu},
      journal=arxiv,
      year={2025},
}
}

\clearpage
\appendix
\raggedbottom
\setlength{\textfloatsep}{5mm plus 1pt minus 1pt}
\setlength{\intextsep}{4mm plus 1pt minus 1pt}
  {
   \newpage
        \centering

        {
        \Large
        \bf
        SONIC: Spectral Optimization of Noise for \\ Inpainting with Consistency
        }
        
        {
        \vspace{0.5em}Supplementary Material \\
        }
        \vspace{2.5em}
   }

In this supplementary, we provide further analysis, examples, and data excluded from the main paper due to spatial limits.
We first present a user study in \cref{sec:supp_userstudy}, followed by
an illustration of the linearity of \sdthreefive denoising trajectories in \cref{sec:supp_linearity}.
We then further discuss the convergence details in \cref{fig:optimization_sweep} in \cref{sec:supp_spectral_vs_spatial} and provide a qualitative example of applying our method to video models in \cref{sec:supp_video}.
To show the general applicability of our method, we apply our method to non-flow models in \cref{sec:non-flow}.
For further grounding, we provide a theoretical analysis in \cref{sec:supp_theory}.
We include a wall-clock comparison of all methods in \cref{sec:supp_wall_clock}, comparisons against state-of-the-art noise optimization methods in \cref{sec:supp_sota-noiseopt-comparison}, a comparison with a trained inpainting baseline in \cref{sec:supp_controlnet}, and additional ablations in \cref{sec:supp_ablations}.
We conclude with the exact prompt settings for reproducing our results in \cref{sec:supp_prompt} and more qualitative results in \cref{sec:supp_extra}.
Note that the self-contained interactive website provides further visual results.

\section{User study}
\label{sec:supp_userstudy}

\begin{figure}[H]
    \centering
    \vspace{-0.5em}
    \includegraphics[width=0.28\linewidth,height=0.46\linewidth]{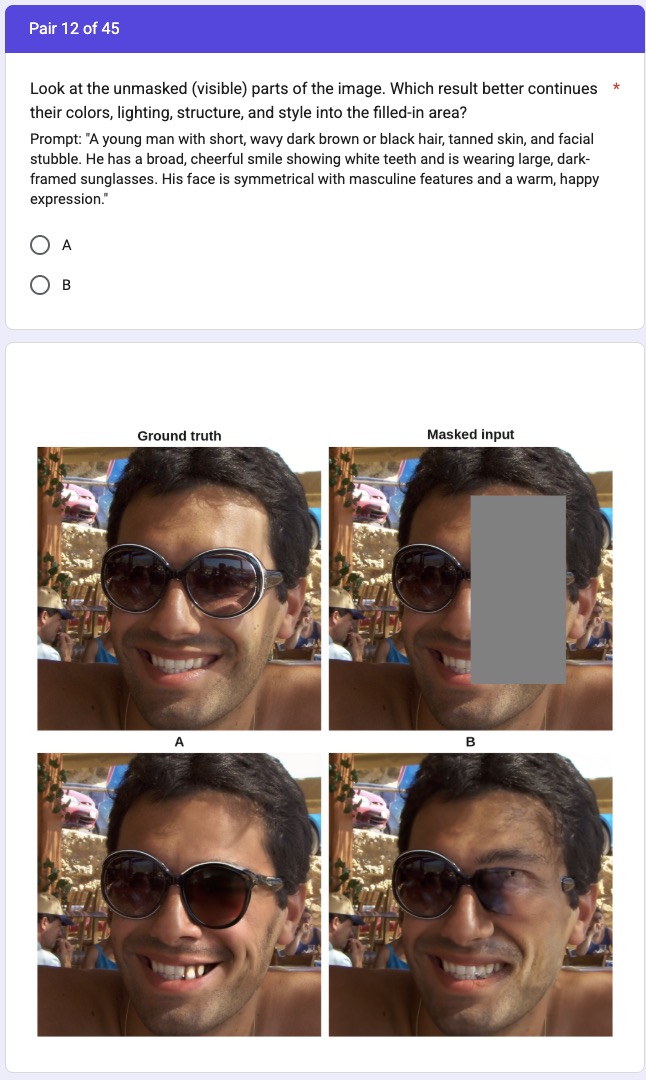}\hspace{0.5em}%
    \includegraphics[width=0.28\linewidth,height=0.46\linewidth]{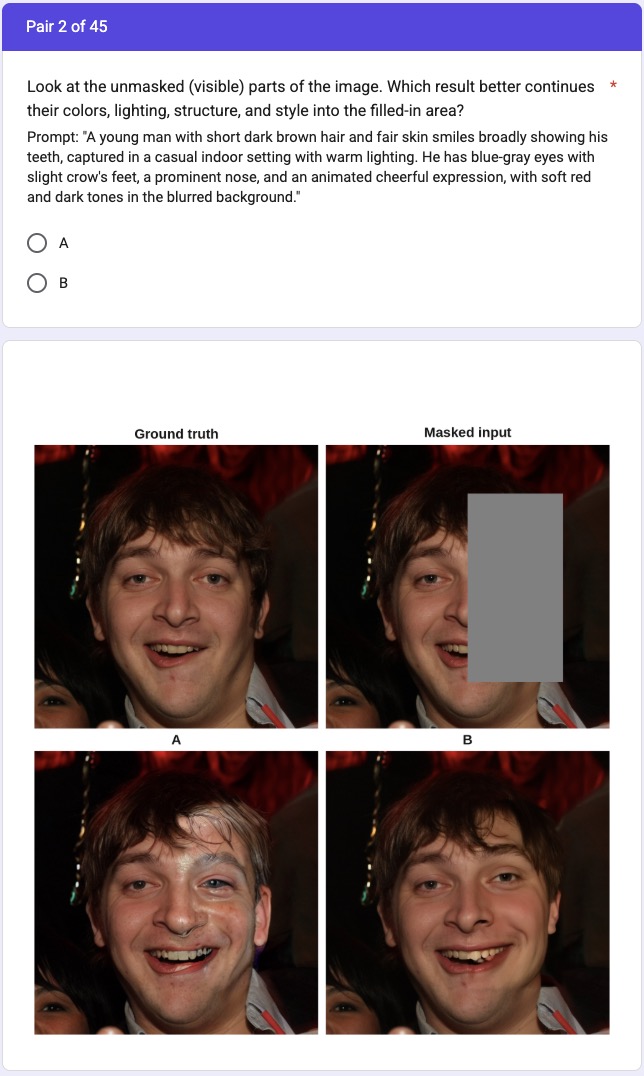}\hspace{0.5em}%
    \includegraphics[width=0.28\linewidth,height=0.46\linewidth]{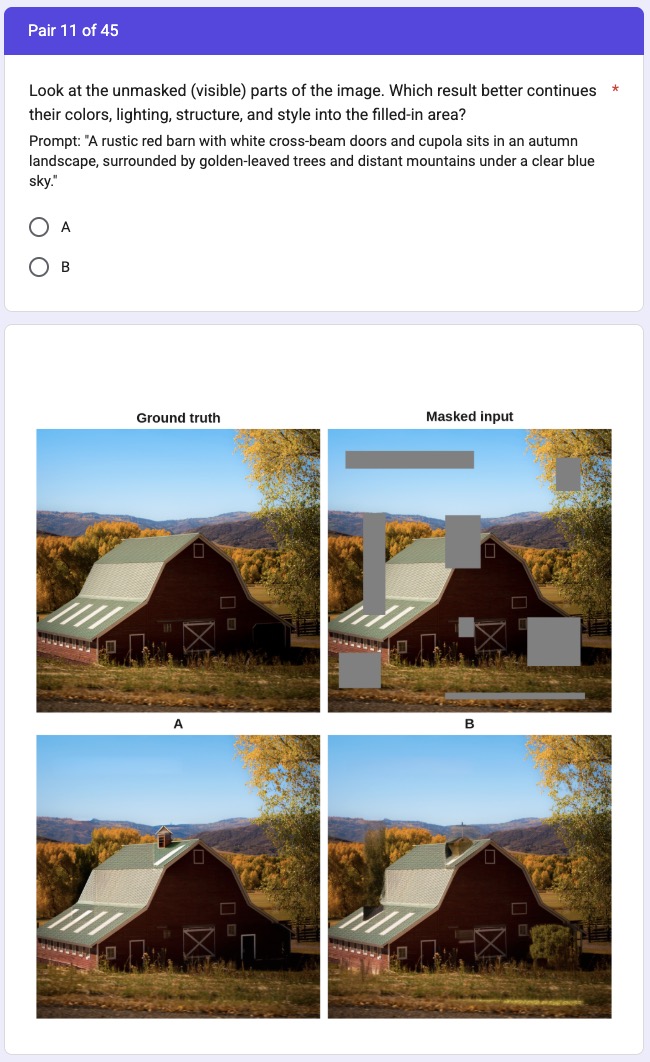}
    \vspace{-0.5em}
    \caption{{\bf Examples of 2AFC user study questions} -- Raters saw the ground truth, masked input and two candidate inpaintings ($A$/$B$), and chose the one that better preserves colors, lighting, structure, and style of the visible region. The placement of our results and the question order were randomized.}
    \label{fig:user_test_example}
    \vspace{-1em}
\end{figure}

We conducted a single-blind two-alternative forced-choice (2AFC) study with $N{=}37$ raters comparing our method against 5 baselines on 45 images spanning 45 scenes randomly sampled from FFHQ, DIV2K, and BrushBench (30/10/5), each scene shown once and each baseline evaluated on 9 pairs.
For every pair, the placement of our method ($A$ or $B$) and the question order were randomized, and all baselines were run at their default NFE.
Raters were asked to choose the image that better preserves the colors, lighting, structure, and style of the visible region. See \cref{fig:user_test_example} for example questions.
Our method was preferred in $\mathbf{90.0\%}$ of comparisons ($1{,}498/1{,}665$; 95\% CI $[88.4,91.4]$), with statistically significant gains over all baselines; full per-baseline breakdown is reported in \cref{tab:userstudy}.

\begin{table}[H]
\centering
\setlength{\tabcolsep}{6pt}
\resizebox{\linewidth}{!}{
\begin{tabular}{@{}lccccc@{}}
\toprule
 & \makecell[t]{vs.\ FLAIR}
 & \makecell[t]{vs.\ BrushNet}
 & \makecell[t]{vs.\ BLD-SD3.5}
 & \makecell[t]{vs.\ FlowChef}
 & \makecell[t]{vs.\ FlowDPS} \\
\midrule
Win-rate$\uparrow$ 
& 80.5\% & 79.9\% & 94.0\% & 99.1\% & 96.4\% \\

95\% Conf. Interv. 
& {\scriptsize[75.9,84.4]} 
& {\scriptsize[75.2,83.8]} 
& {\scriptsize[90.9,96.1]} 
& {\scriptsize[97.4,99.7]} 
& {\scriptsize[93.8,97.9]} \\
\bottomrule
\end{tabular}
}
\vspace{0.5em}
\caption{{\bf 2AFC user study -- } Per-baseline win rate of our method ($N{=}37$ raters, $1{,}665$ pairwise comparisons across FFHQ, DIV2K, and BrushBench). Our method is preferred over every baseline with statistically significant margins.}
\label{tab:userstudy}
\end{table}

\section{Linearity of \sdthreefive denoising trajectories}
\label{sec:supp_linearity}

\begin{figure}[H]
\centering
\includegraphics[width=0.95\linewidth]{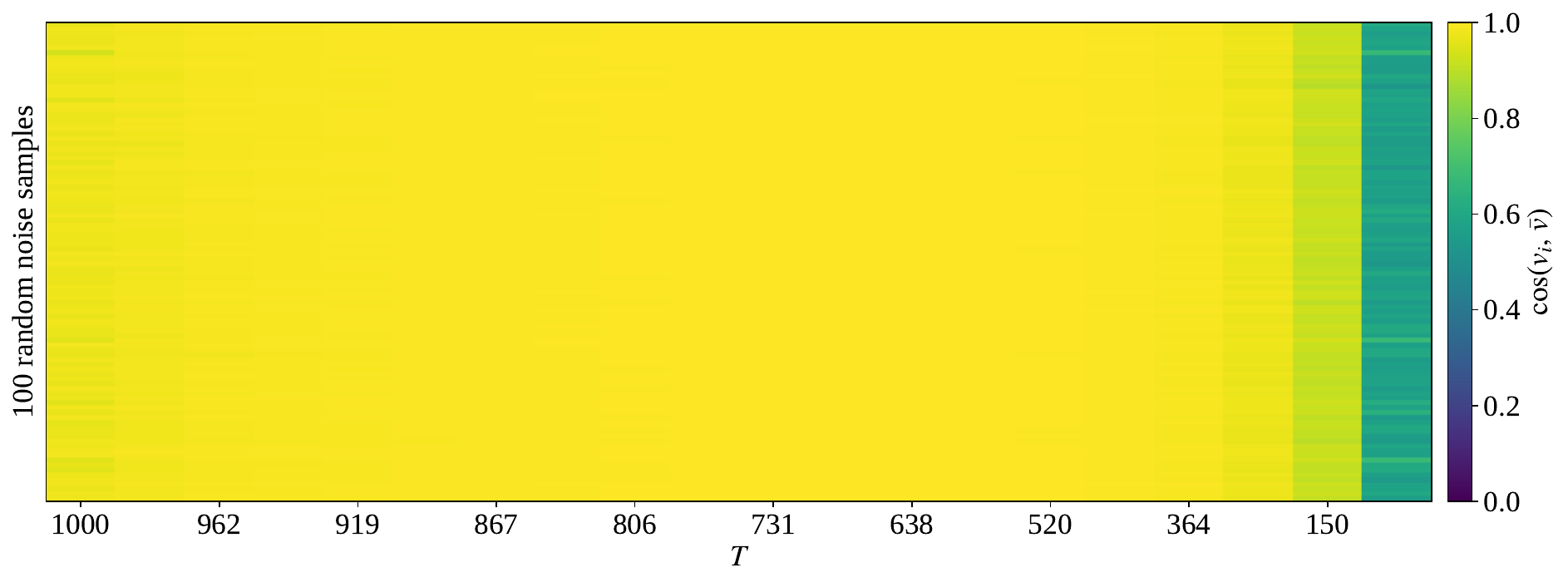}
    \vspace{-1em}
    \caption{%
      {\bf Velocity linearity --} We plot the similarity between the velocity from each time step and the average velocity value of all time steps---if the trajectory is entirely linear, these should be identical.
      Trajectory is mostly linear with the exception of only a few final denoising steps.
      }%
    \label{fig:linearity-velocity}
\vspace{-1em}
\end{figure}

To illustrate the linearity of the denoising trajectory of a flow model (\sdthreefive), we visualize the cosine similarity between the predicted velocities of each time step versus the average velocity in \cref{fig:linearity-velocity}.
We use 100 randomly drawn noise samples, a fixed prompt,  $T=20$, and a classifier-free guidance scale of 2.0. 
For an ideal linear trajectory, the individual velocities would be identical to the average velocity.
Note how the trajectory is almost always linear except for the final denoising steps, which are when the latent is nearly noiseless and can be safely omitted when considering the initial noise sample.
This is further supported by our empirical results in the main paper, which strongly suggest that linear approximation is sufficient.

\enlargethispage{10\baselineskip}

\begin{figure}[H]
\centering
\begin{minipage}[c]{0.48\linewidth}
    \centering
\includegraphics[width=0.85\linewidth]{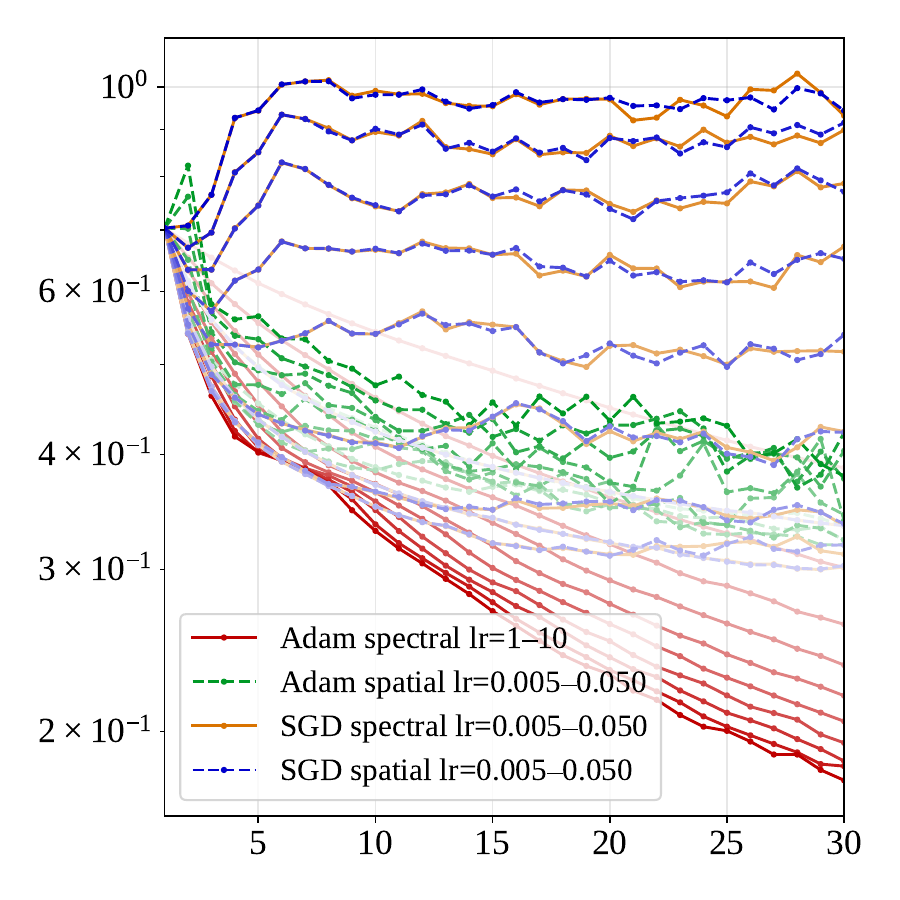}
    \vspace{-1em}
    \caption{%
      {\bf Convergence of denoised latent $x_0$} -- Optimization trajectory with varying optimizer and learning rate setups, mapping the average root mean-squared error (RMSE) between the \textbf{denoised latents}.
      }%
    \label{fig:x0_graph_suppl}
\end{minipage}
\hfill
\begin{minipage}[c]{0.48\linewidth}
    \centering
    \includegraphics[width=0.85\linewidth]{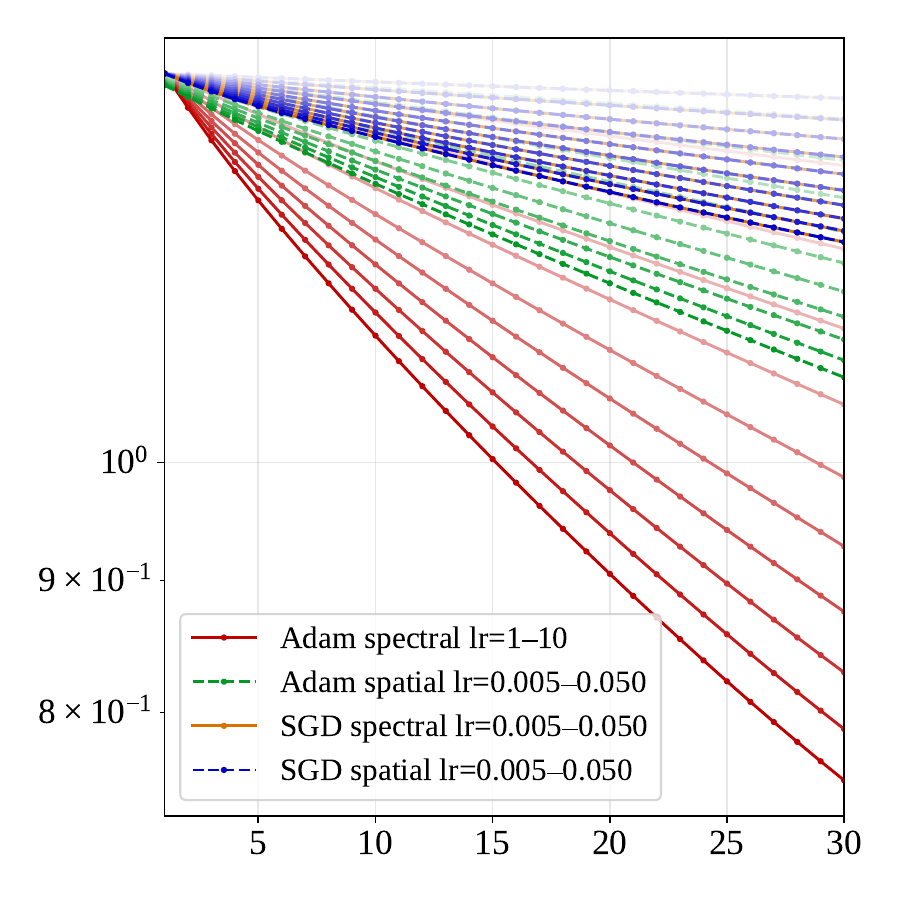}
    \vspace{-1em}
    \caption{%
      {\bf Convergence of denoised latent $X_T$} -- Optimization trajectory with varying optimizer and learning rate setups, mapping the average root mean-squared error (RMSE) between the \textbf{initial noise samples}.
      }
    \label{fig:epsilon_graph}
\end{minipage}
\vspace{-1em}
\end{figure}

\section{Comparing spectral vs. spatial optimization}
\label{sec:supp_spectral_vs_spatial}

We graph the optimization trajectory of the denoised latent $x_0$ in \cref{fig:x0_graph_suppl} (same graph as~\cref{fig:optimization_sweep}), and the optimization trajectory of the initial noise $X_T$ in \cref{fig:epsilon_graph}.
With Adam~\cite{kingma2017adammethodstochasticoptimization} spatial optimization, though the optimization with a high-enough learning rate may reduce the initial noise sample loss, it is not able to converge to a valid denoised outcome. 

\begin{figure}[H]
    \centering
    \newcommand{\imwidth}{0.187\linewidth}
    \setlength{\tabcolsep}{1pt}
    \renewcommand{\arraystretch}{0.85}
    \begin{tabular}{c ccccc}
        {\rotatebox[origin=l]{90}{\parbox[b]{2.2cm}{\centering\tiny Low LR \\ Adam Spatial}}} &
        \includegraphics[width=\imwidth]{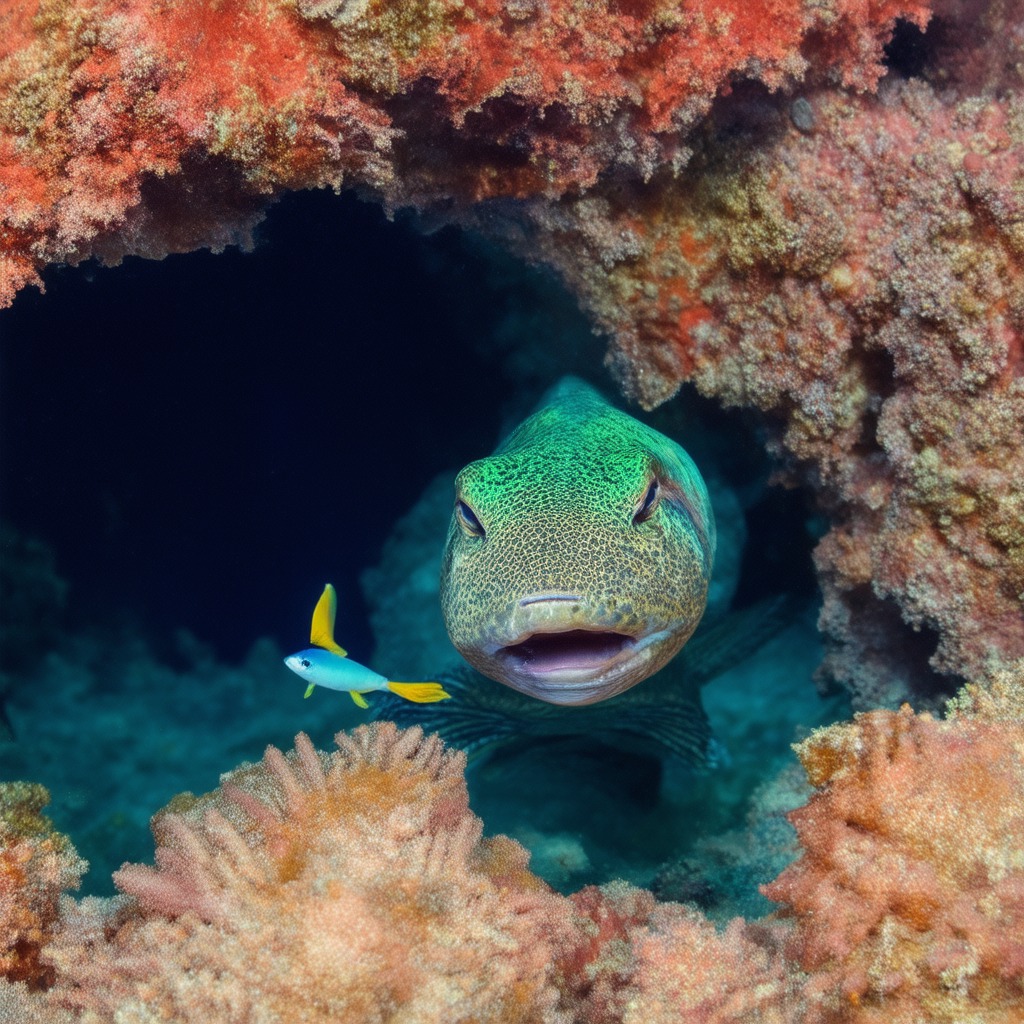} &
        \includegraphics[width=\imwidth]{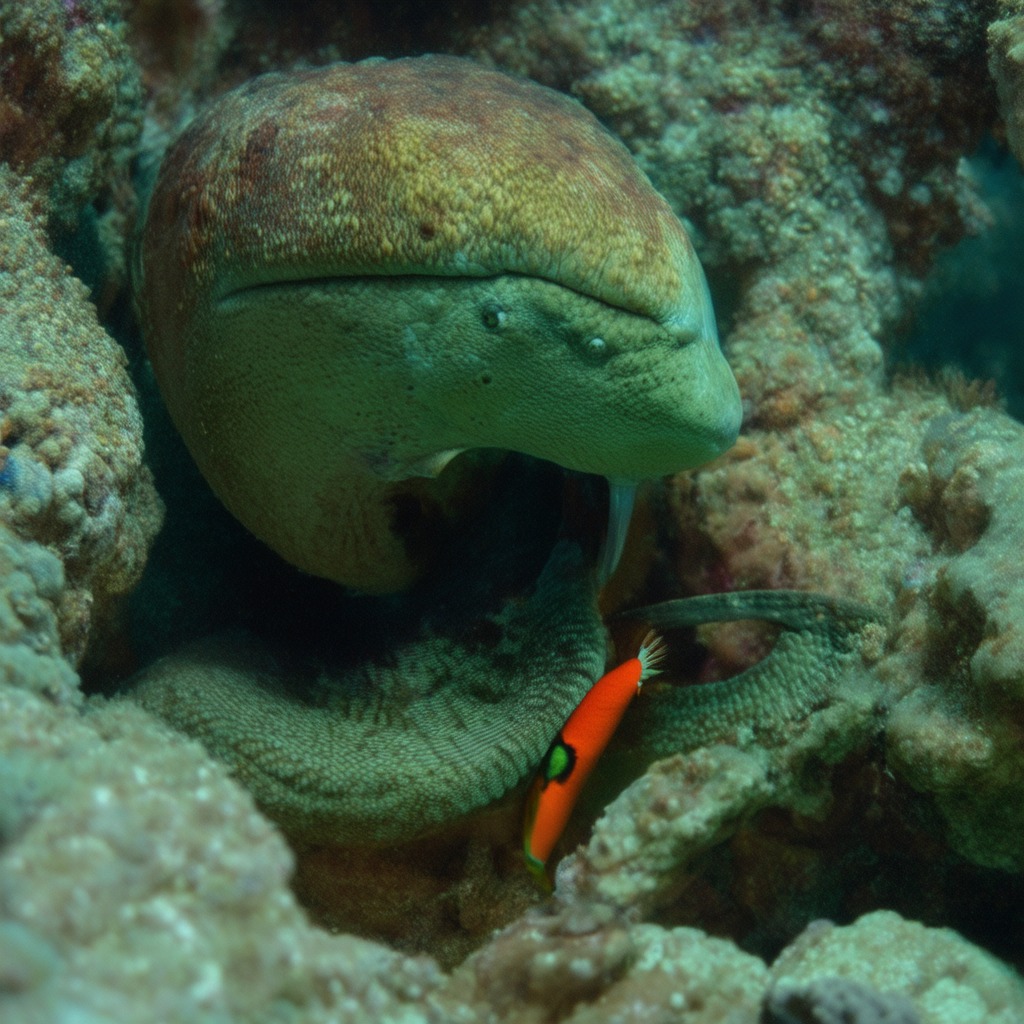} &
        \includegraphics[width=\imwidth]{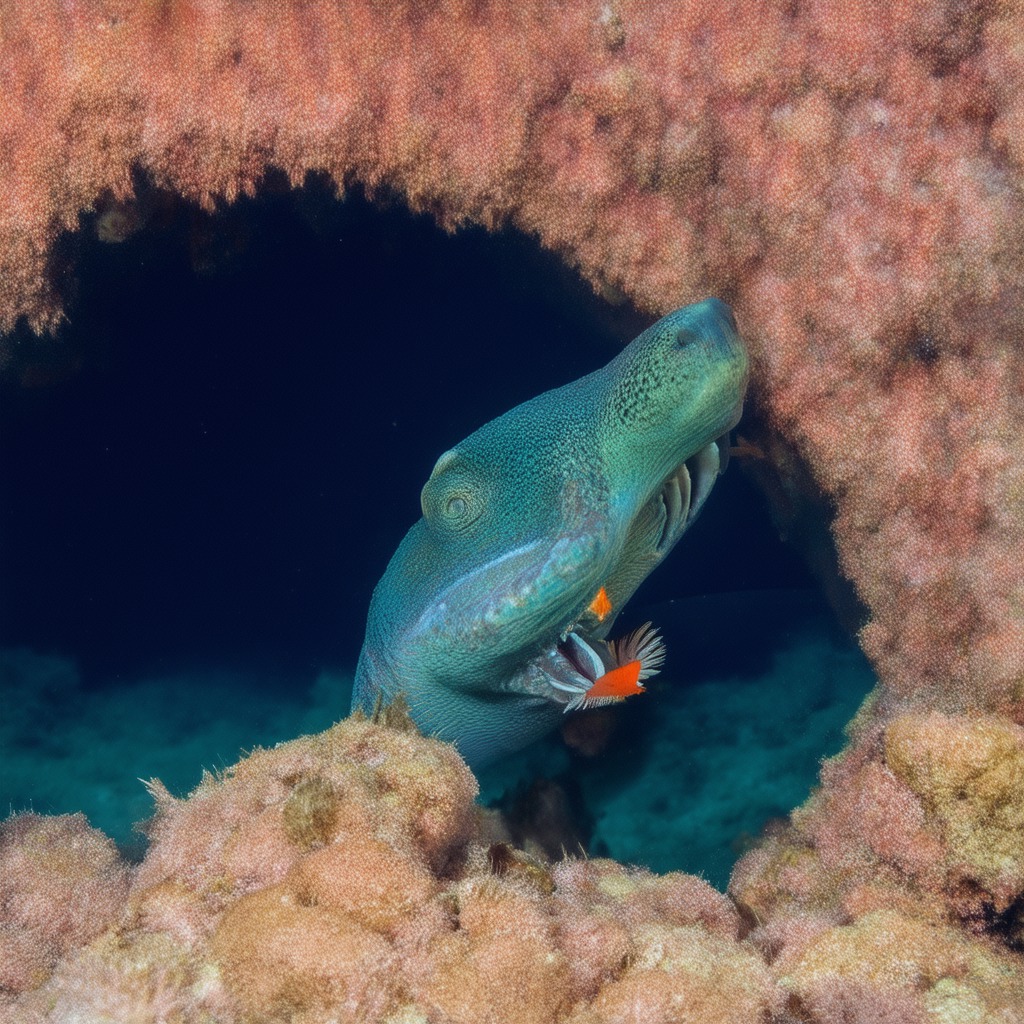} &
        \includegraphics[width=\imwidth]{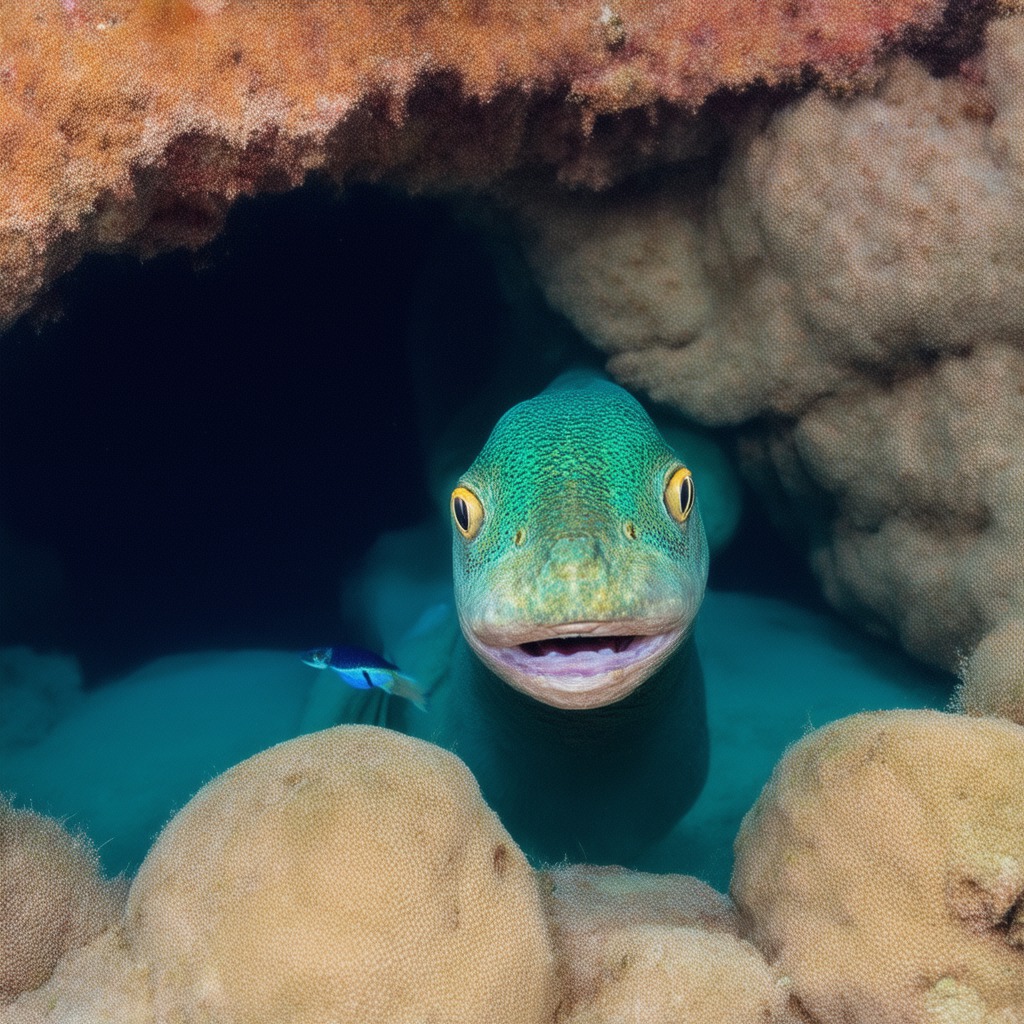} &
        \includegraphics[width=\imwidth]{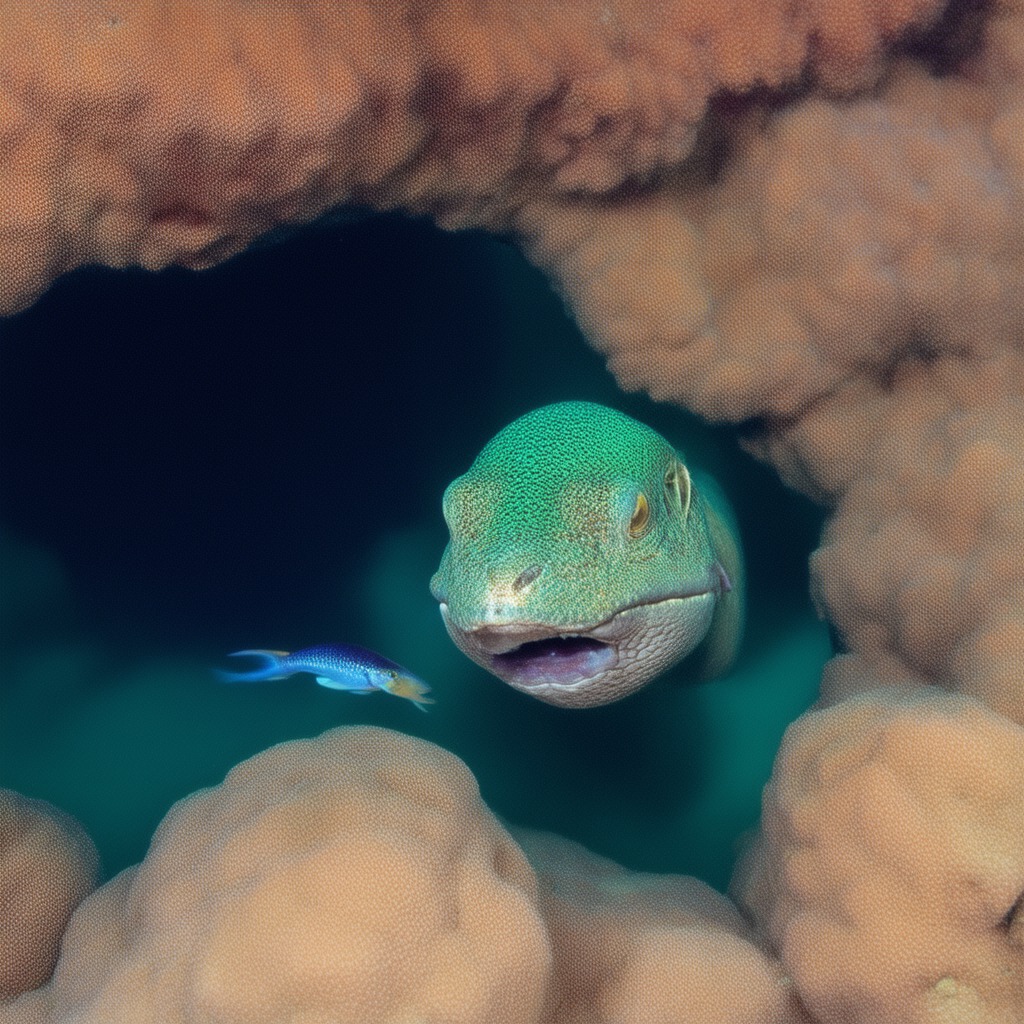} \\
        {\rotatebox[origin=l]{90}{\parbox[b]{2.2cm}{\centering\tiny High LR \\ Adam Spatial}}} &
        \includegraphics[width=\imwidth]{figures/suppl_fig/qualitative/00098.jpg} &
        \includegraphics[width=\imwidth]{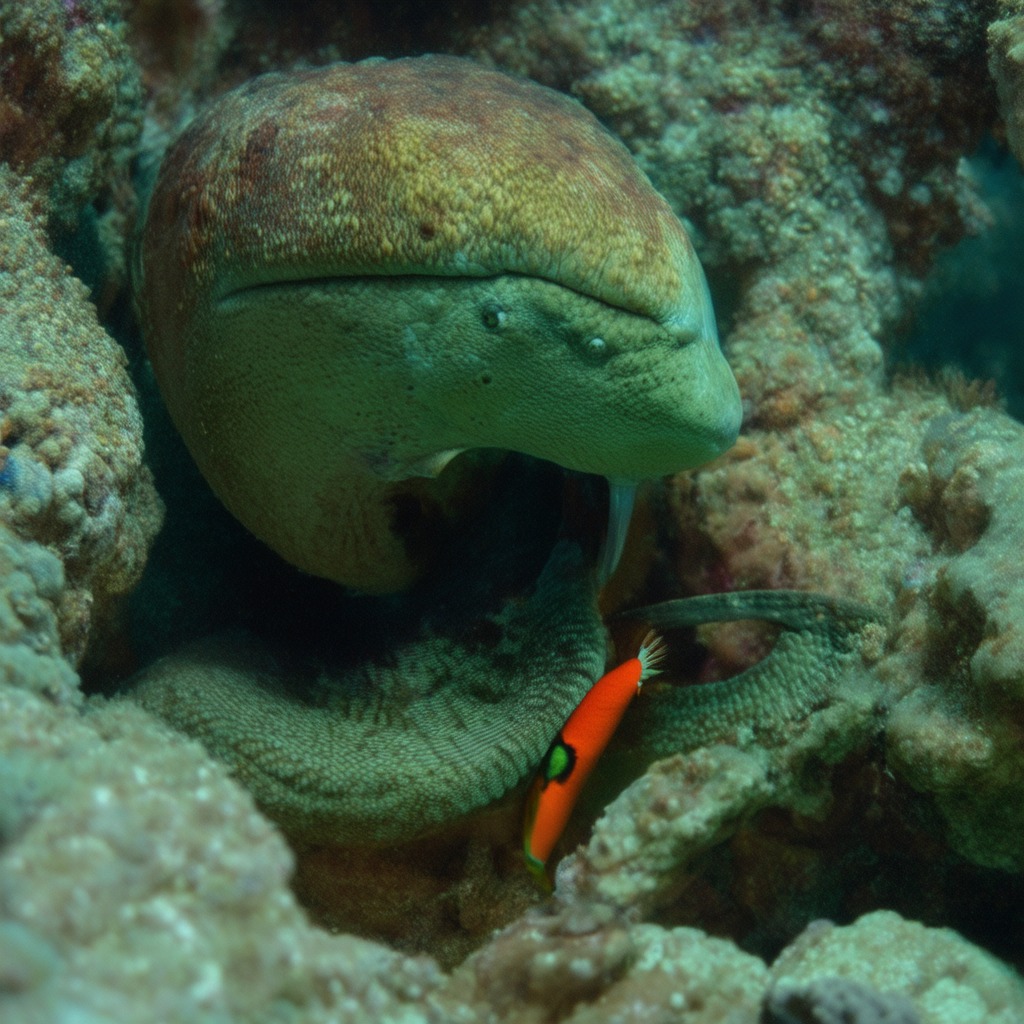} &
        \includegraphics[width=\imwidth]{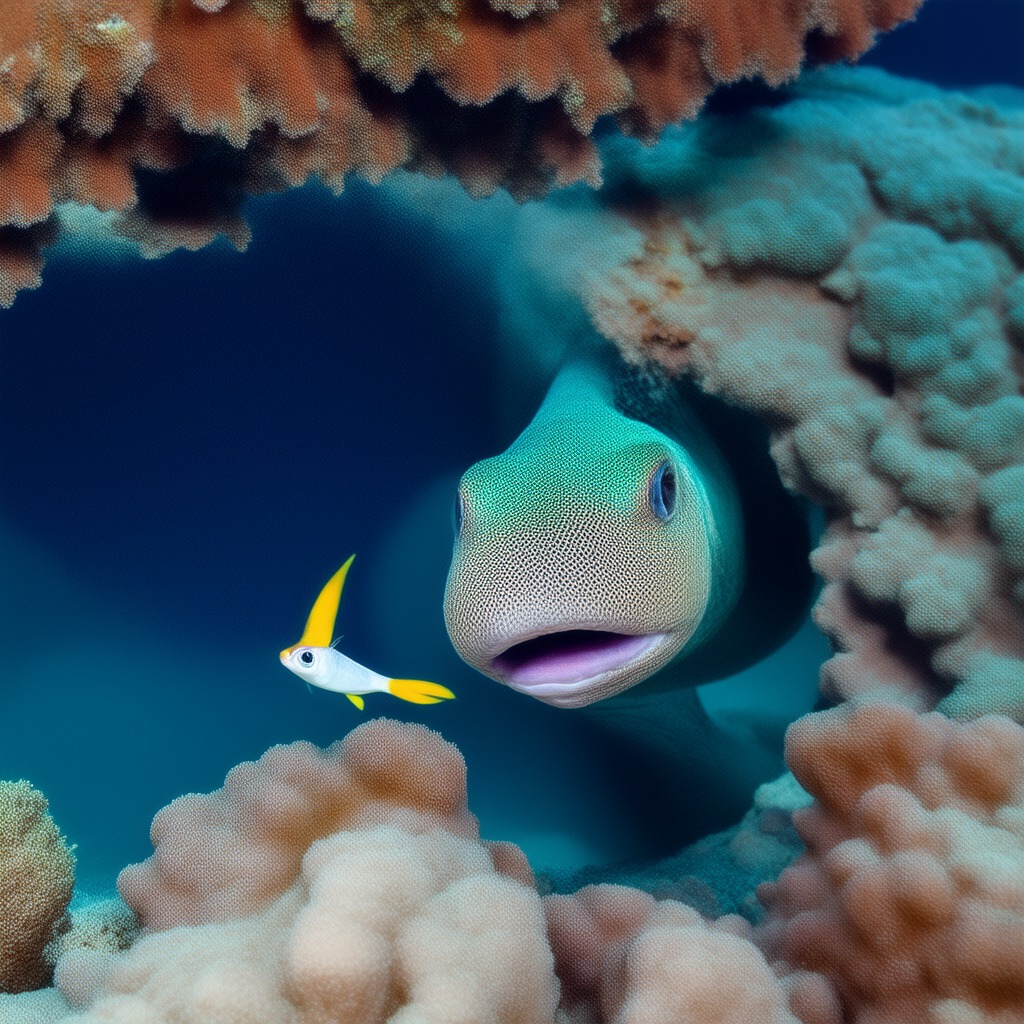} &
        \includegraphics[width=\imwidth]{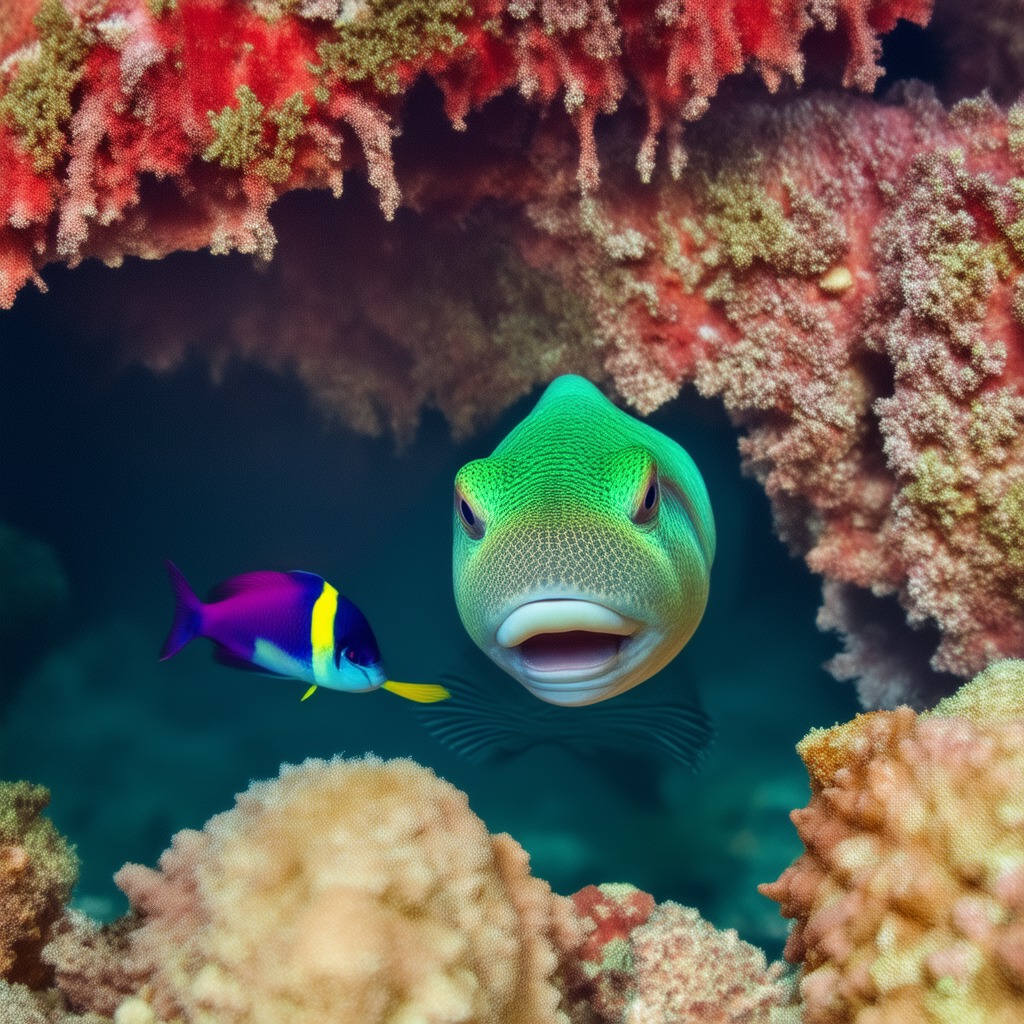} &
        \includegraphics[width=\imwidth]{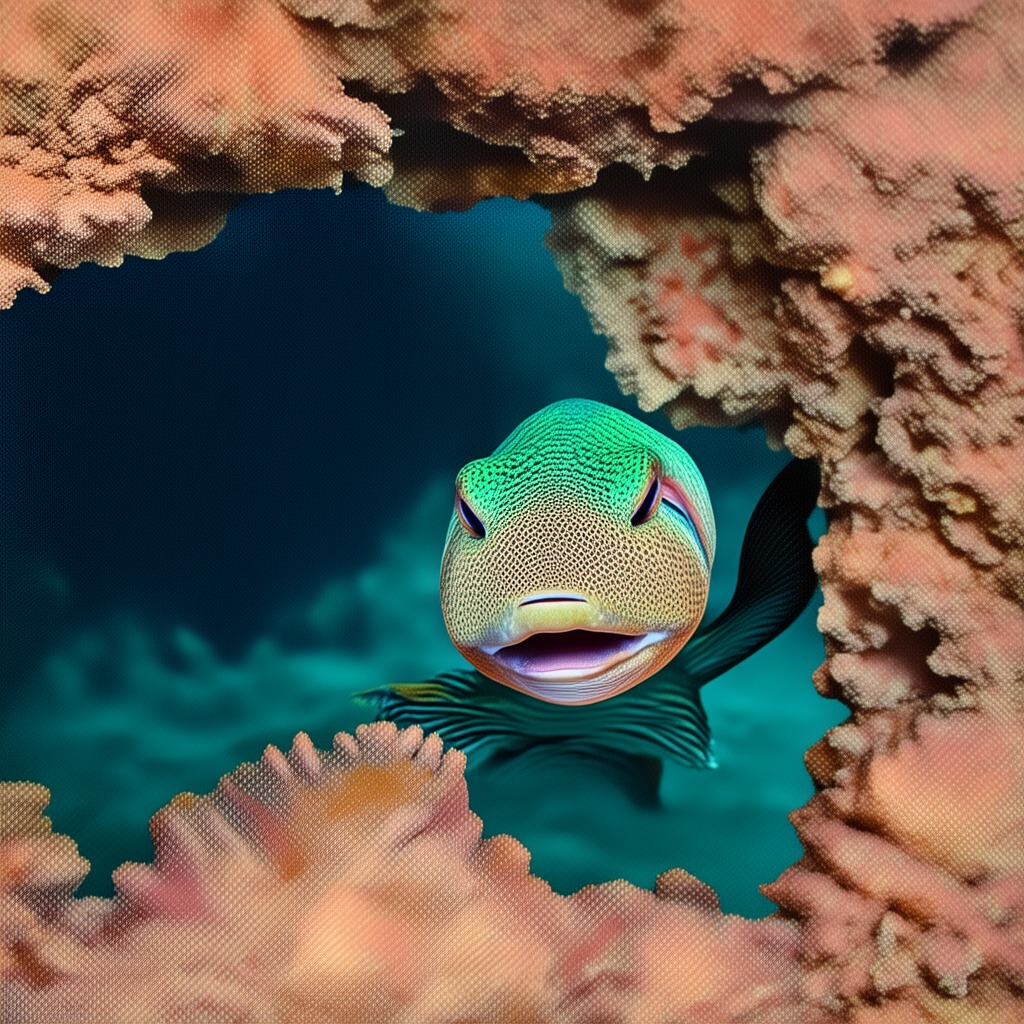} \\
        {\rotatebox[origin=l]{90}{\parbox[b]{2.2cm}{\centering\tiny Adam Spectral}}} &
        \includegraphics[width=\imwidth]{figures/suppl_fig/qualitative/00098.jpg} &
        \includegraphics[width=\imwidth]{figures/suppl_fig/qualitative/spatial_lowlr/001.jpg} &
        \includegraphics[width=\imwidth]{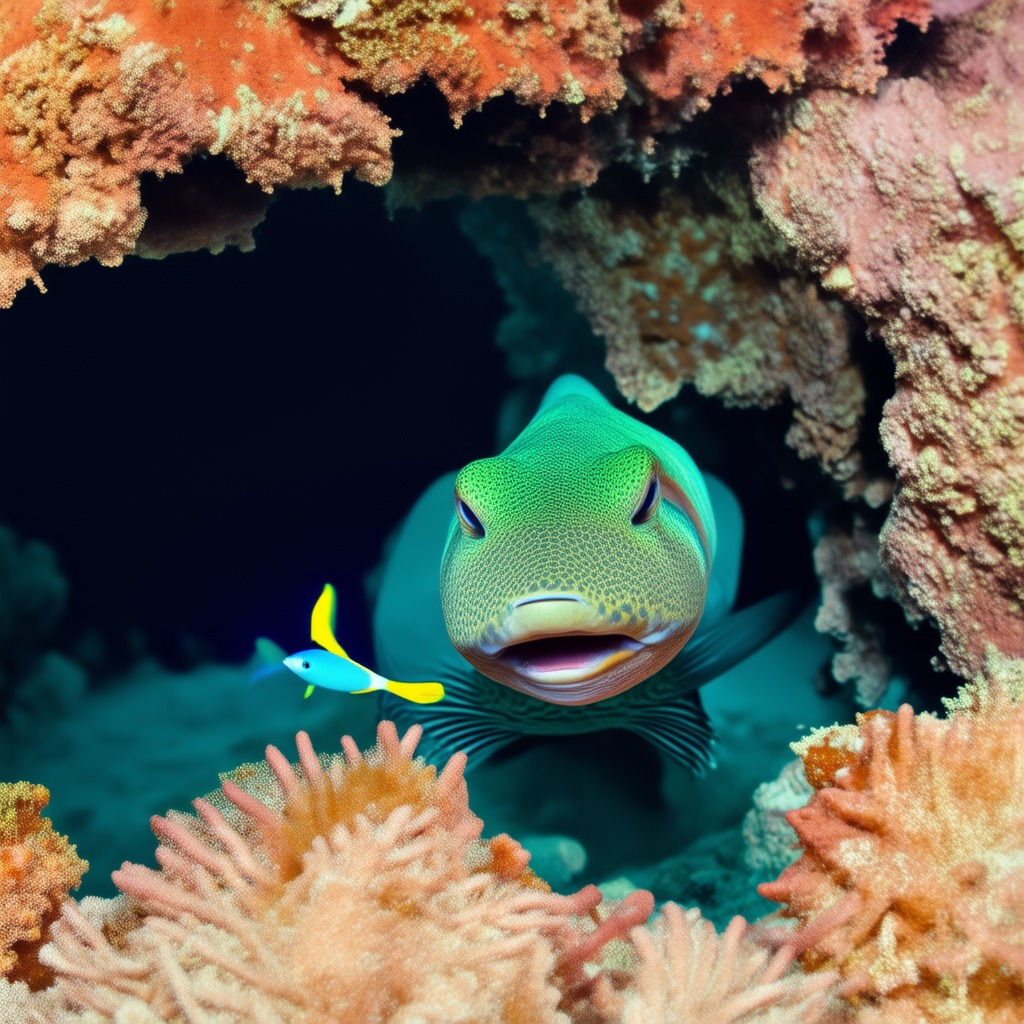} &
        \includegraphics[width=\imwidth]{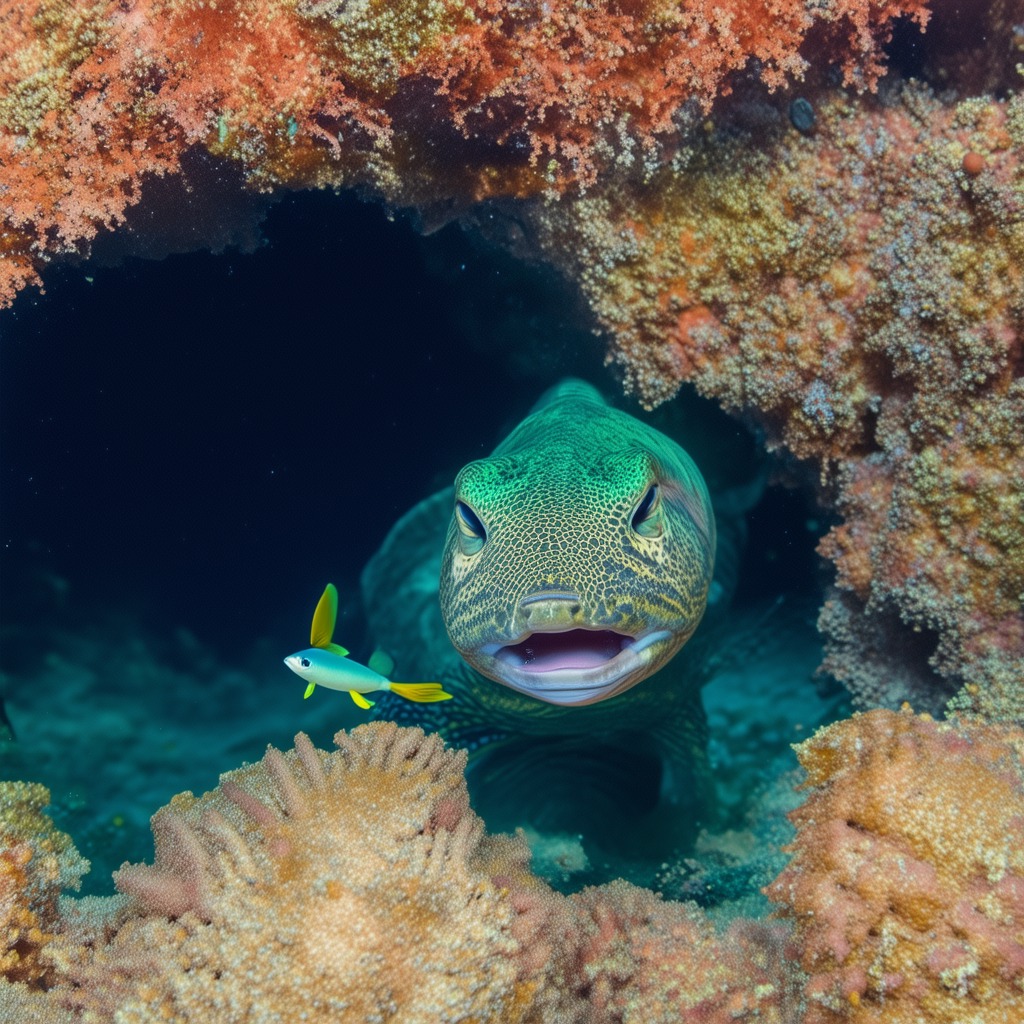} &
        \includegraphics[width=\imwidth]{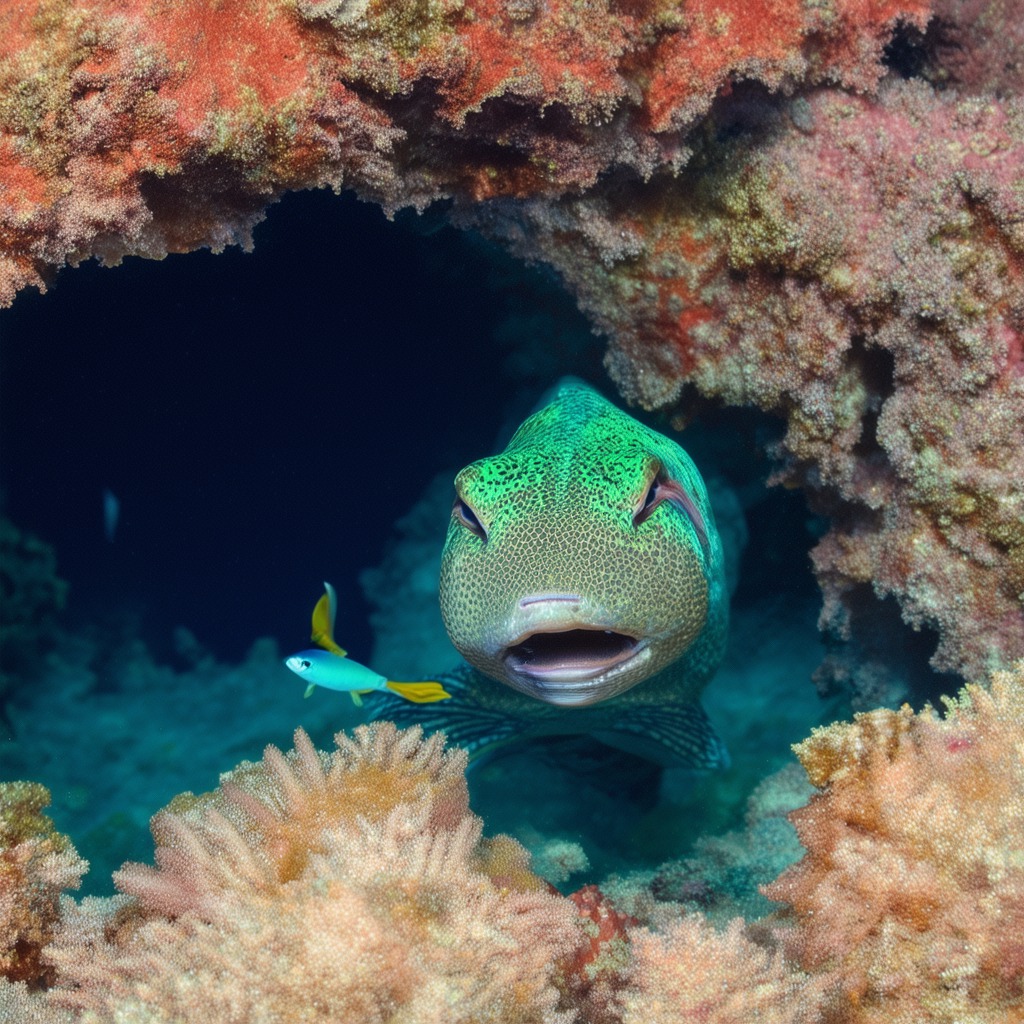} \\
        &
        {\tiny \makecell{Ground Truth}} & {\tiny \bf Step 0} & {\tiny \bf Step 10} & {\tiny \bf Step 20} & {\tiny \bf Step 30}
    \end{tabular}
    \renewcommand{\arraystretch}{1.0}
    \setlength{\tabcolsep}{6pt}
    \vspace{-0.5em}
    \caption{
    {\bf 
        Qualitative comparisons between varying optimization setups
        } -- 
        Our spectral preconditioning delivers stable and enhanced convergence.
    }
    \label{fig:qualitative_optimization_comparison}
    \vspace{-1em}
\end{figure}

In~\cref{fig:qualitative_optimization_comparison}, we include qualitative comparisons between denoised outcomes using varying optimization setups for a randomly selected sample. 
Though the optimization trajectory may appear stable with a low learning rate with spatial Adam optimization, it is unable to reconstruct fine details in the denoised image. 
Increasing the learning rate causes erratic optimization trajectories, leading to over-saturation, errors in the reconstruction, or a blurry and dotted outcome.

\enlargethispage{6\baselineskip}

\section{Applying our method to video models}
\label{sec:supp_video}

Optimizing the initial noise sample in the spectral domain can be done with video generative models such as Wan~\cite{wan2025}, to perform inpainting. To mask the latent in the temporal space, we follow the encoder's logic---first masking the temporal layer of the latent with the first mask frame, then masking consecutive layers with the maxpool of 4 mask frames. 
Please see the interactive result on the project website.

\section{Applying our method to non-flow models}
\label{sec:non-flow}

\begin{figure}[H]
    \centering
    \newcommand{\imwidth}{0.185\linewidth}
    \newcommand{\oursimg}[1]{{\setlength{\fboxsep}{0pt}\setlength{\fboxrule}{2pt}\fcolorbox{red}{white}{#1}}}
    \setlength{\tabcolsep}{1pt}
    \renewcommand{\arraystretch}{0.85}
    \begin{tabular}{ccccc}
        \includegraphics[width=\imwidth]{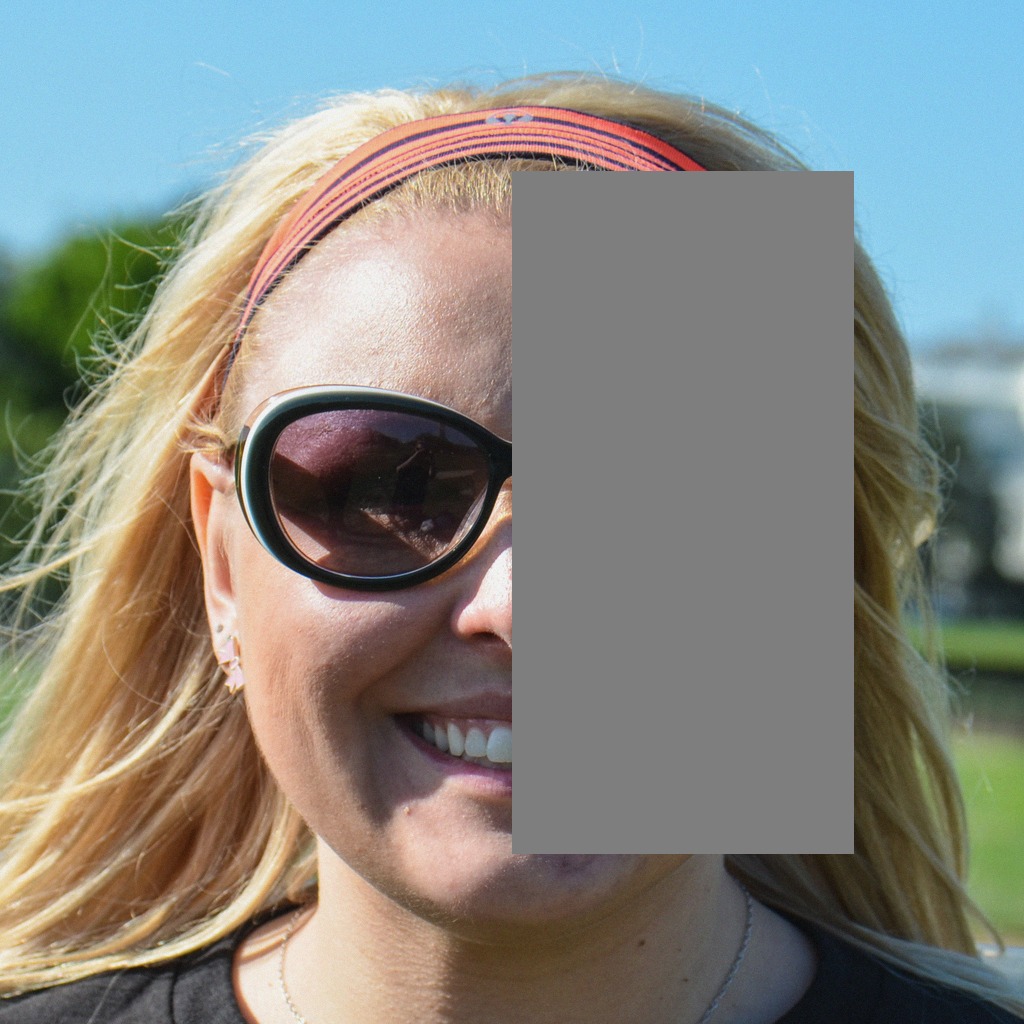} &
        \includegraphics[width=\imwidth]{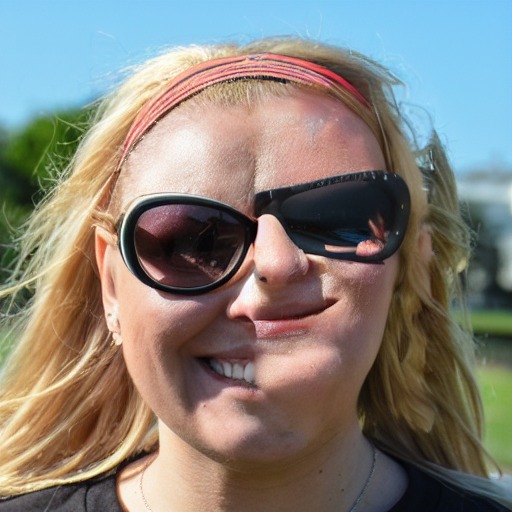} &
        \oursimg{\includegraphics[width=\imwidth]{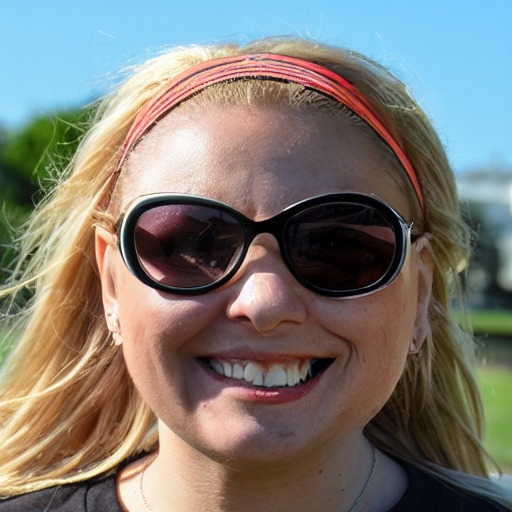}} &
        \includegraphics[width=\imwidth]{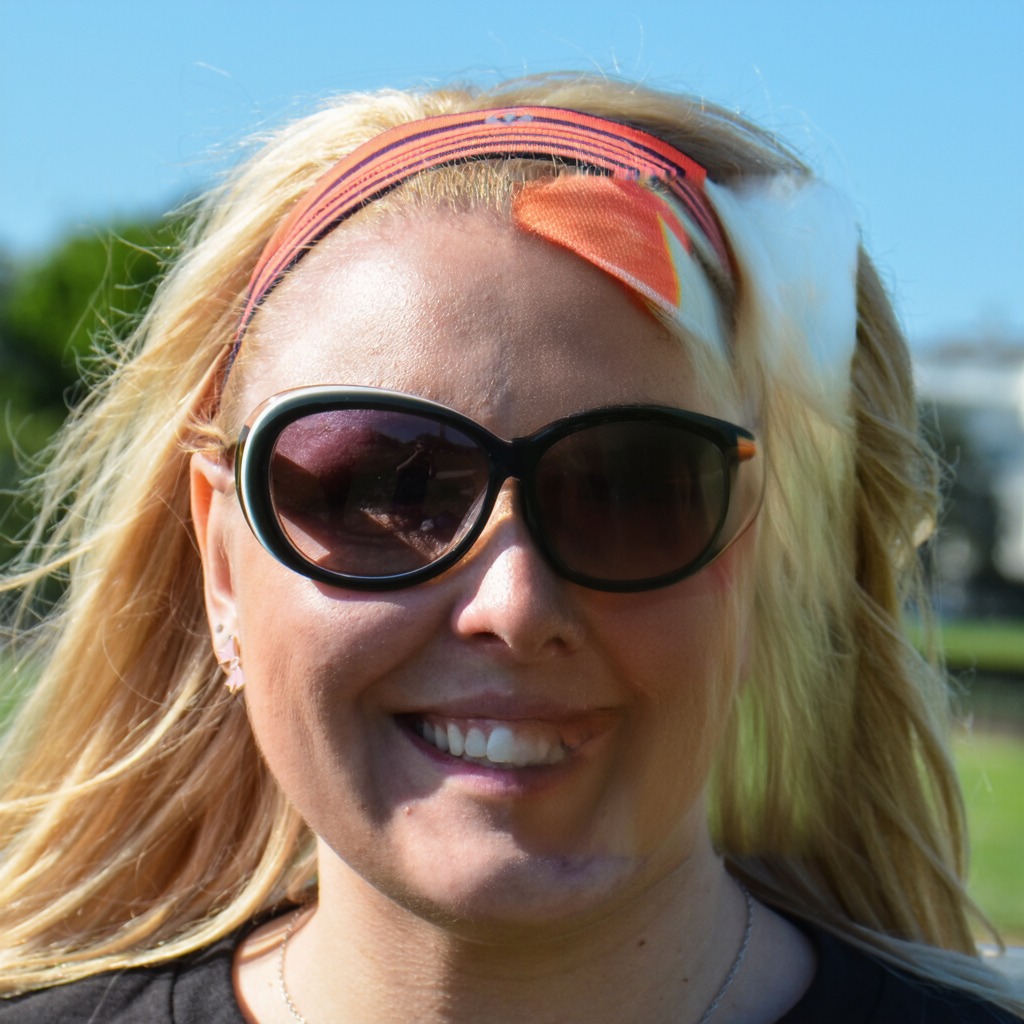} &
        \oursimg{\includegraphics[width=\imwidth]{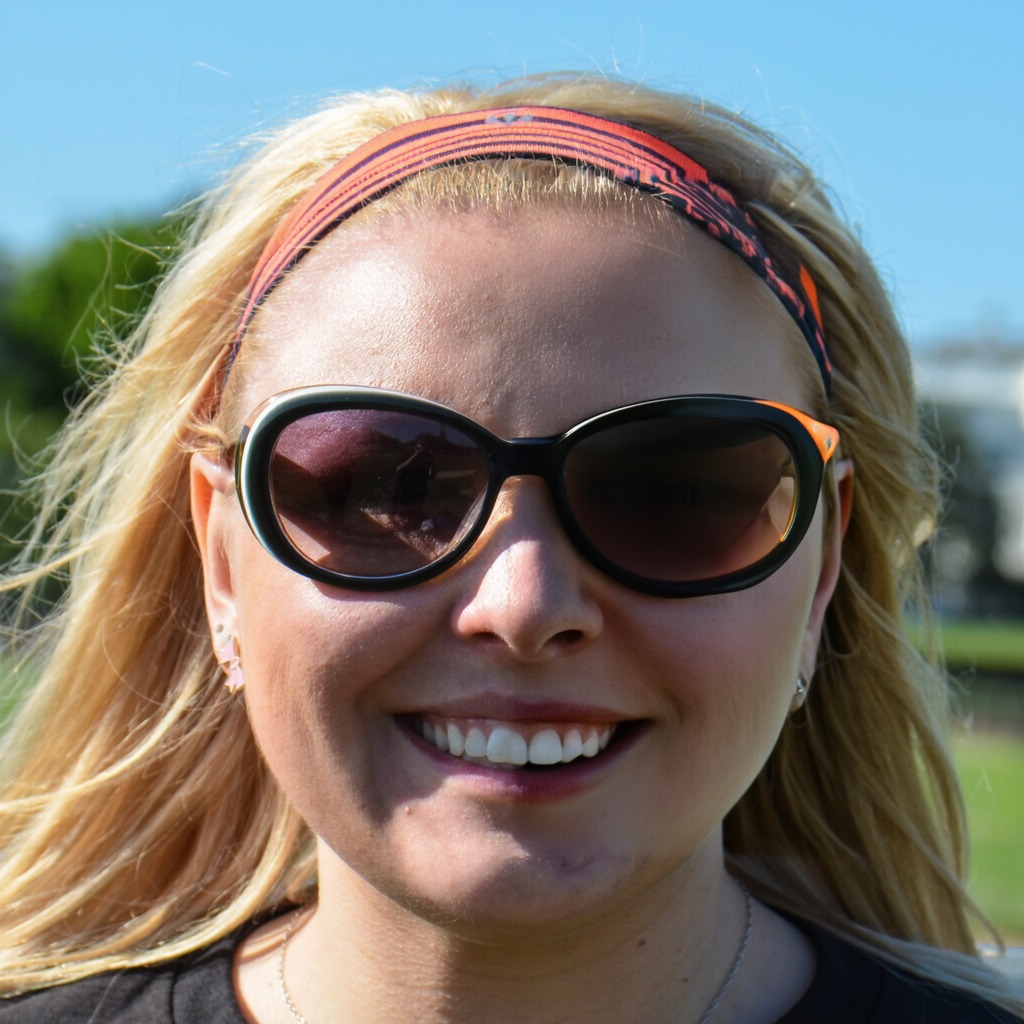}} \\[3pt]
        \includegraphics[width=\imwidth]{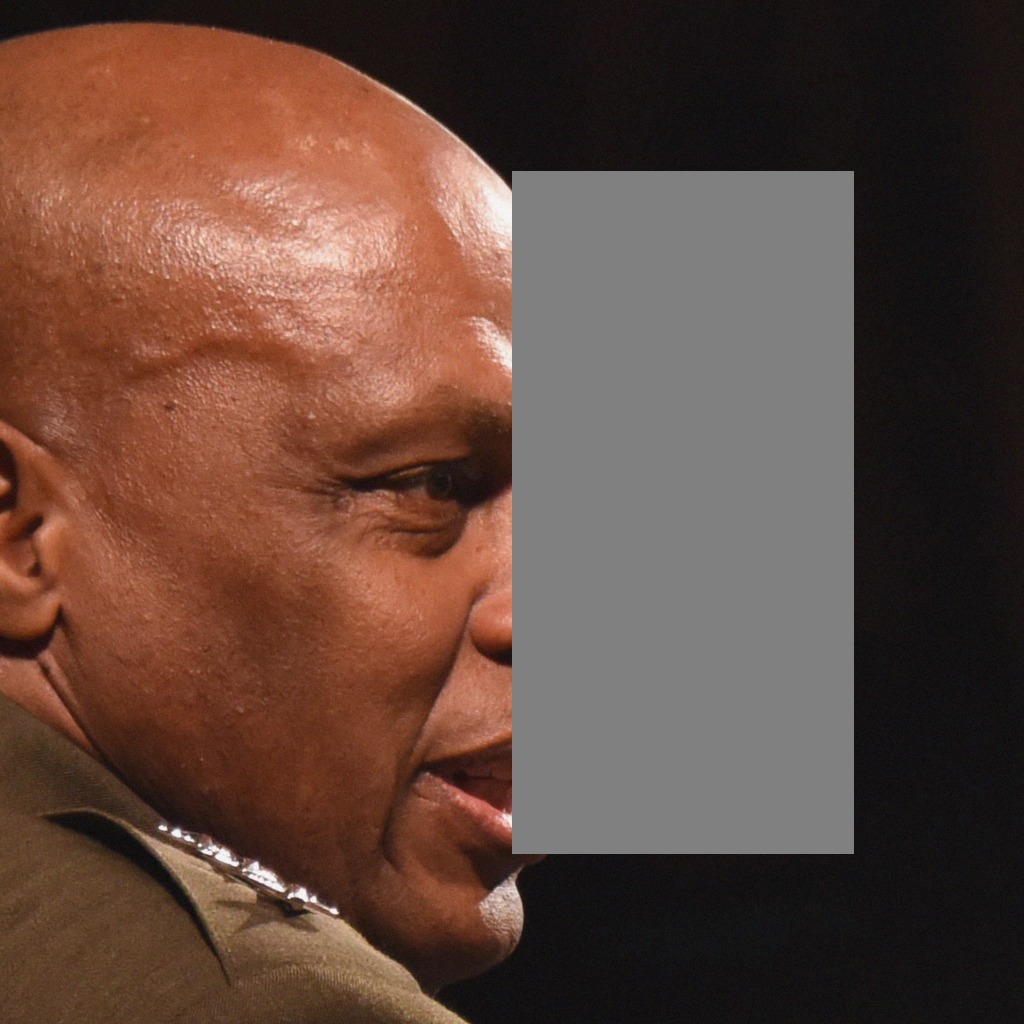} &
        \includegraphics[width=\imwidth]{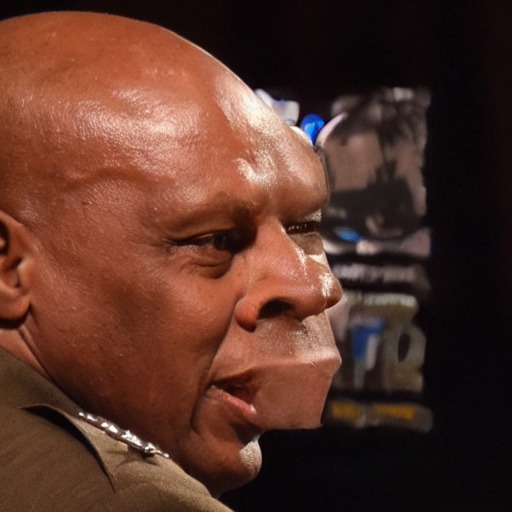} &
        \oursimg{\includegraphics[width=\imwidth]{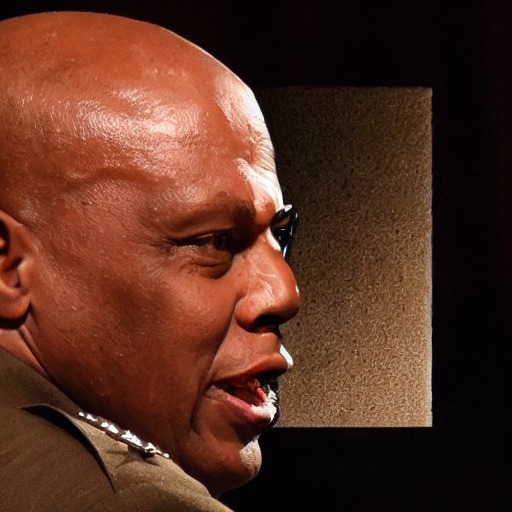}} &
        \includegraphics[width=\imwidth]{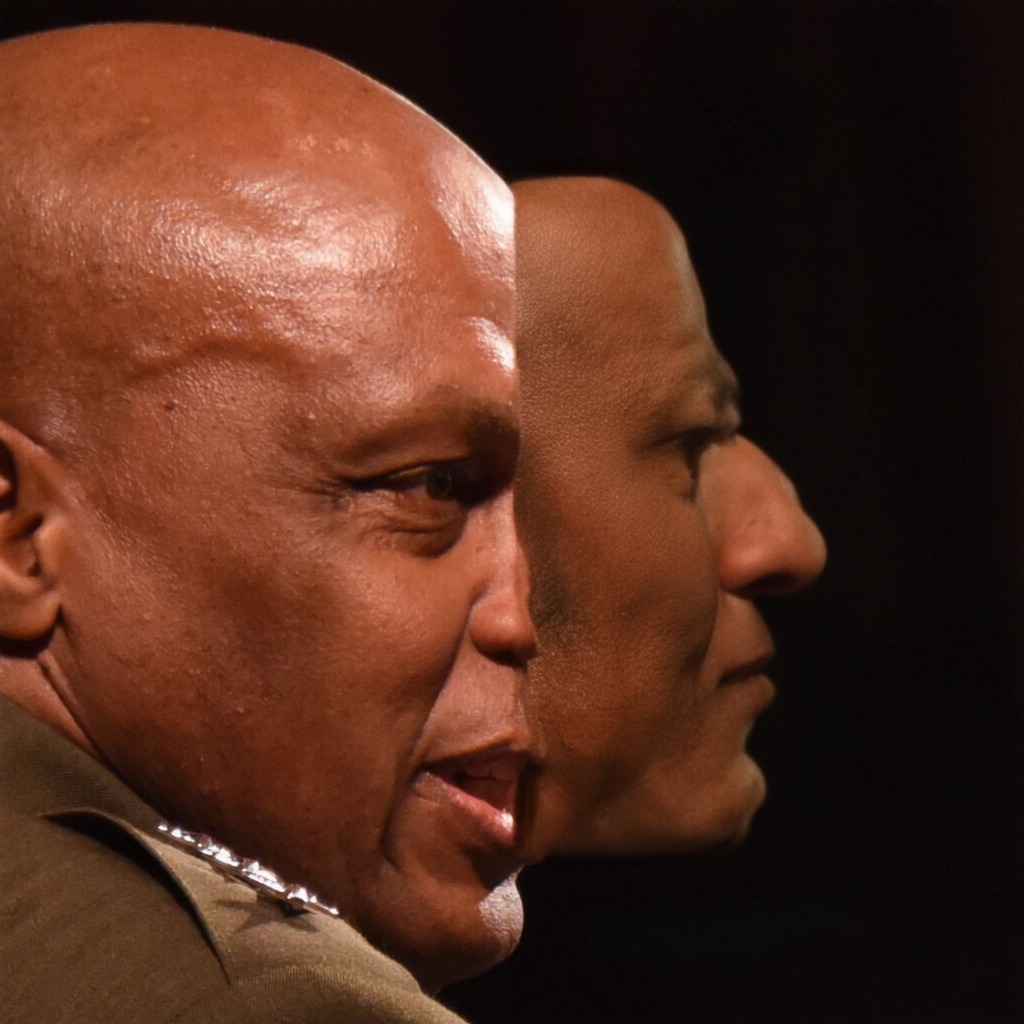} &
        \oursimg{\includegraphics[width=\imwidth]{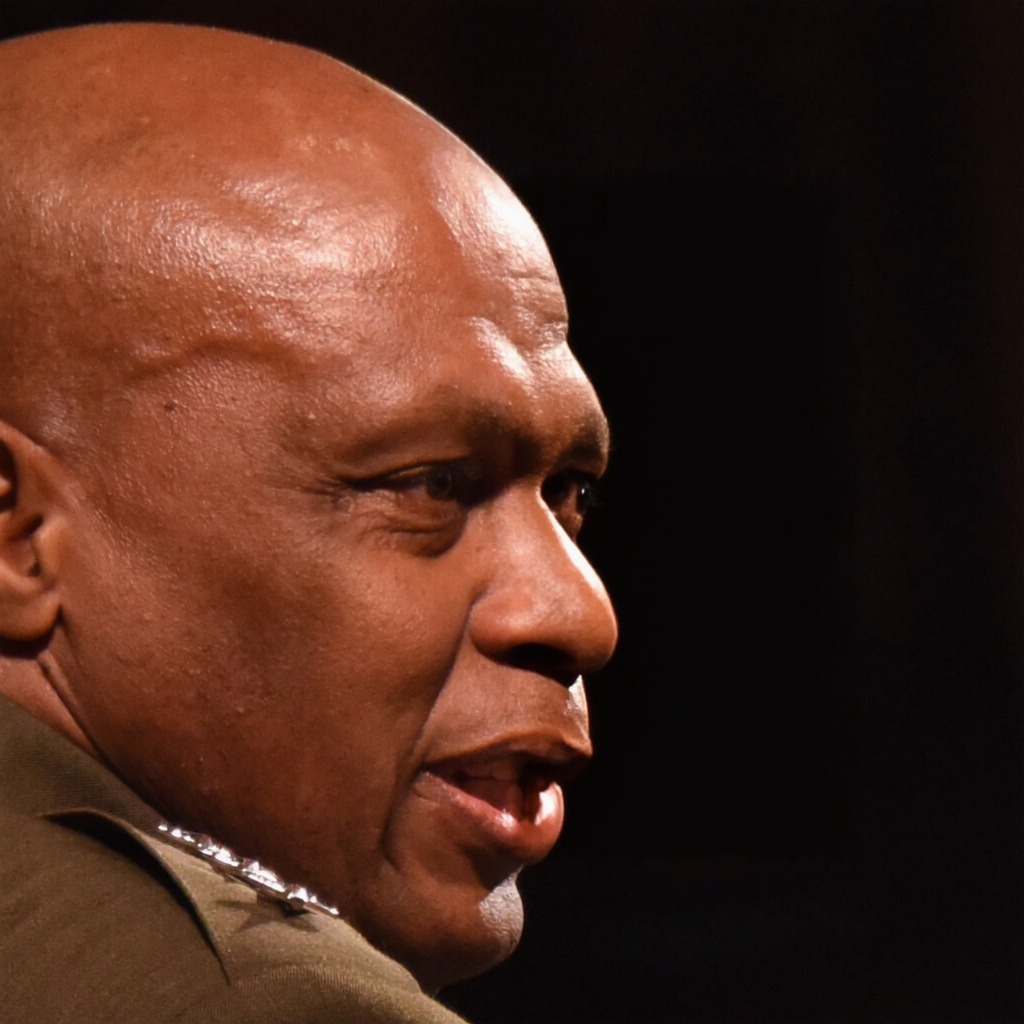}} \\[3pt]
        \includegraphics[width=\imwidth]{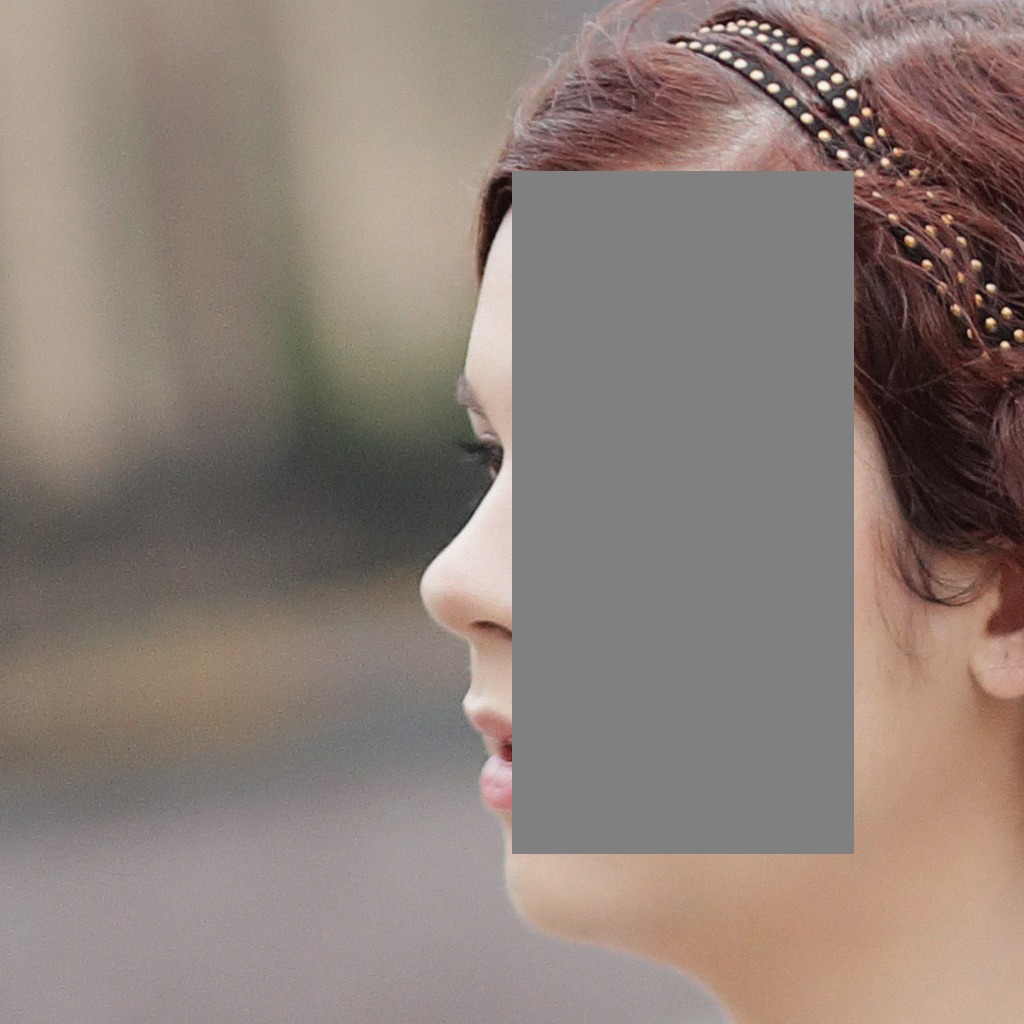} &
        \includegraphics[width=\imwidth]{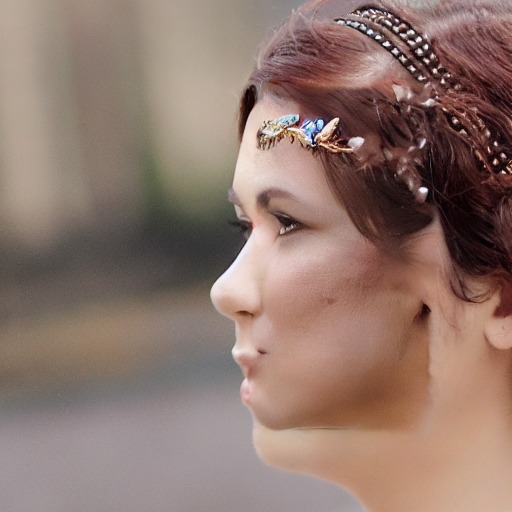} &
        \oursimg{\includegraphics[width=\imwidth]{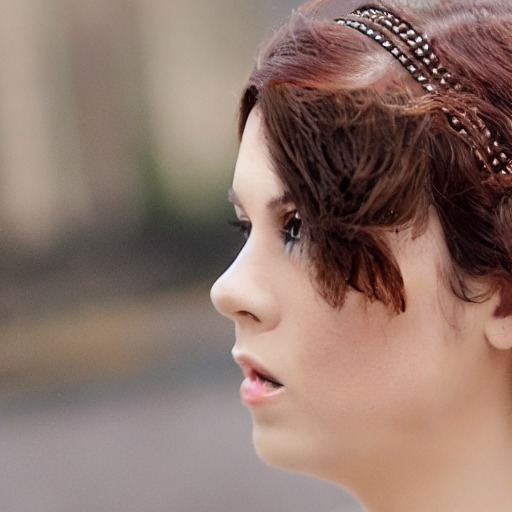}} &
        \includegraphics[width=\imwidth]{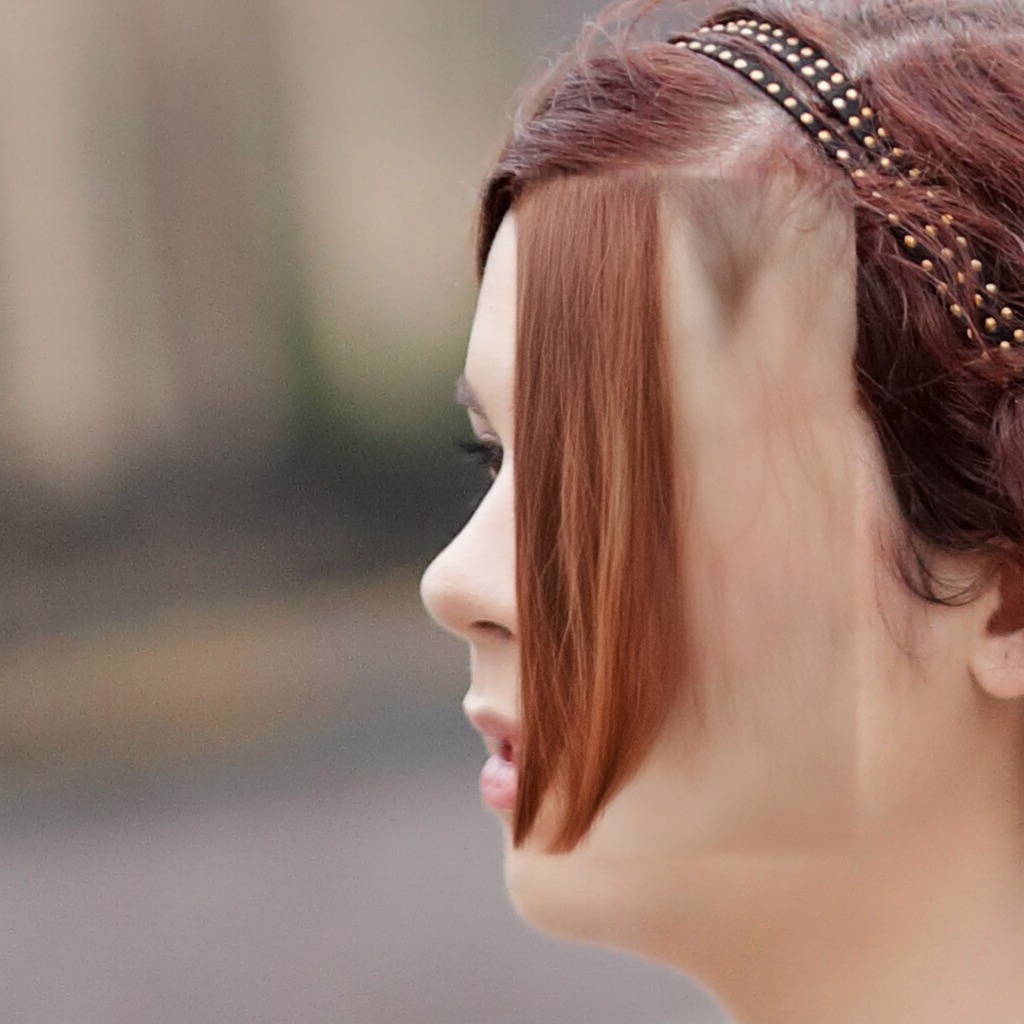} &
        \oursimg{\includegraphics[width=\imwidth]{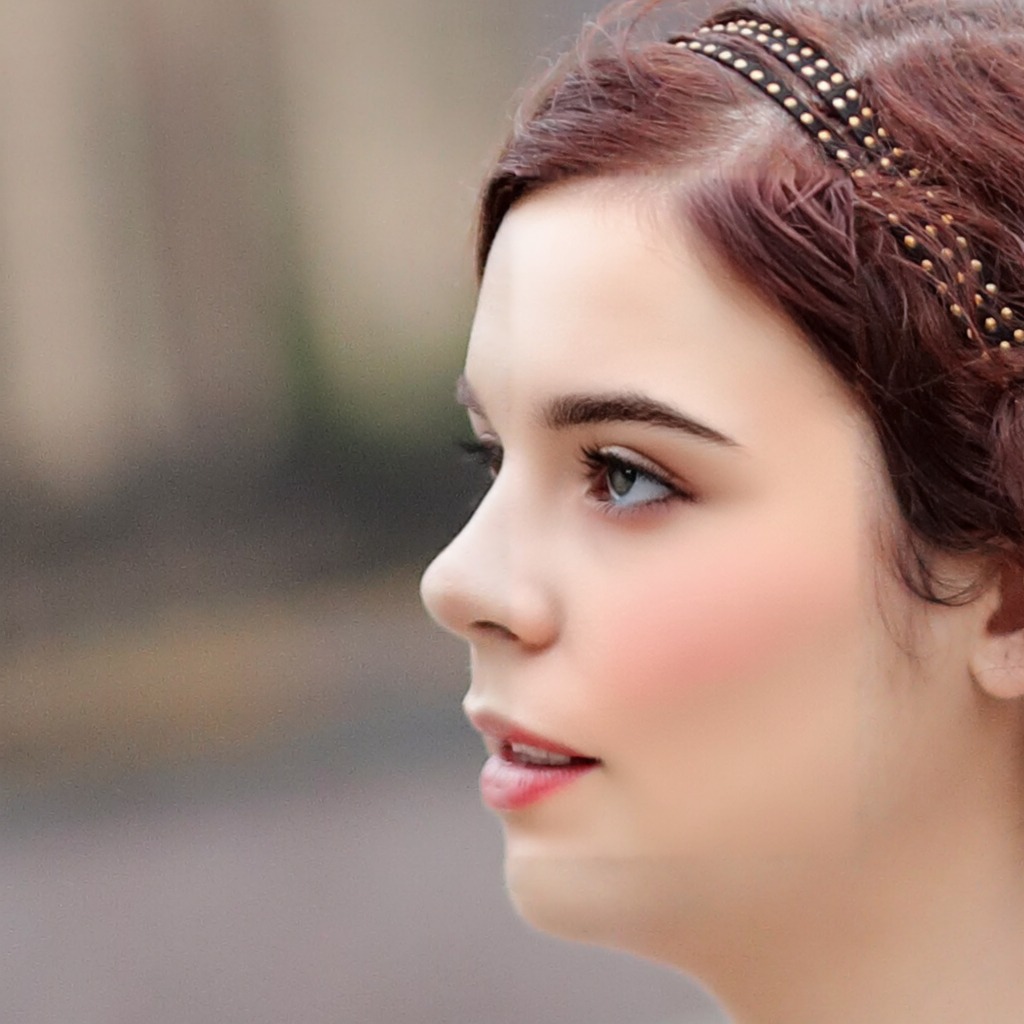}} \\[2pt]
        {\tiny Input} &
        {\tiny \makecell{BLD-SD1.5\\(rand)}} &
        {\tiny \makecell{\bf BLD-SD1.5\\+ Our method}} &
        {\tiny \makecell{BLD-SDXL\\(rand)}} &
        {\tiny \makecell{\bf BLD-SDXL\\+ Our method}} \\
    \end{tabular}
    \renewcommand{\arraystretch}{1.0}
    \setlength{\tabcolsep}{6pt}
    \vspace{-0.5em}
    \caption{{\bf Qualitative results for non-flow model backbones} -- 
    Our method works even with diffusion models where the trajectories are not as straight.
    }
    \label{fig:bld_nonflow}
    \vspace{-1em}
\end{figure}

We also apply our spectral noise optimization for diffusion backbones such as StableDiffusion 1.5~\cite{rombach2022high} and SDXL~\cite{podell2024sdxl}.
Though the denoising trajectories of these models may not be as straight as a flow model, our optimization method prepended to BLD~\cite{Avrahami_2023} is able to achieve reasonable inpainted outputs, outperforming BLD results with randomly initialized noise distributions.
Quantitative results, comparing against other state-of-the-art noise optimization methods, are included in \cref{sec:supp_sota-noiseopt-comparison}.
Qualitative results are shown in \cref{fig:bld_nonflow}.

\section{Theoretical analysis of spatial vs. spectral optimization}
\label{sec:supp_theory}

In the main text and the supplementary materials above, we illustrated the advantages of our spectral optimization via qualitative and quantitative analysis.
In this section, we more rigorously analyze the theoretical foundations of this choice, first proving the fundamental difference between vanilla gradient descent and spectral optimization with Adam~\cite{kingma2017adammethodstochasticoptimization}.
We then further discuss the implications and benefits of this for optimization over short optimization horizons.

\subsection{Non-equivalence under coordinate-wise non-linear solvers}

We formally demonstrate that optimizing a latent representation in the spectral (Fourier) domain using a coordinate-wise optimizer (e.g., Adam) yields a fundamentally different update trajectory than optimizing via gradient descent with a fixed step size.

Let $x_T \in \mathbb{R}^N$ be the \textbf{spatial} latent tensor of the initial noise distribution, and $X_T \in \mathbb{C}^N$ be its \textbf{spectral} representation such that $X_T = \mathcal{F}x$, where $\mathcal{F}$ is the Discrete Fourier Transform (DFT) matrix, dropping parenthesis for clarity.
Since $\mathcal{F}$ is unitary (up to a scaling factor), $\mathcal{F}^{-1} \propto \mathcal{F}^\dagger$, where $\dagger$ is the conjugate transpose, or Hermitian.

Let the objective function be $\mathcal{L}(x_T) = \mathcal{L}(\mathcal{F}^{-1}X_T)$. By the chain rule, the gradient in the spectral domain $g_{X}$ is a linear transformation of the spatial gradient:
\[
g_{X} = \nabla_{X}  \mathcal{L} = \mathcal{F} (\nabla_{x} \mathcal{L}) = \mathcal{F} g_{x}
\]

If we apply standard Gradient Descent (SGD without momentum), the spatial update is $\Delta x_{sgd} = -\eta g_x$. The spectral update is $\Delta X_{sgd} = -\eta \mathcal{F} g_x$. Projecting the spectral update back to the spatial domain yields:
\[
\Delta x_{effective} = \mathcal{F}^{-1}(-\eta \mathcal{F} g_{x}) = -\eta (\mathcal{F}^{-1}\mathcal{F}) g_{x} = -\eta g_{x}
\]
Under vanilla SGD, the domains are \textbf{strictly equivalent}, as already observed empirically in \cref{fig:optimization_sweep}.
However, modern solvers like Adam apply a pointwise non-linear function $h(\cdot)$ to the gradients, governed by moving averages of the first and second moments.
In the early stages of optimization (\eg, step $t=1$), the Adam update direction is heavily dominated by the sign of the gradient: $h(g) \approx \text{sgn}(g)$ \footnote{Assuming zero-initialized moment buffers and omitting the numerical stability term $\epsilon$, the asymptotic first step update simplifies to precisely $g / \sqrt{g^2} = \text{sgn}(g)$.}.

The \textbf{spatial} Adam update is therefore:
\[
\Delta x_{T} \approx -\eta \cdot \text{sgn}(g_x)
\]

In contrast, the \textbf{spectral} Adam update applies the non-linear preconditioner to the frequency coordinates:
\[
\Delta X_{T} \approx -\eta \cdot \text{sgn}(g_X) = -\eta \cdot \text{sgn}(\mathcal{F} g_x)
\]

To compare the two, we project the spectral update back to the spatial domain:
\[
\Delta x_{effective} = \mathcal{F}^{-1} \Delta X_{T} \approx -\eta \cdot \mathcal{F}^{-1} \text{sgn}(\mathcal{F} g_x)
\]

\paragraph{Proof of Non-Equivalence:} 
If spectral optimization were merely a spatial update with a larger effective learning rate $c$, there must exist a scalar $c$ such that:
\[
\mathcal{F}^{-1} \text{sgn}(\mathcal{F} g_x) = c \cdot \text{sgn}(g_x)
\]
\emph{Because $\mathcal{F}$ is a dense, global linear transformation and $\text{sgn}(\cdot)$ is a pointwise non-linear operator, \textbf{they do not commute}.} The equality only holds if $h(\cdot)$ is strictly linear, which violates the definition of coordinate-wise adaptive optimizers. Therefore, spectral Adam takes a fundamentally distinct vector step in $\mathbb{R}^N$ that cannot be replicated by altering the scalar learning rate $\eta$ in the spatial domain.

\subsection{Implications for short-horizon optimization ($T \le 20$)}

In our setting, optimization is constrained to a highly limited horizon (e.g., 20 steps). This further highlights the necessity of the spectral basis:

\paragraph{Global Spatial Entanglement:} 
A single parameter update in the spatial domain $\Delta x_{T}$ acts locally. It requires $\mathcal{O}(N)$ steps for gradient information to propagate and form coherent global structures. Conversely, a single coordinate in $X$ represents a global sinusoidal basis function. The update $\mathcal{F}^{-1}\text{sgn}(\mathcal{F}g_x)$ applies a dense, perfectly correlated update across all spatial pixels simultaneously, enabling structural formation within a mere 20 steps.

\paragraph{Spectral Disentanglement \& Divergence:} 
The loss landscape of diffusion latents is spectrally skewed. Low frequencies exhibit massive gradient variance, while high frequencies exhibit minute variance. Spatial Adam entangles these frequencies within every pixel. If one attempts to emulate the fast convergence of spectral methods by merely increasing the spatial learning rate, the local updates become dominated by low-frequency overshoot, leading to catastrophic color shifts and structural divergence. Spectral optimization acts as a per-frequency preconditioner, natively normalizing these variances via Adam's $1/\sqrt{v_t}$ term in the orthogonal basis, allowing stable convergence of high frequencies without destroying low-frequency structures.

\section{Wall clock table}
\label{sec:supp_wall_clock}
\vspace{-0.5em}
We report wall-clock runtimes against all baselines in \cref{tab:runtime}, measured on an NVIDIA GeForce RTX 5090 GPU.
\begin{table}[H]
\centering
\setlength{\tabcolsep}{2pt}
\resizebox{\linewidth}{!}{
\begin{tabular}{@{}lcccccccc|c@{}}
\toprule
 & \makecell[t]{BrushNet}
 & \makecell[t]{BLD SD3.5}
 & \makecell[t]{FLAIR}
 & \makecell[t]{FLAIR (400 NFEs)}
 & \makecell[t]{FlowChef}
 & \makecell[t]{FlowChef (400 NFEs)}
 & \makecell[t]{FlowDPS}
 & \makecell[t]{FlowDPS (400 NFEs)}
 & \makecell[t]{\textbf{Ours}} \\
\midrule
Time (s)$\downarrow$ 
& 5 & 4 & 57 & 489 & 13 & 103 & 22 & 175 & 67 \\
Mem. (GB)$\downarrow$ 
& 4.5 & 19.5 & 21.6 & 21.7 & 24.3 & 24.3 & 13.1 & 13.1 & 19.5 \\
\bottomrule
\vspace{0.2em}
\end{tabular}
}
\caption{{\bf Wall-clock time and memory -- } Inference time (seconds) and GPU memory (GB) per image. Our method runs within a comparable budget to other optimization-based methods.}
\label{tab:runtime}
\end{table}

\section{Additional comparisons with other initial noise optimization methods}
\label{sec:supp_sota-noiseopt-comparison}
\vspace{-0.5em}

\begin{figure}[H]
\vspace{-0.5em}
    \centering
    \begin{subfigure}[t]{0.15\columnwidth}
        \centering
        \includegraphics[width=\linewidth]{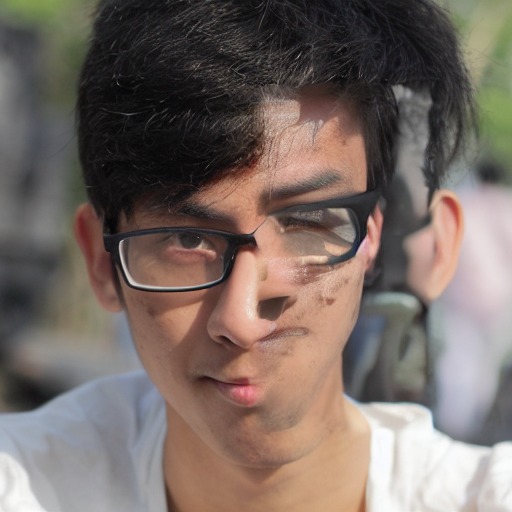}
        \caption*{\tiny BLD-SD1.5}
    \end{subfigure}\hspace{0.004\columnwidth}
    \begin{subfigure}[t]{0.15\columnwidth}
        \centering
        \includegraphics[width=\linewidth]{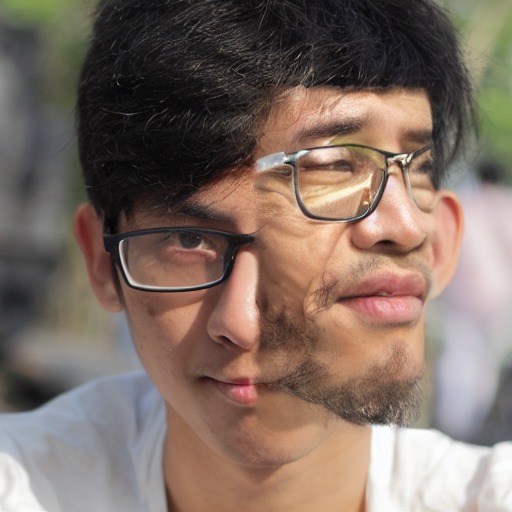}
        \caption*{\tiny w/ InitNo}
    \end{subfigure}\hspace{0.004\columnwidth}
    \begin{subfigure}[t]{0.15\columnwidth}
        \centering
        \includegraphics[width=\linewidth]{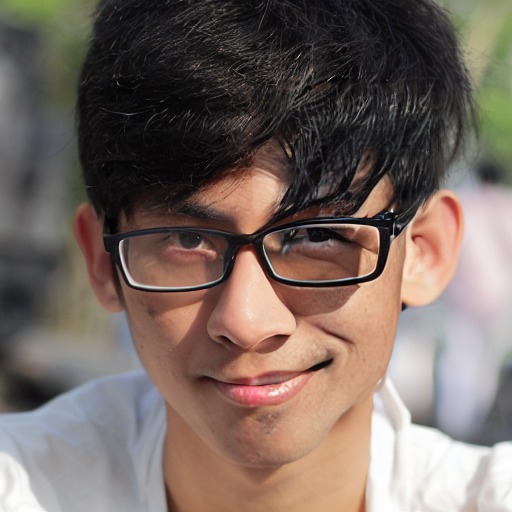}
        \caption*{\tiny w/ Our Method}
    \end{subfigure}\hspace{0.004\columnwidth}
    \begin{subfigure}[t]{0.15\columnwidth}
        \centering
        \includegraphics[width=\linewidth]{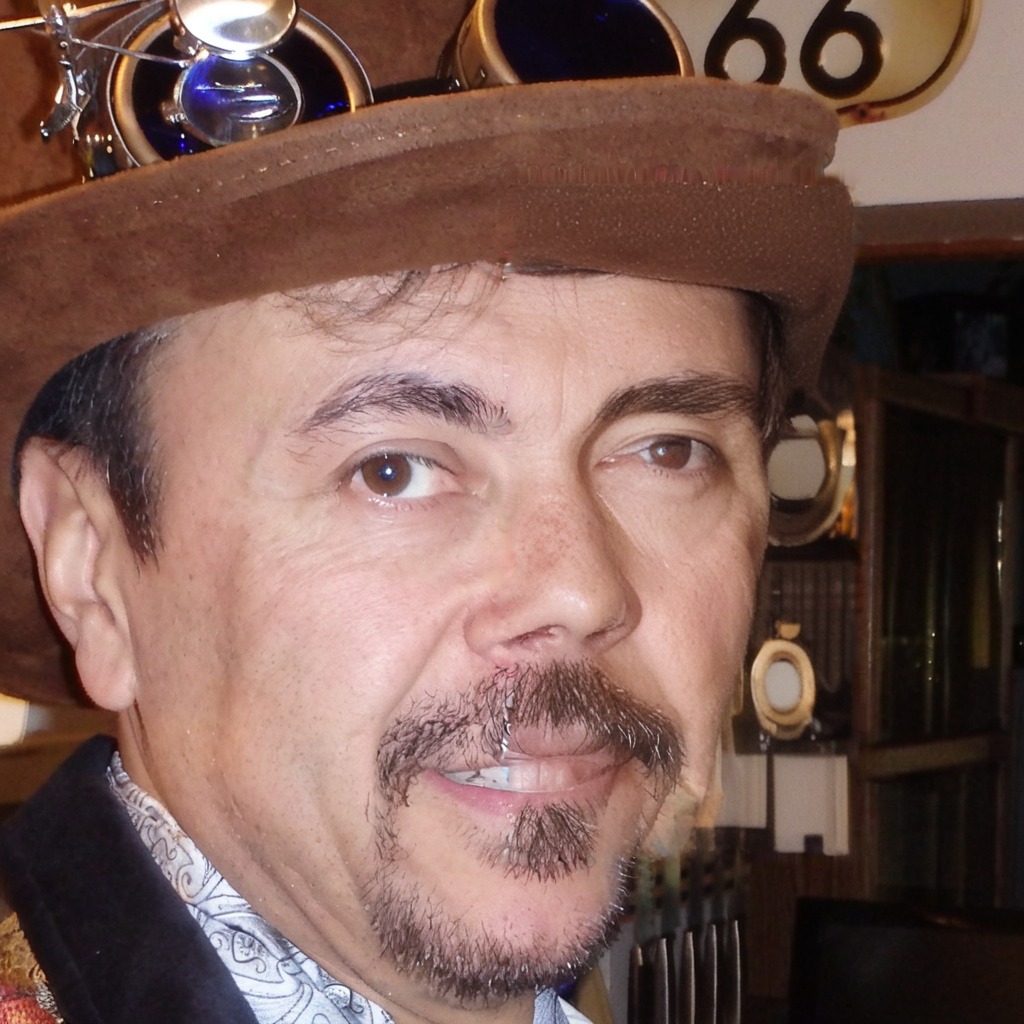}
        \caption*{\tiny BLD-SDXL}
    \end{subfigure}\hspace{0.004\columnwidth}
    \begin{subfigure}[t]{0.15\columnwidth}
        \centering
        \includegraphics[width=\linewidth]{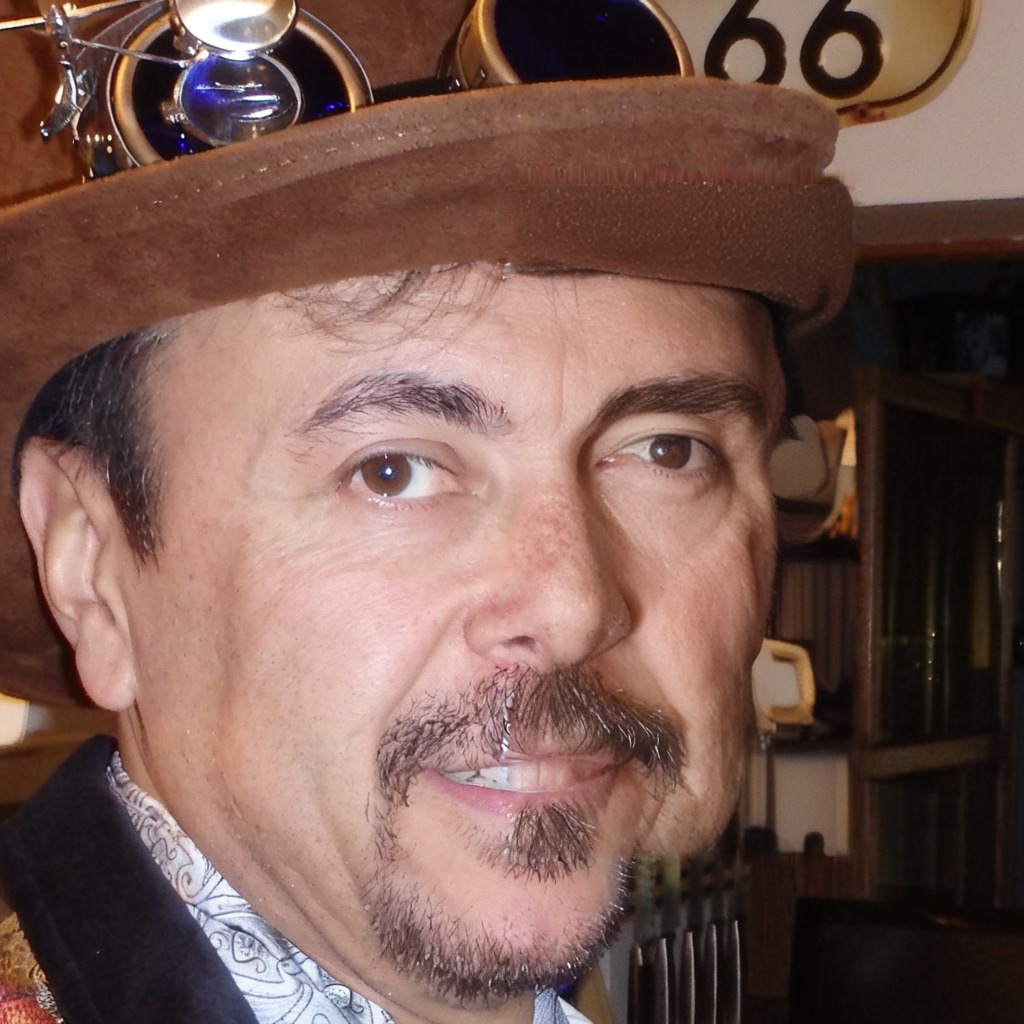}
        \caption*{\tiny w/ GoldenNoise}
    \end{subfigure}\hspace{0.004\columnwidth}
    \begin{subfigure}[t]{0.15\columnwidth}
        \centering
        \includegraphics[width=\linewidth]{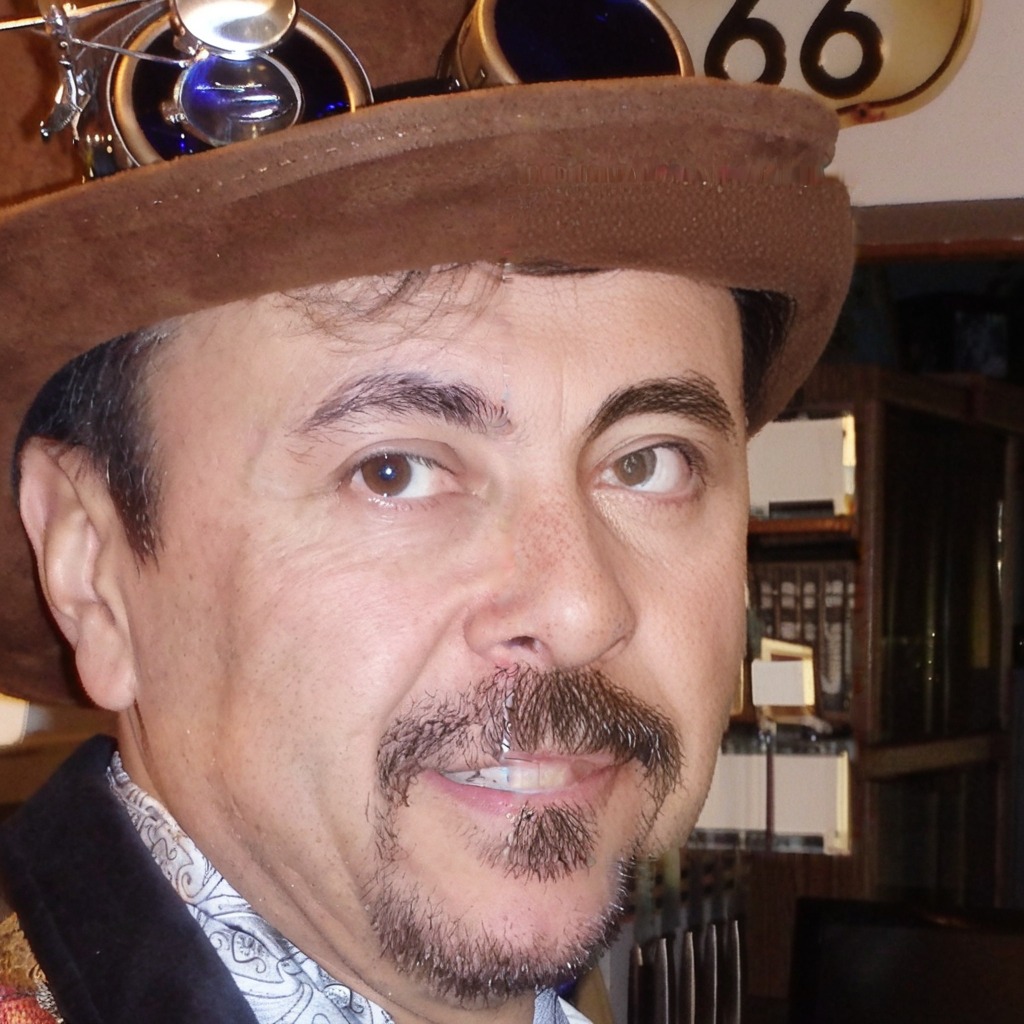}
        \caption*{\tiny w/ Our Method}
    \end{subfigure}
    \vspace{-0.5em}
    \caption{{\bf Qualitative comparison with InitNo and GoldenNoise --} InitNo (SD1.5 backbone) and GoldenNoise (SDXL backbone) improve generation quality but remain agnostic of the observed pixels, unlike our method which maintains coherence with the unmasked region.}
    \label{fig:siximages}
    \vspace{-1em}
\end{figure}

While initial noise optimization methods for image models exist, most are not directly comparable to our inpainting objective, as they optimize or learn to find an initial noise distribution that enhances image quality without accounting for partial pixel observations.
Nonetheless, we compare against the following methods on FFHQ:
for \textbf{InitNo}~{\cite{guo2024initno}}, we use the SD1.5~{\cite{rombach2022high}} backbone, as this method was developed for the SD1.x family;
for \textbf{GoldenNoise}~{\cite{zhou2025golden}}, we use the official pretrained network for SDXL~{\cite{podell2024sdxl}}.
\textbf{FreeInpaint}~{\cite{gong2026freeinpaint}} optimizes the initial noise distribution with reference to the partial observation, but only prepends the optimized noise to a \textit{trained} inpainting backbone. Thus, we compare against both the official code using BrushNet~{\cite{ju2024brushnet}} and against the BLD-SD3.5~{\cite{Avrahami_2023,esser2024}} setup for fairness.
As shown in \cref{tab:sota-noiseopt}, \textbf{InitNo}, while improving CLIP scores, degrades every other metric.
\textbf{GoldenNoise} improves all metrics slightly, but our method still outperforms it.
\textbf{FreeInpaint} with BrushNet provides the best IR, CLIP, and AS scores, but the wide gap in LPIPS and FID indicates worse alignment with the unmasked region; this gap is exacerbated when applied with BLD.
The qualitative examples in \cref{fig:siximages} further illustrate how \textbf{InitNo} and \textbf{GoldenNoise} improve generation quality but remain agnostic of the observed pixels.
\vspace{-0.5em}
\begin{table}[H]
\centering
\small
\setlength{\tabcolsep}{10pt}
\resizebox{\linewidth}{!}{%
\begin{tabular}{@{}lcccccccc}
\toprule
Method & PSNR$\uparrow$ & SSIM$\uparrow$ & LPIPS$\downarrow$ & FID$\downarrow$ & IR$\uparrow$ & HPS v2$\uparrow$ & CLIP$\uparrow$ & AS$\uparrow$\\
\midrule
SD1.5-BLD          & \best{19.341} & \best{0.740} & \second{0.273} & \second{26.897} & \second{-0.144} & \second{0.212} & \third{19.772} & \second{5.337} \\
SD1.5-BLD + InitNO     & \third{17.836} & \third{0.726} & \third{0.293} & \third{35.945} & \third{-0.294} & \third{0.209} & \best{22.372} & \third{5.068} \\
\textbf{SD1.5-BLD + Our Method} & \second{19.118} & \second{0.734} & \best{0.264} & \best{20.925} & \best{0.074} & \best{0.223} & \second{20.731} & \best{5.565} \\
\midrule
SDXL-BLD (standard) & 19.724 & 0.790 & 0.170 & 19.784 & 0.046 & 0.226 & 23.081 & 5.425 \\
SDXL-BLD + GoldenNoise & \best{19.945} & \best{0.794} & \best{0.168} & 18.817 & 0.061 & 0.227 & 23.131 & 5.443 \\
\textbf{SDXL-BLD + Our Method} & 19.782 & 0.792 & \best{0.168} & \best{18.006} & \best{0.083} & \best{0.236} & \best{23.320} & \best{5.449} \\
\midrule
BrushNet + FreeInpaint (Official)          & 21.587 & 0.824 & 0.125 & 19.233 & \best{0.350} & 0.243 & \best{24.338} & \best{5.966} \\
SD3.5-BLD + FreeInpaint& 16.878 & 0.823 & 0.191 & 29.754 & -0.491 & 0.202 & 21.955 & 4.867 \\
\textbf{SD3.5-BLD + Our Method}  & \best{22.562} & \best{0.857} & \best{0.121} & \best{12.883} & 0.280 & \best{0.245} & 24.042 & 5.848 \\
\bottomrule
\end{tabular}
}
\setlength{\tabcolsep}{6pt}
\caption{{\bf Comparison with initial-noise optimization baselines on FFHQ -- } Our method is evaluated against InitNo (SD1.5 backbone), GoldenNoise (SDXL backbone), and FreeInpaint (BrushNet and SD3.5-BLD backbones), each compared within its respective backbone for fairness. Our method outperforms InitNo and GoldenNoise across nearly all metrics, and delivers far better pixel fidelity (LPIPS, FID).}
\label{tab:sota-noiseopt}
\vspace{-1em}
\end{table}

\vspace{-1em}
\begin{figure}[H]
    \centering
    \newcommand{\imwidth}{0.152\linewidth}
    \newcommand{\oursimg}[1]{{\setlength{\fboxsep}{0pt}\setlength{\fboxrule}{2pt}\fcolorbox{red}{white}{#1}}}
    \setlength{\tabcolsep}{1pt}
    \renewcommand{\arraystretch}{0.85}
    \begin{tabular}{cccccc}
        \includegraphics[width=\imwidth]{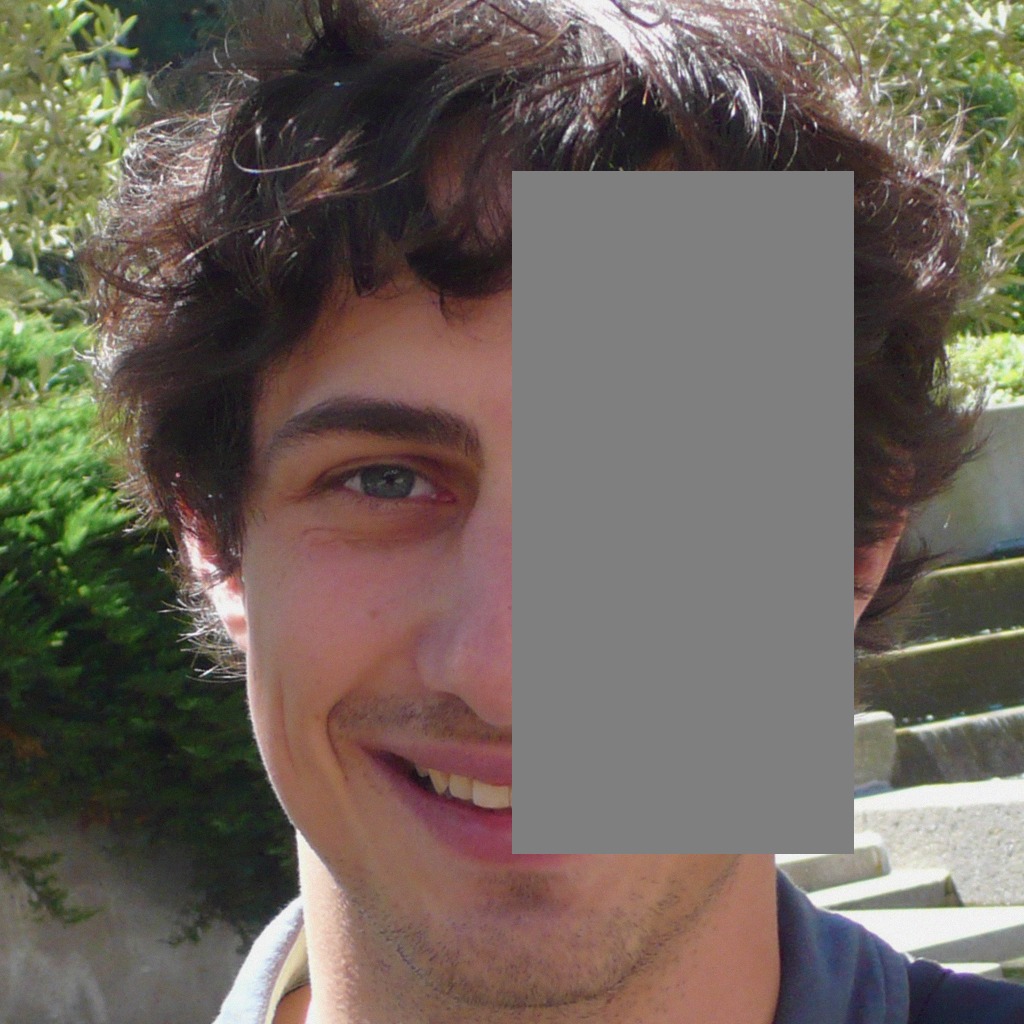} &
        \includegraphics[width=\imwidth]{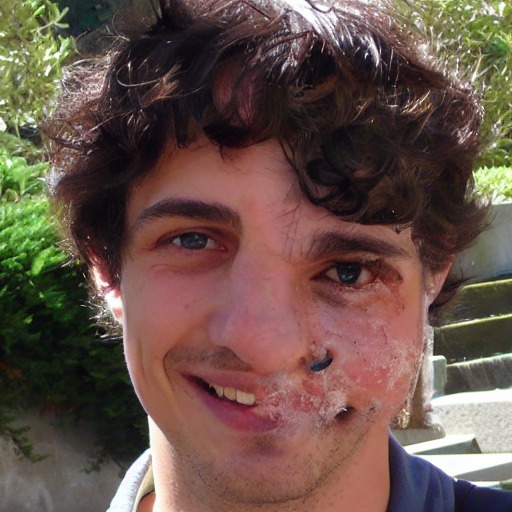} &
        \includegraphics[width=\imwidth]{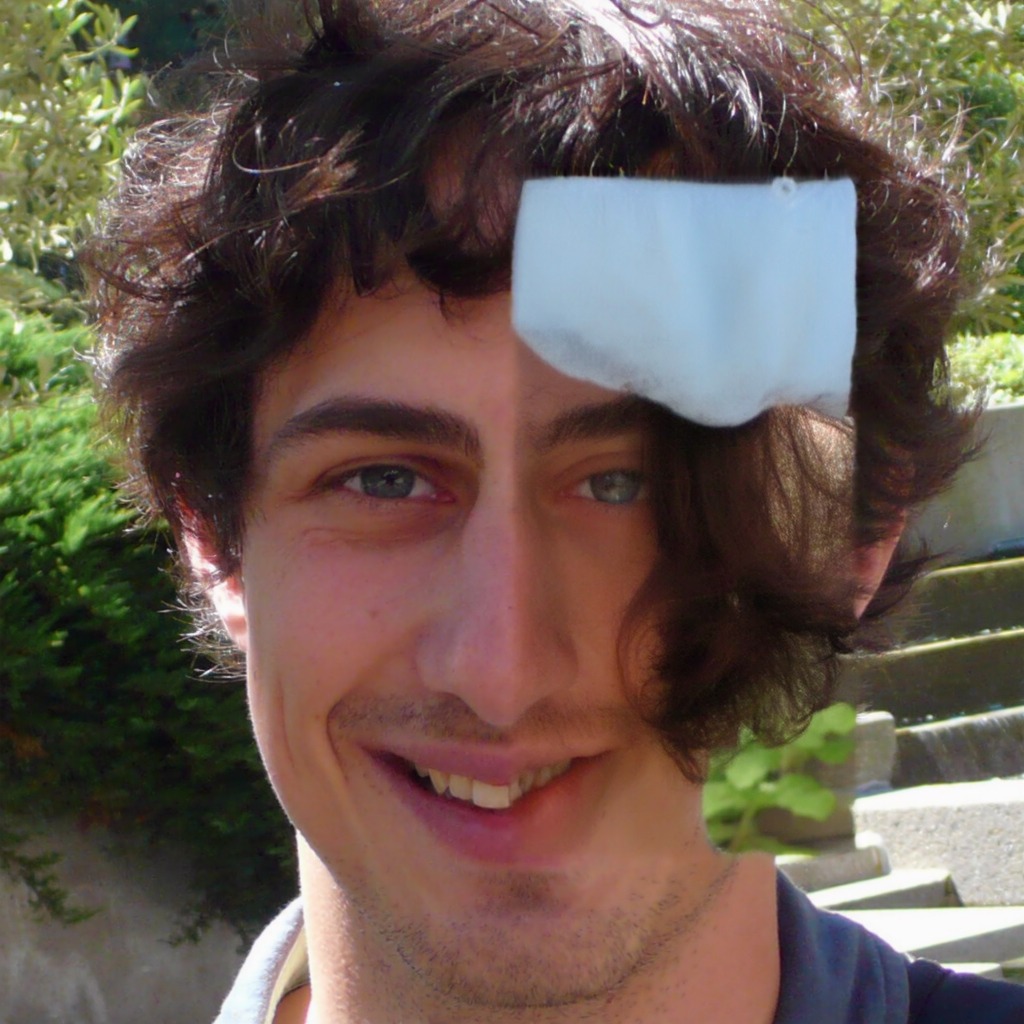} &
        \includegraphics[width=\imwidth]{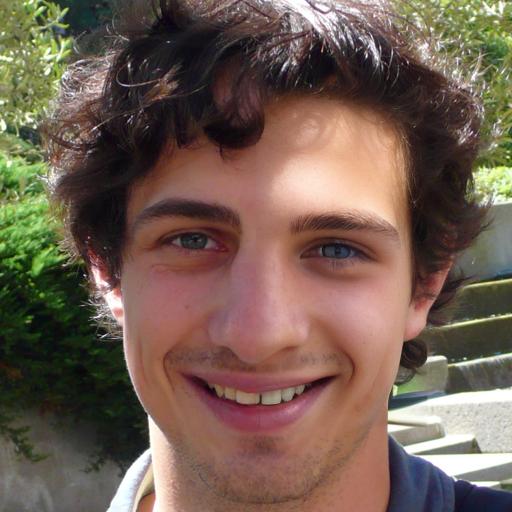} &
        \includegraphics[width=\imwidth]{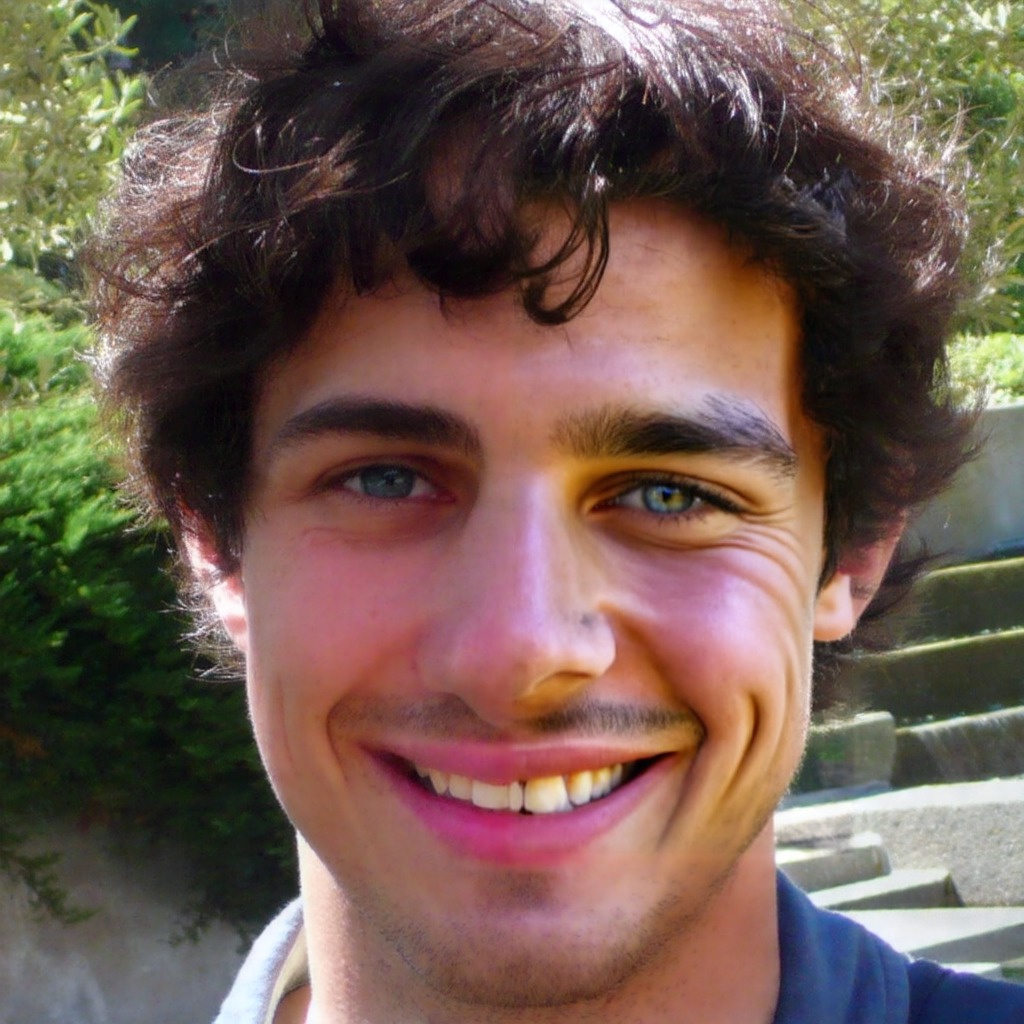} &
        \oursimg{\includegraphics[width=\imwidth]{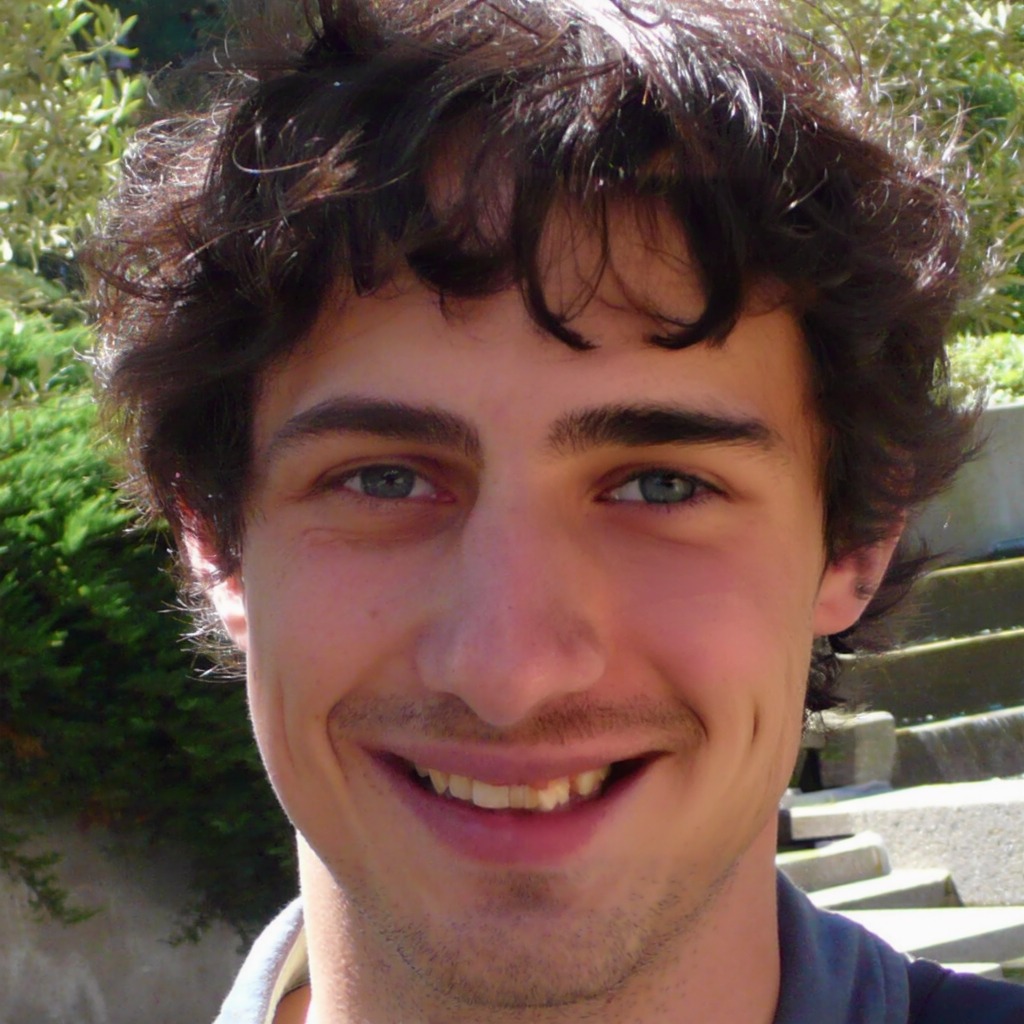}} \\[-2pt]
        {\tiny --} & {\tiny CLIP: 0.218} & {\tiny CLIP: 0.206} & {\tiny \second{CLIP: 0.232}} & {\tiny \best{CLIP: 0.244}} & {\tiny \third{CLIP: 0.219}} \\[3pt]
        \includegraphics[width=\imwidth]{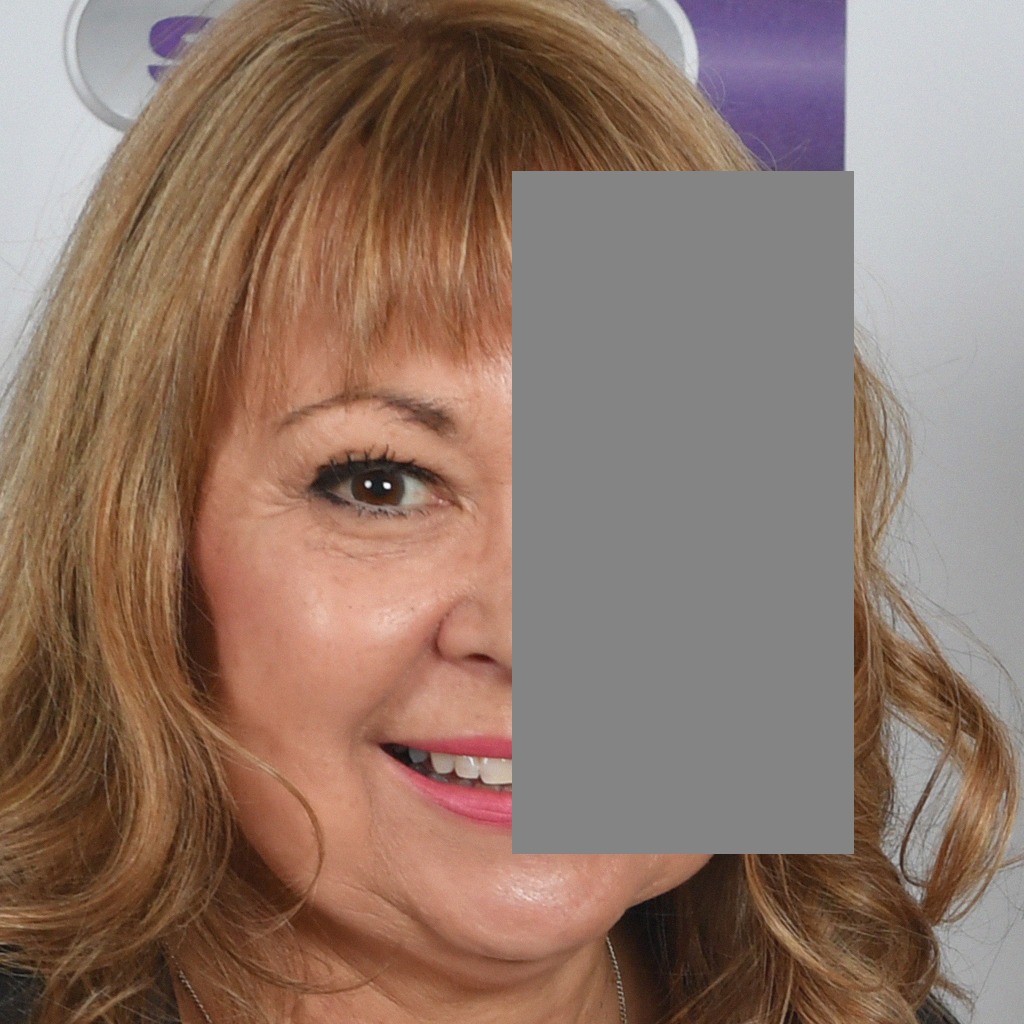} &
        \includegraphics[width=\imwidth]{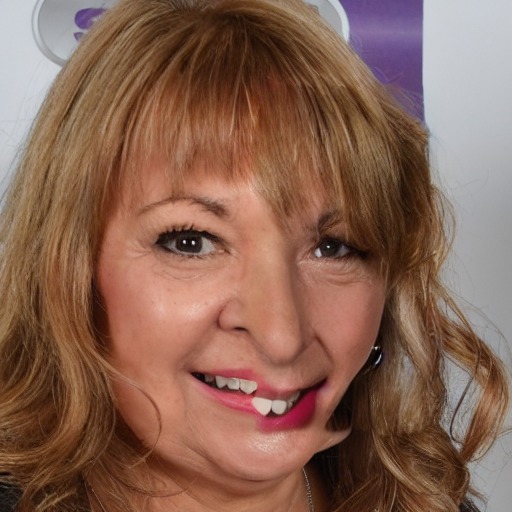} &
        \includegraphics[width=\imwidth]{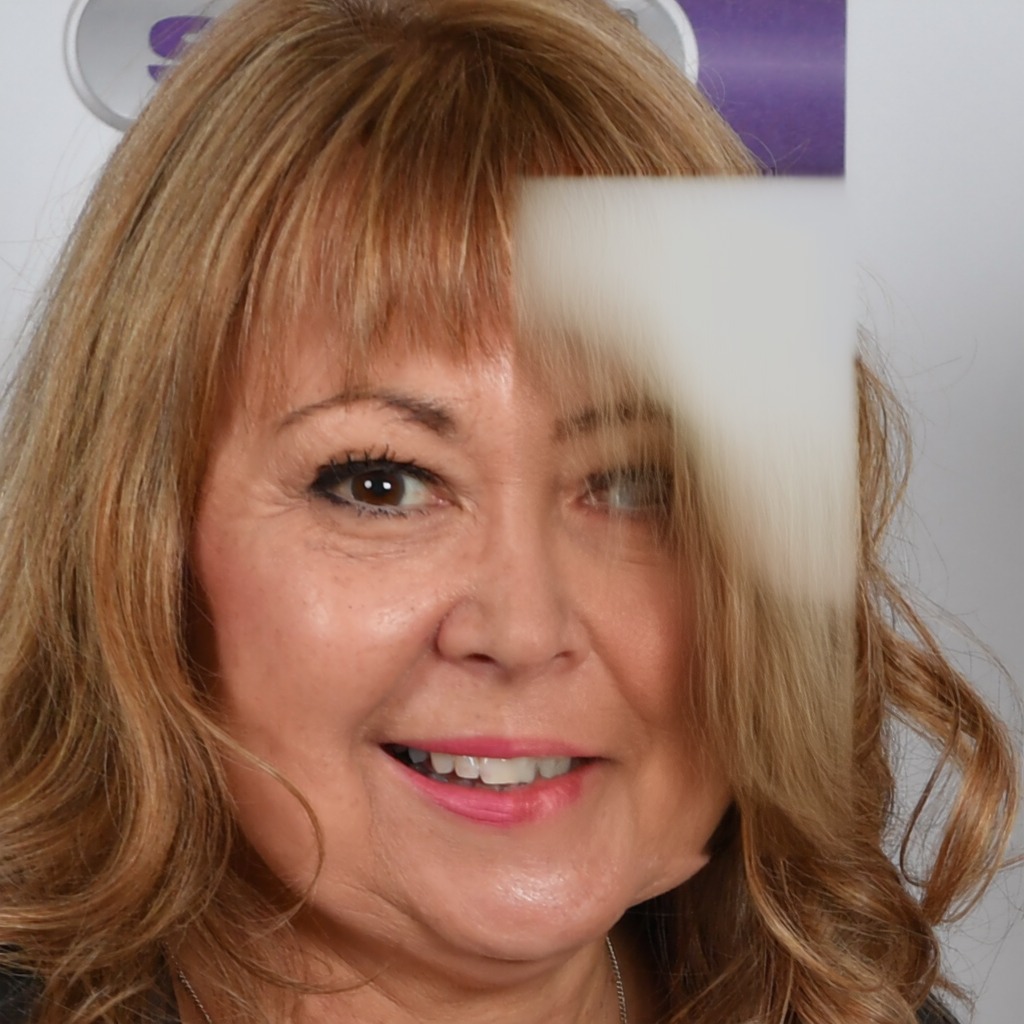} &
        \includegraphics[width=\imwidth]{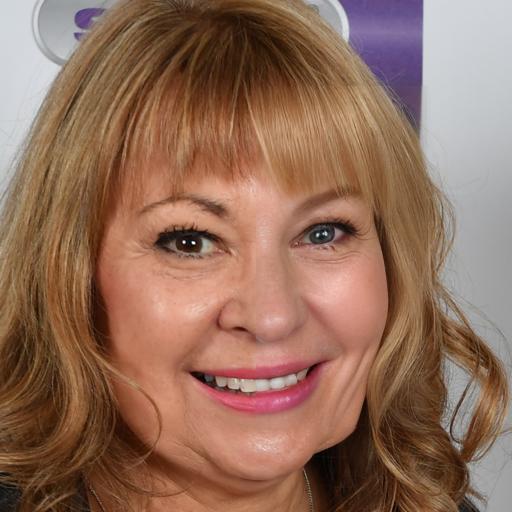} &
        \includegraphics[width=\imwidth]{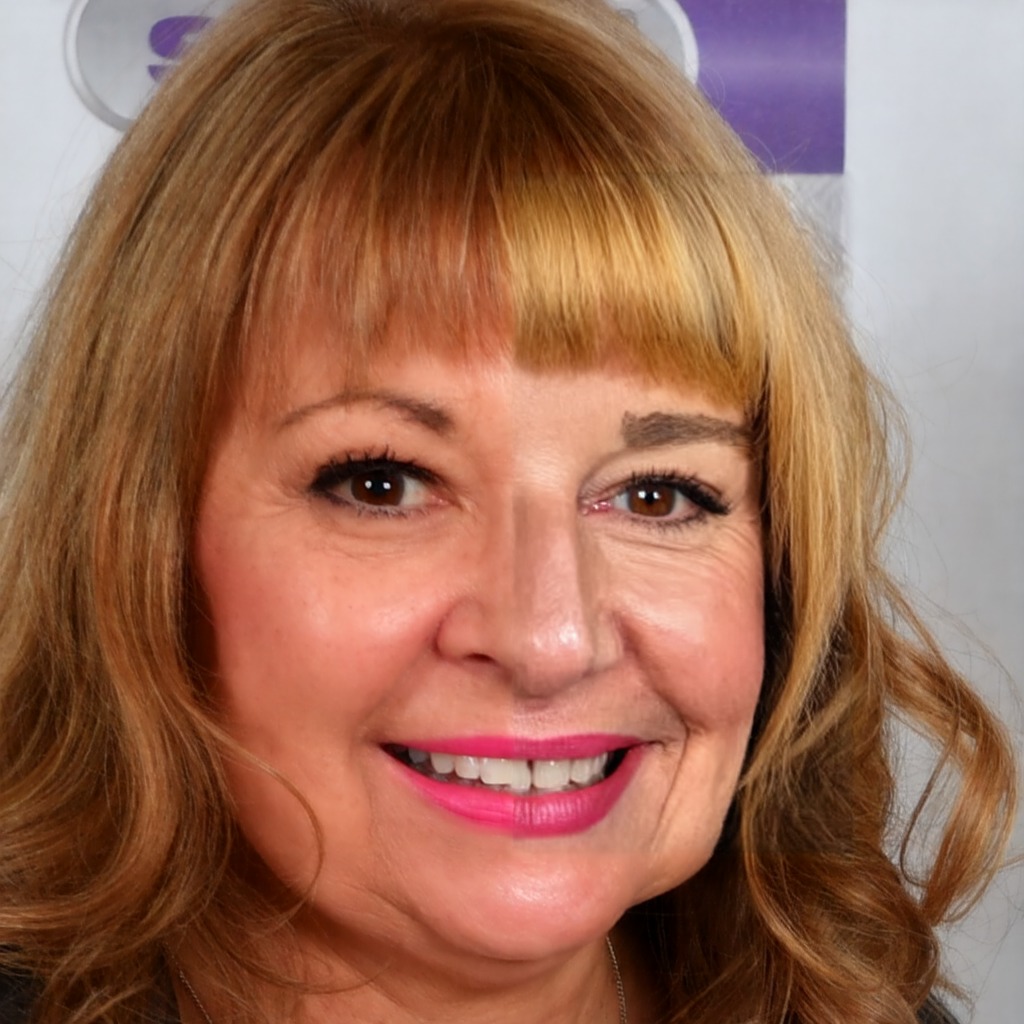} &
        \oursimg{\includegraphics[width=\imwidth]{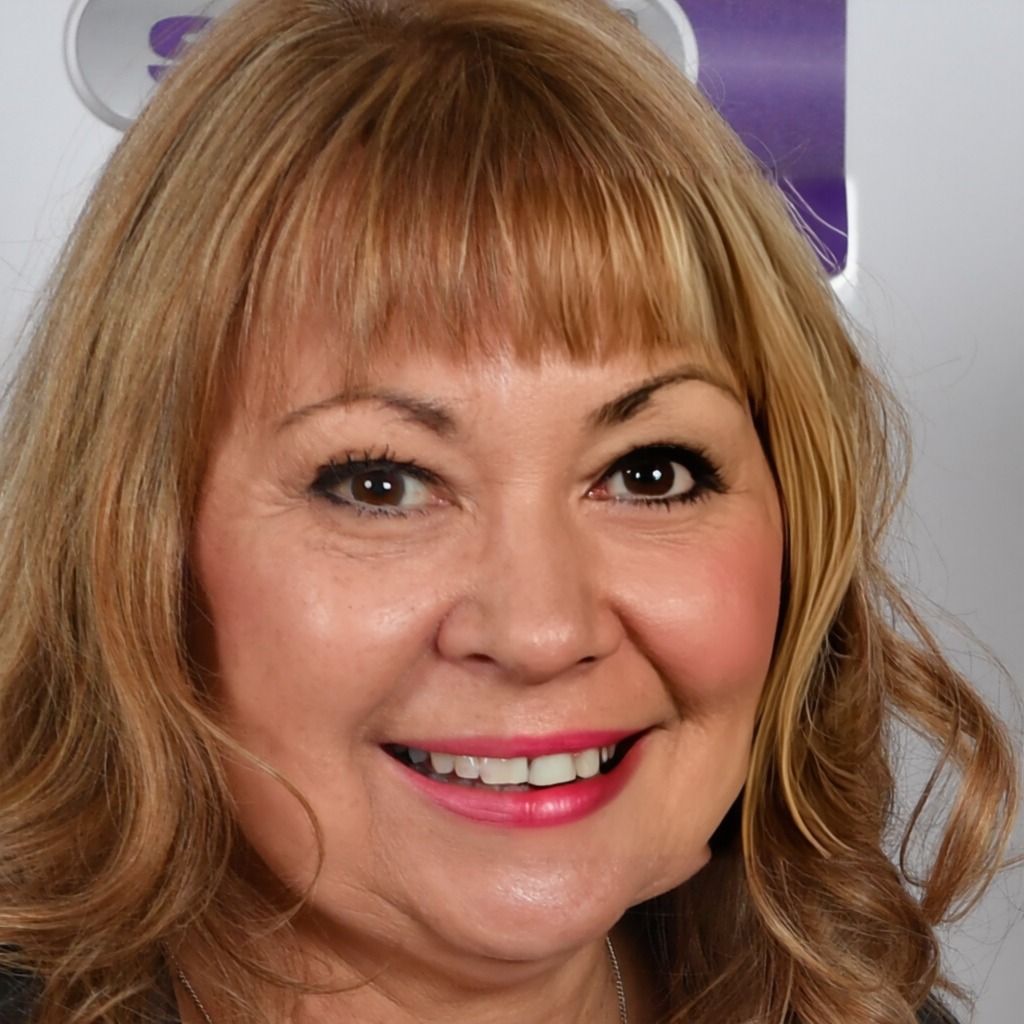}} \\[-2pt]
        {\tiny --} & {\tiny \second{CLIP: 0.224}} & {\tiny CLIP: 0.209} & {\tiny \best{CLIP: 0.229}} & {\tiny \third{CLIP: 0.217}} & {\tiny CLIP: 0.192} \\[3pt]
        \includegraphics[width=\imwidth]{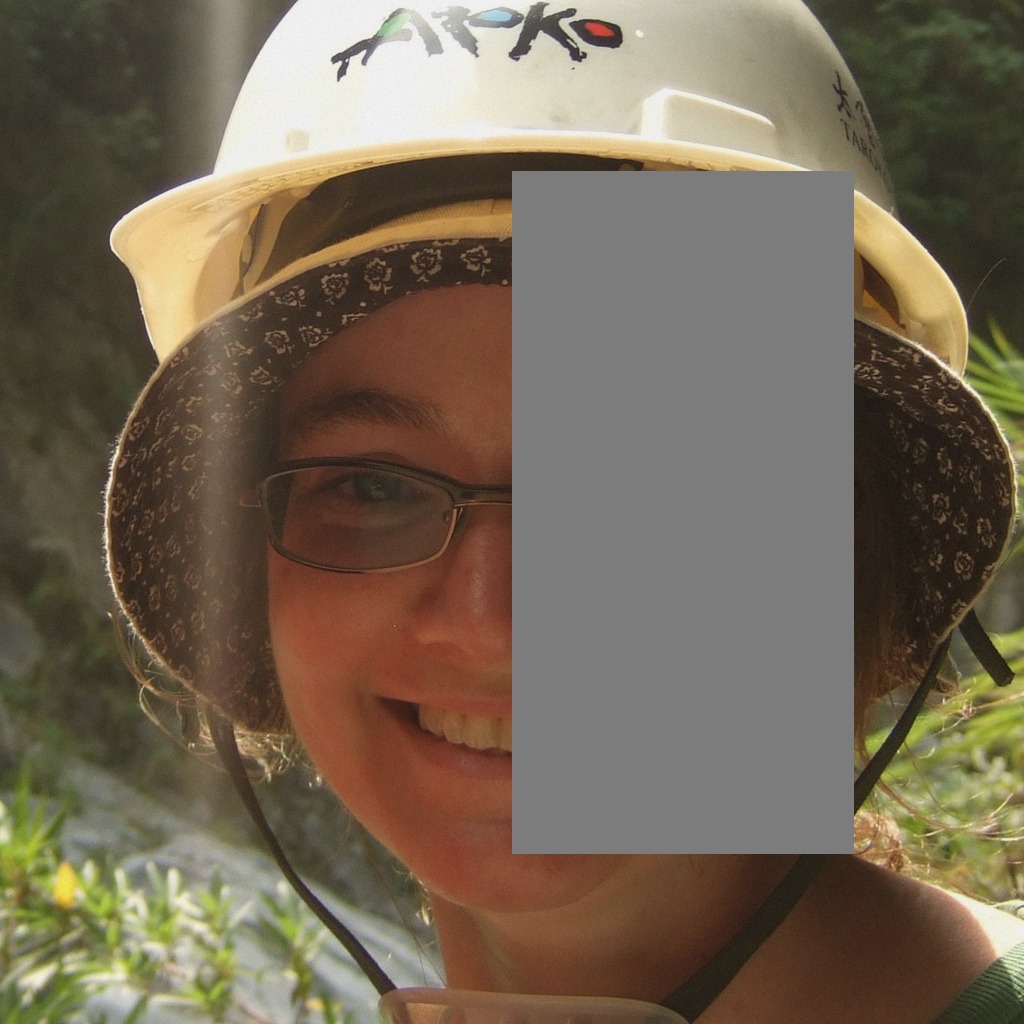} &
        \includegraphics[width=\imwidth]{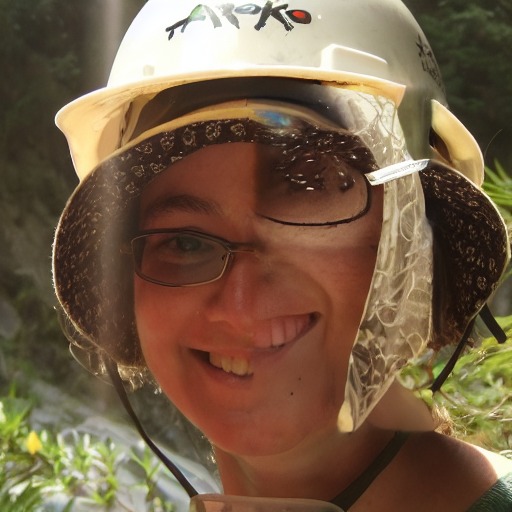} &
        \includegraphics[width=\imwidth]{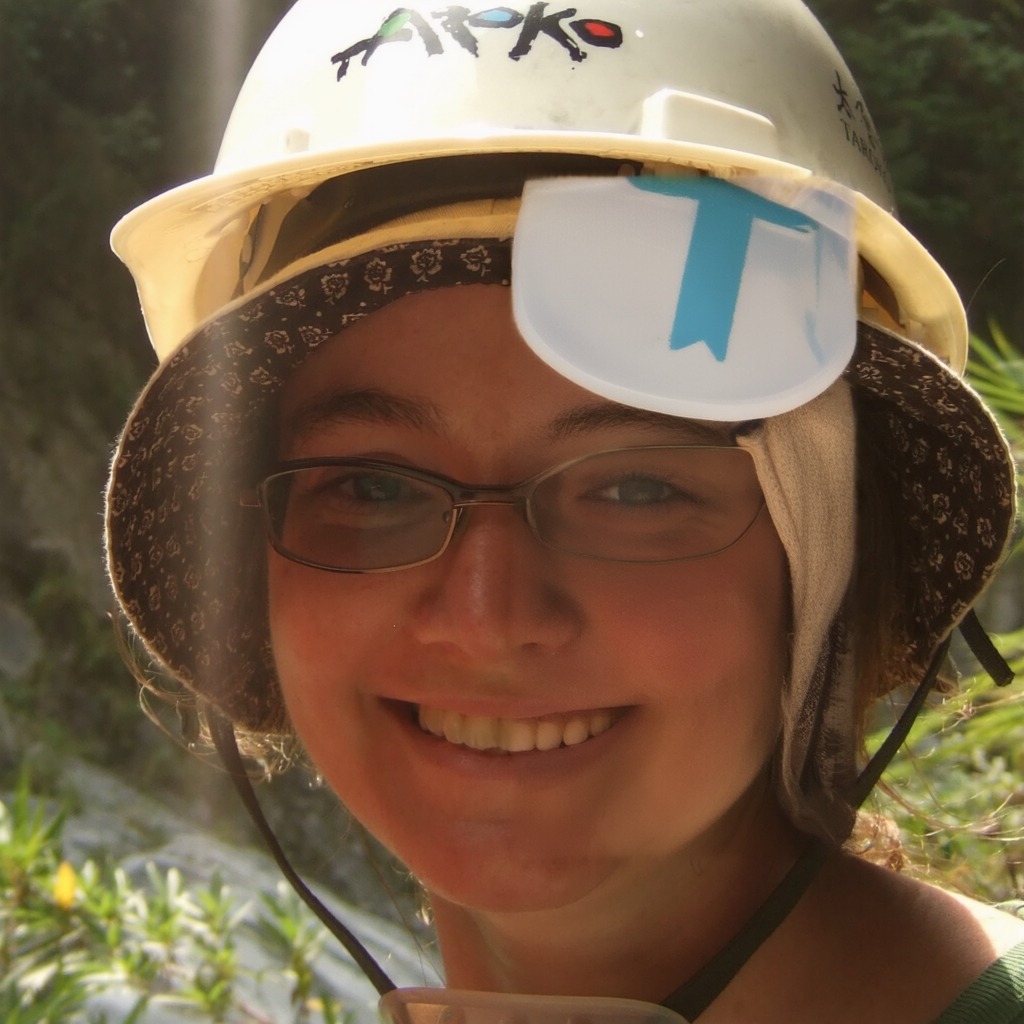} &
        \includegraphics[width=\imwidth]{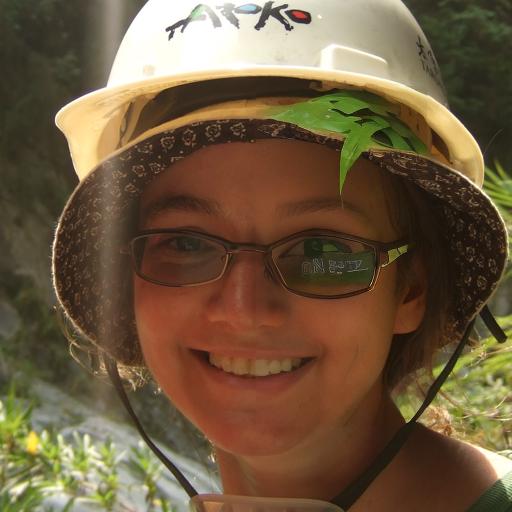} &
        \includegraphics[width=\imwidth]{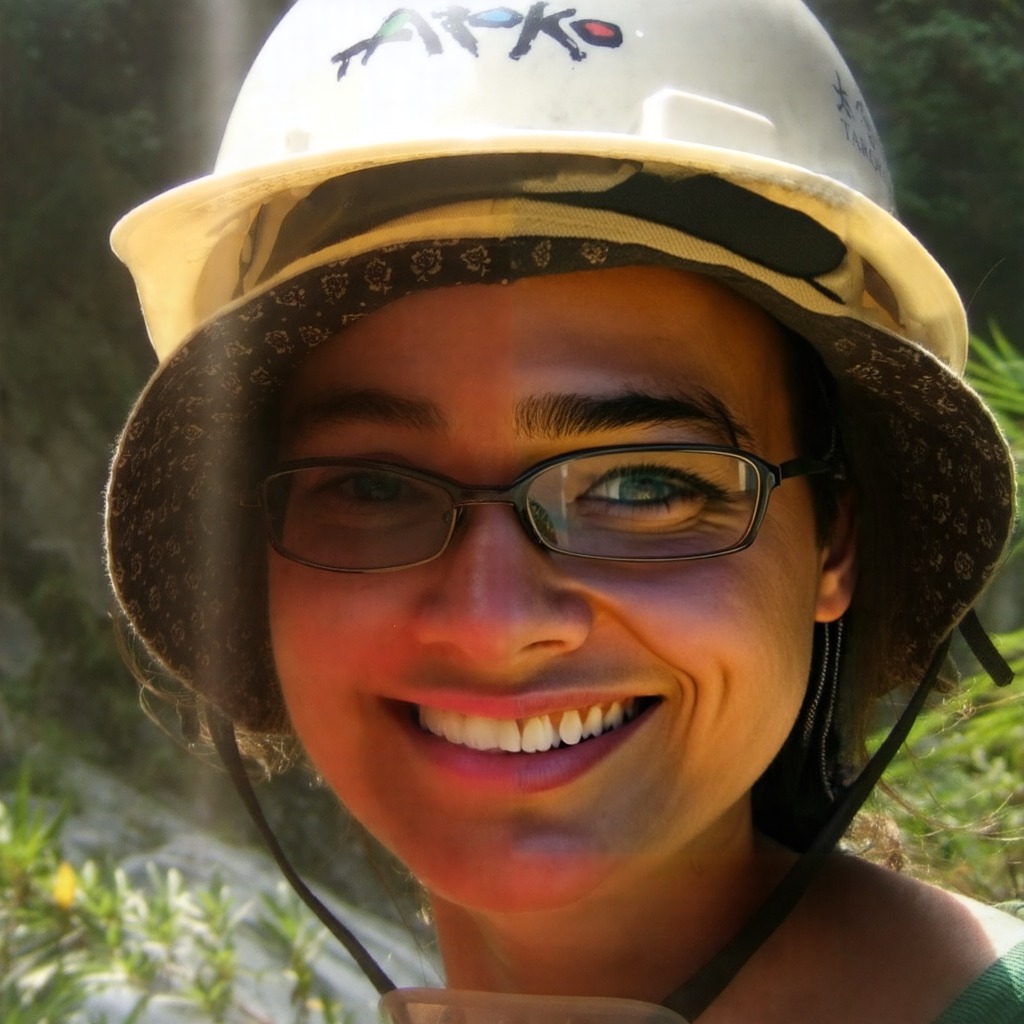} &
        \oursimg{\includegraphics[width=\imwidth]{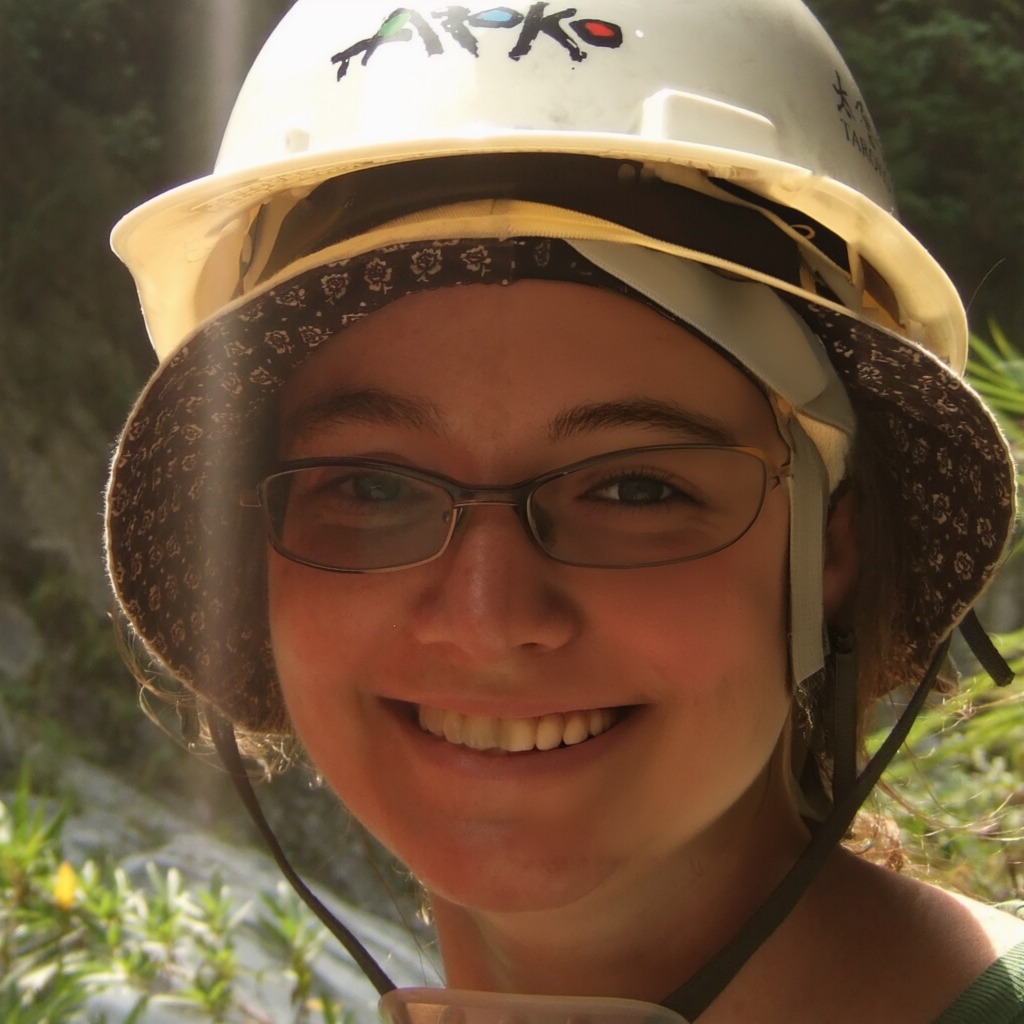}} \\[-2pt]
        {\tiny --} & {\tiny \second{CLIP: 0.282}} & {\tiny \third{CLIP: 0.281}} & {\tiny CLIP: 0.264} & {\tiny \best{CLIP: 0.289}} & {\tiny CLIP: 0.266} \\[2pt]
        {\tiny Input} & {\tiny InitNo} &
        {\tiny \makecell{FreeInpaint\\(SD3.5)}} & {\tiny \makecell{FreeInpaint\\(BrushNet)}} & {\tiny \makecell{SD3\\ControlNet}} &
        {\tiny \bf Our method} \\
    \end{tabular}
    \renewcommand{\arraystretch}{1.0}
    \setlength{\tabcolsep}{6pt}
    \vspace{-0.5em}
    \caption{{\bf Qualitative comparison with CLIP scores --} Our result better matches the lighting, colour and facial structure of the observed region, while scoring lower for CLIP.}
    \label{fig:clip_qualitative}
    \vspace{-1em}
\end{figure}

\vspace{-0.5em}
Note that a higher CLIP score does not mean better outcome; it only measures alignment to prompt, and will not penalize unnatural compositions.
For example, as shown in \cref{fig:clip_qualitative}, inpainted outputs from InitNo, FreeInpaint, and SD3 Inpainting ControlNet~\cite{alimama2024sd3controlnet,controlnet,esser2024} may score higher than ours in terms of CLIP score, but are not
\textit{coherent} with the surroundings.
\vspace{-0.5em}
\enlargethispage{8\baselineskip}

\section{Comparison with trained inpainting baseline}
\label{sec:supp_controlnet}
We compare our method against SD3 ControlNet~{\cite{alimama2024sd3controlnet,controlnet,esser2024}} Inpainting on FFHQ. We use the 1024 resolution to be compatible with a pre-trained SD3 ControlNet. Our method outperforms ControlNet by a large margin in PSNR, SSIM, LPIPS, FID, and AS.

For IR, HPSv2, and CLIP, ControlNet scores are higher, but they are not well aligned with the surrounding pixels, as indicated by SSIM, LPIPS, and FID; see \cref{tab:controlnet_vs_ours} for quantitative results and \cref{fig:ffhq_comparison} for qualitative comparison as well.%

\begin{table}[H]
\centering
\setlength{\tabcolsep}{8pt}
\resizebox{\linewidth}{!}{%
\begin{tabular}{@{}lcccccccc@{}}
\toprule
Method & PSNR$\uparrow$ & SSIM$\uparrow$ & LPIPS$\downarrow$ & FID$\downarrow$ & IR$\uparrow$ & HPS v2$\uparrow$ & CLIP$\uparrow$ & AS$\uparrow$\\
\midrule
ControlNet (50 NFEs) & 19.068 & 0.761 & 0.199 & 21.341 & \best{0.451} & \best{0.250} & \best{24.732} & 5.710 \\
\textbf{Our Method} & \best{22.562} & \best{0.857} & \best{0.121} & \best{12.883} & 0.280 & 0.245 & 24.042 & \best{5.848} \\
\bottomrule
\end{tabular}
}
\setlength{\tabcolsep}{6pt}
\caption{{\bf Comparison with SD3 ControlNet inpainting on FFHQ ($1024{\times}1024$) -- } Our method outperforms ControlNet by a large margin in PSNR, SSIM, LPIPS, FID, and AS. ControlNet scores higher in IR, HPSv2, and CLIP, but the concurrent gaps in SSIM, LPIPS, and FID indicate poor alignment with the unmasked region.}
\label{tab:controlnet_vs_ours}
\end{table}

\section{Additional ablations}
\label{sec:supp_ablations}

\subsection{Effect of number of optimization iterations}
We evaluate inpainting quality on FFHQ as a function of the number of optimization iterations.
As shown in \cref{tab:iteration-count}, the performance improves from 10 to 20 iterations, with diminishing returns beyond that.

\begin{table}[H]
\vspace{-0.5em}
\centering
\small
\setlength{\tabcolsep}{10pt}
\resizebox{\linewidth}{!}{%
\begin{tabular}{@{}lcccccccc@{}}
\toprule
Num of Iterations & PSNR$\uparrow$ & SSIM$\uparrow$ & LPIPS$\downarrow$ & FID$\downarrow$ & IR$\uparrow$ & HPS v2$\uparrow$ & CLIP$\uparrow$ & AS$\uparrow$\\
\midrule
10 (200NFE) & \third{22.088} & \second{0.855} & \third{0.127} & \best{12.783} & \second{0.096} & \second{0.230} & \third{20.464} & \second{5.825} \\
20 (400NFE) & \best{22.562} & \best{0.857} & \best{0.121} & \third{12.883} & \best{0.114} & \best{0.233} & \best{20.773} & \best{5.849} \\
30 (600NFE) & \second{22.543} & \third{0.854} & \second{0.122} & \second{12.840} & \third{0.094} & \third{0.226} & \second{20.754} & \third{5.802} \\
\bottomrule
\vspace{0.2em}
\end{tabular}

}
\setlength{\tabcolsep}{6pt}
\caption{{\bf Effect of number of optimization iterations on FFHQ} --  Performance improves from 10 to 20 iterations, with diminishing returns beyond that.}
\label{tab:iteration-count}
\end{table}

\section{Prompts}
\label{sec:supp_prompt}

We provide the exact prompts to reproduce results in the paper.

\subsection{Prompt for example figures}

The following prompt was used for \cref{fig:init_noise} and \cref{fig:freq_optimization}.

\begin{lstlisting}[style=promptstyle]
A high quality, 4K, realistic image of bright red mushrooms with white speckles growing in a cluster at the base of a smooth gray tree trunk. Fallen orange and brown autumn leaves cover the ground with patches of green grass and a couple of round gray stones nearby. Yellow-green leaves hang above.
\end{lstlisting}

\subsection{Exact prompt for generating prompts}
Given the masked image directories, we instruct Claude \cite{claude-sonnet-4-5} to generate image prompts using the following templates. Variables denoted in \textcolor{blue}{\textbf{blue}} are populated programmatically in a loop.

\paragraph{FFHQ}
\begin{lstlisting}[style=promptstyle]
Manually generate caption ID @[INDEX]@ of the folder @[Masked Image Folder]@ and output each caption to @[Output Text File Path]@ as caption_@[INDEX]@.txt. Process one image at a time (not in batches) and overwrite existing files. Each caption must be 2--3 concise sentences that describe the complete, fully restored scene as it should look after inpainting. Use the visible context to infer plausible 
colors, materials, geometry, and lighting;
extend existing structures smoothly and remain consistent with the overall style of the source image. In the final prompt that you output, definitely remove any descriptions related to "mask" or "grey area". We want a prompt that describes the 
final inpainted image without the mask. Also, remove anything about symmetrical in
the caption. Be succinct, using only a few
sentences. Directly use your write operation. Do not do any bash commands such as cat and >>.
\end{lstlisting}

\paragraph{DIV2K}
\begin{lstlisting}[style=promptstyle]
Manually generate one caption per PNG inside @[Masked Image Folder]@, processing the images sequentially and never 
batching. Reference the mask layout encoded in @[Mask Tensor Path]@ so you can infer what content should appear in the hidden region, but do not mention the mask 
itself in the final writing. For each image, write the caption directly (no shell redirection) to @[Output Caption Folder]@/caption_@[image_INDEX]@.txt (e.g., caption_00039.txt). Each caption must be 2--3 tight sentences that describe the complete, fully restored scene as it should look after inpainting. Use the visible context to infer plausible colors, 
materials, geometry, and lighting; extend existing structures smoothly and stay consistent with the overall style of the source image. Do not introduce new objects, camera moves, or artistic flourishes that are not already implied. Absolutely avoid words such as "mask," "grey area," "missing," "occluded," or any hint that parts of the image were blocked out---the caption should read like a finished description of the final inpainted result only.
\end{lstlisting}

\section{Additional qualitatives}
\label{sec:supp_extra}

\begin{figure*}[p]
    \centering
    \includegraphics[width=\linewidth, height=0.92\textheight, keepaspectratio]{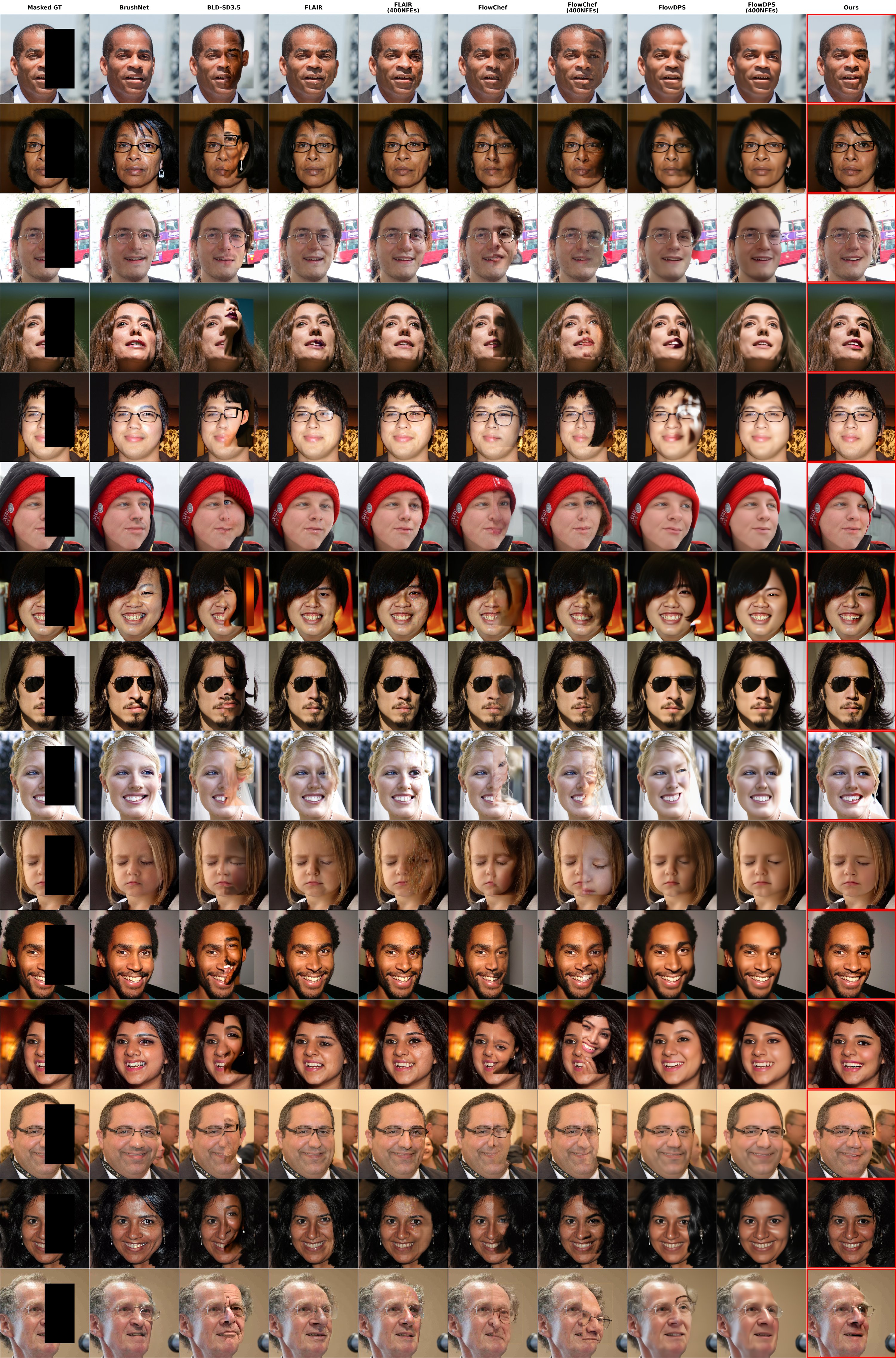}
    \captionsetup{width=0.8\textwidth}
    \caption{Qualitative comparison of inpainting results on FFHQ dataset. 15 samples drawn at random from the full set of 1000 outputs.}
    \label{fig:ffhq_comparison}
\end{figure*}

\begin{figure*}[p]
    \centering
    \includegraphics[width=\linewidth, height=0.92\textheight, keepaspectratio]{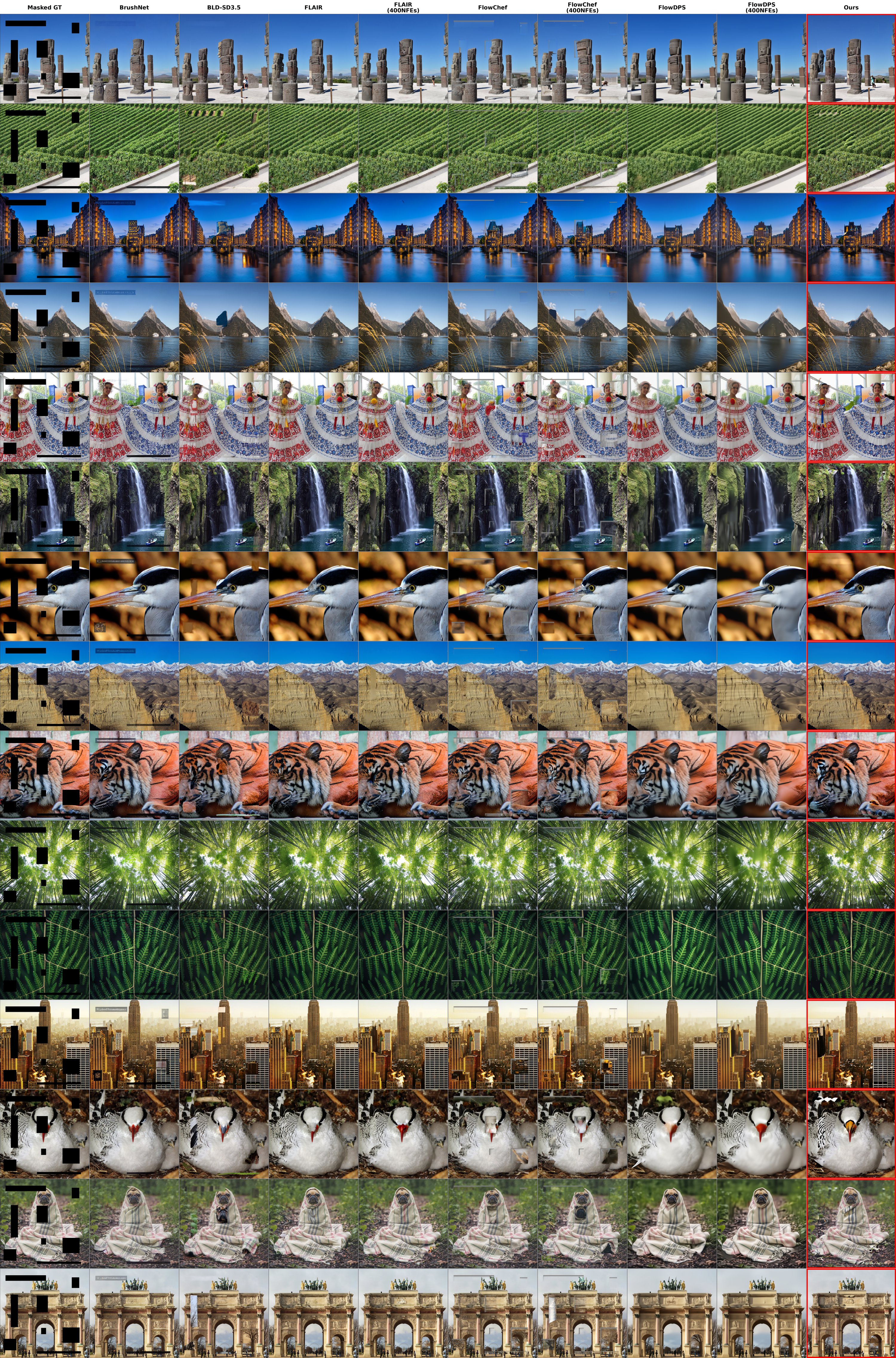}
    \captionsetup{width=0.8\textwidth}
    \caption{Qualitative comparison of inpainting results on DIV2K dataset. 15 samples drawn at random from the full set of 800 outputs.}
    \label{fig:div2k_comparison}
\end{figure*}

\begin{figure*}[p]
    \centering
    \includegraphics[width=\linewidth, height=0.92\textheight, keepaspectratio]{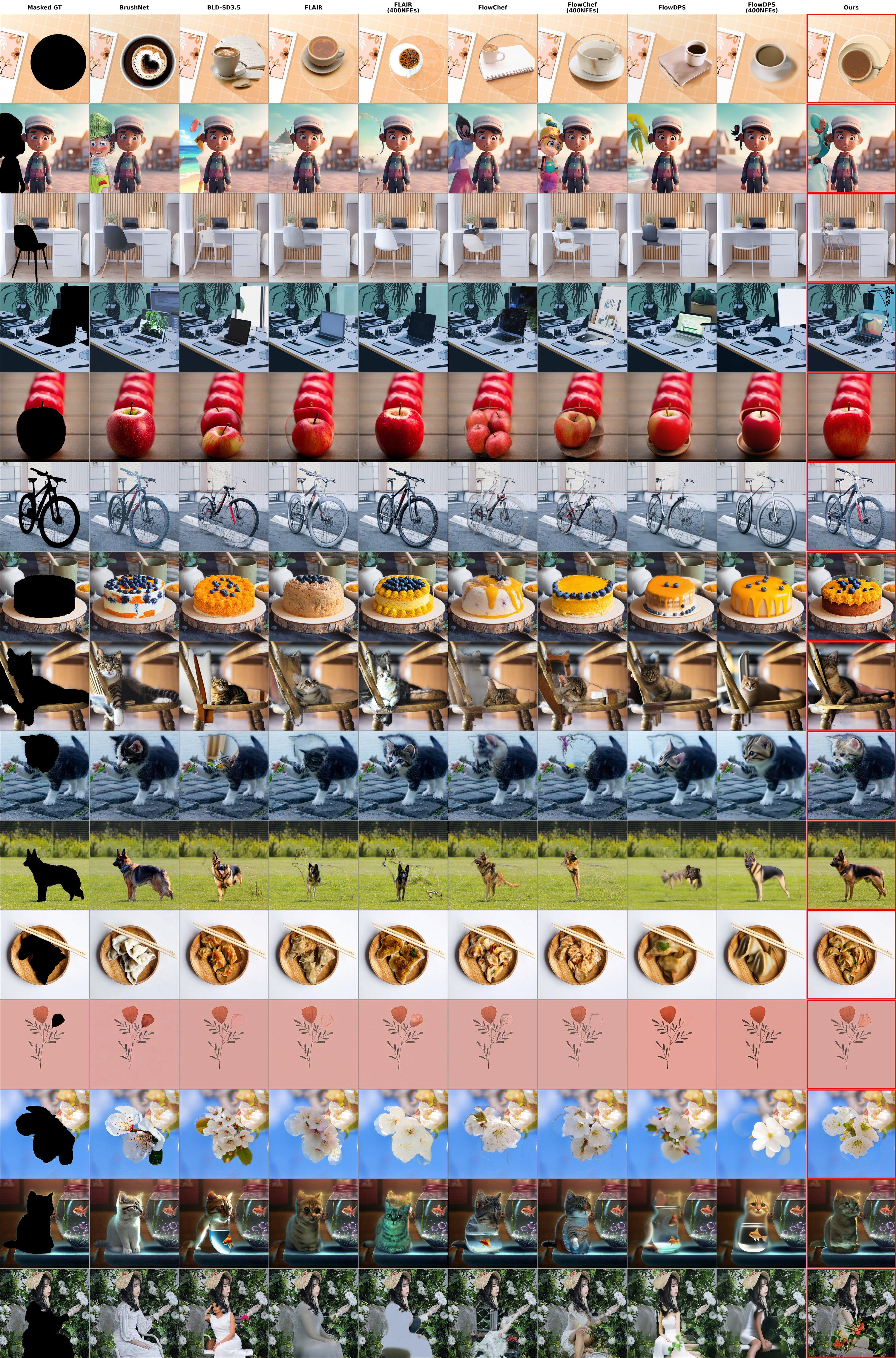}
    \captionsetup{width=0.8\textwidth}
    \caption{Qualitative comparison of inpainting results on BrushBench dataset. 15 samples drawn at random from the full set of 600 outputs.}
    \label{fig:brushbench_comparison}
\end{figure*}

To provide a fair qualitative comparison, we show randomly selected samples of inpainted results for all baselines and our method on FFHQ (\cref{fig:ffhq_comparison}), DIV2K (\cref{fig:div2k_comparison}), and BrushBench (\cref{fig:brushbench_comparison}).
Zoom in for better detail.

\end{document}